        \documentclass{article}

        \usepackage{arxiv}

        \usepackage{iftex}
        \ifLuaTeX
          \usepackage[haranoaji,nfssonly]{luatexja-preset}
          \newcommand{\japanesetext}[1]{#1}
        \else
          \usepackage[utf8]{inputenc}
          \usepackage[T1]{fontenc}
          \usepackage{CJKutf8}
          \newcommand{\japanesetext}[1]{\begin{CJK*}{UTF8}{min}#1\end{CJK*}}
        \fi

        \usepackage{amsmath,amssymb}
        \usepackage{booktabs}
        \usepackage{longtable}
        \usepackage{array}
        \usepackage{calc}
        \usepackage{graphicx}
        \usepackage{caption}
        \usepackage{enumitem}
        \usepackage{newunicodechar}
        \usepackage{xurl}
        \usepackage{fvextra}
        \usepackage{etoolbox}
        \usepackage[unicode,hidelinks]{hyperref}

        \newunicodechar{≥}{\ensuremath{\ge}}
        \newunicodechar{≤}{\ensuremath{\le}}
        \newunicodechar{×}{\ensuremath{\times}}
        \newunicodechar{±}{\ensuremath{\pm}}
        \newunicodechar{≈}{\ensuremath{\approx}}
        \newunicodechar{χ}{\ensuremath{\chi}}
        \newunicodechar{²}{\ensuremath{^2}}
        \newunicodechar{−}{\ensuremath{-}}

        \DefineVerbatimEnvironment{verbatim}{Verbatim}{breaklines=true,breakanywhere=true,fontsize=\small}
        \AtBeginEnvironment{longtable}{\small}
        \providecommand{\tightlist}{\setlength{\itemsep}{0pt}\setlength{\parskip}{0pt}}

        \newlength{\cslhangindent}
        \newenvironment{CSLReferences}[2]{%
          \setlength{\parindent}{0pt}%
          \ifodd #1 \setlength{\leftskip}{\cslhangindent}\setlength{\parindent}{-\cslhangindent}\fi
          \setlength{\parskip}{0.6em}%
        }{}

        \title{Distinct Profiles of Run-to-Run Score Reliability and Expert-Panel Alignment Across Four LLM Evaluators of Simulated Japanese-Language AI-to-AI Counseling}
        \author{Keita Kiuchi\textsuperscript{1}\thanks{Correspondence concerning this article should be sent to Keita Kiuchi, Japan National Institute of Occupational Safety and Health, Organization of Occupational Health and Safety, Tama-ku Nagao 6-21-1, Kawasaki, Kanagawa 214-0023, Japan. Email: \texttt{kiuchi@h.jniosh.johas.go.jp}}, Yoshikazu Fujimoto\textsuperscript{2}, Hideyuki Gotō\textsuperscript{3}, Tomonori Hosokawa\textsuperscript{4},\\
\textbf{Makoto Nishimura\textsuperscript{5}, Yōsuke Satō\textsuperscript{6}, Izumi Sezai\textsuperscript{7}, Tomohiro Inoue\textsuperscript{8}}\\[0.5em]
\normalfont\small
\textsuperscript{1}Japan National Institute of Occupational Safety and Health, Kawasaki, Kanagawa, Japan\\
\textsuperscript{2}Kaze To Taiyo, Tokyo, Japan\\
\textsuperscript{3}Saga Occupational Health Association, Saga, Japan\\
\textsuperscript{4}Zikei Hospital/Zikei Institute of Psychiatry, Okayama, Japan\\
\textsuperscript{5}Suzuka University of Medical Science, Suzuka, Mie, Japan\\
\textsuperscript{6}Ritsumeikan University, Ibaraki, Osaka, Japan\\
\textsuperscript{7}National Defense Medical College, Tokorozawa, Saitama, Japan\\
\textsuperscript{8}Aoyama Gakuin University, Tokyo, Japan}
        \date{August 27, 2026}

        \renewcommand{\headeright}{Preprint}
        \renewcommand{\undertitle}{Preprint}
        \renewcommand{\shorttitle}{LLM Evaluator Reliability and Alignment}

        \hypersetup{
          pdftitle={Distinct Profiles of Run-to-Run Score Reliability and Expert-Panel Alignment Across Four LLM Evaluators of Simulated Japanese-Language AI-to-AI Counseling},
          pdfsubject={Run-to-run reliability and expert-panel alignment of LLM evaluators of counseling dialogue},
          pdfauthor={Keita Kiuchi, Yoshikazu Fujimoto, Hideyuki Goto, Tomonori Hosokawa, Makoto Nishimura, Yosuke Sato, Izumi Sezai, Tomohiro Inoue},
          pdfkeywords={large language models, automated evaluation, motivational interviewing, expert judgment, AI counseling simulation},
        }

\begin{document}
\maketitle

\begin{abstract}
\textbf{Introduction:} Large language models (LLMs) increasingly evaluate generated dialogue, but reliable scores across repeated runs do not establish alignment with professional judgment.

\textbf{Methods:} This observational fixed-benchmark study compared four configured LLM evaluator systems---GPT-5.5, Gemini 3.5 Flash, Claude Opus 4.8, and Fable 5---with aggregated ratings from 15 counseling experts on 18 complete simulated artificial intelligence (AI)-to-AI counseling sessions conducted in Japanese. Sessions represented three counselor conditions across six prespecified client-profile contexts. Each system scored every transcript three times on four motivational interviewing--informed dimensions and overall quality.

\textbf{Findings:} All four LLM systems assigned higher scores than did the aggregated expert panel for softening sustain talk and overall quality, although score differences varied across systems and constructs. Single-run intraclass correlation coefficients ranged from .33 to .96; high run-to-run score reliability did not ensure closer expert-panel alignment. Claude Opus 4.8 showed the smallest mean absolute difference, whereas Fable 5 showed an intermediate mean absolute difference. In a secondary benchmark analysis, GPT-4-turbo sessions generated with the Structured Multi-step Dialogue Prompt received higher expert-panel ratings than sessions generated by the same model with a minimal instruction on cultivating change talk, partnership, empathy, and overall quality; the contrast for softening sustain talk remained uncertain.

\textbf{Discussion:} The systems showed distinct system- and construct-specific profiles of run-to-run score reliability and expert-panel alignment.

\textbf{Limitations:} The fixed benchmark contained one complete session per counselor-condition-by-profile cell and did not estimate performance across all possible stochastic regenerations.

\textbf{Conclusions:} Run-to-run score reliability, expert-panel alignment, and condition discrimination are separate properties of automated counseling evaluation. This benchmark provides a concrete basis for assessing LLM evaluators against professional judgment at the construct level.
\end{abstract}


\hypertarget{introduction}{%
\section{Introduction}\label{introduction}}

Large language models (LLMs), a form of artificial intelligence (AI), are increasingly examined as tools for counseling simulation and skills training and as evaluators of generated dialogue (Maurya, 2024; Qiu \& Lan, 2024; Stade et al., 2024). Mental health chatbot research indicates that engagement, user attitudes, and outcome evidence should be evaluated rather than inferred from conversational availability or positive impressions alone (Limpanopparat et al., 2024), and perceptions of LLM-generated advice vary with advice style and user characteristics (Wester et al., 2024). The capacity to generate fluent, supportive language does not establish counseling quality, and the capacity to assign consistent scores does not establish valid evaluation. A dialogue may appear empathic and orderly while overlooking ambivalence, moving prematurely toward advice, or failing to preserve client autonomy. When generative models both produce and judge counseling dialogue, stable automated scores may therefore encourage unwarranted trust in the evaluation process (Volpato et al., 2025).

Motivational interviewing (MI) provides a framework for separating relational qualities---empathy, collaboration, and respect for autonomy---from technical responsiveness to change talk and sustain talk (Magill et al., 2018; Miller \& Rose, 2009). The \emph{Motivational Interviewing Treatment Integrity Coding Manual} 4.2.1 (MITI; Moyers et al., 2014) operationalizes observable aspects of these processes. Partnership and Empathy primarily represented relational qualities in this study, whereas Cultivating Change Talk and Softening Sustain Talk represented technically demanding responses to client language. An LLM may reproduce the surface form of empathic communication without responding contingently to the psychological meaning of the client's statements. MITI-derived measures have been applied to AI-generated counseling dialogue (Cohen et al., 2024; Steenstra et al., 2024; Yang et al., 2025), although the adapted global ratings used here omitted behavior counts and should not be interpreted as full MITI fidelity coding.

\hypertarget{automated-evaluation-of-counseling-quality}{%
\subsection{Automated Evaluation of Counseling Quality}\label{automated-evaluation-of-counseling-quality}}

Automated evaluation is not a single property. An evaluator may reproduce its scores across repeated calls while remaining systematically shifted from professional judgments, and it may match average score levels while discriminating dialogue conditions differently from experts. LLM-based scoring should therefore distinguish run-to-run score reliability, expert-panel alignment, and discrimination among counseling conditions. Agreement with human ratings varies across models, datasets, and evaluated properties (Bavaresco et al., 2025). In synthetic conversations with a mental health chatbot, the model with the highest classification accuracy differed from that with the strongest human-rating correlation (Reichenpfader et al., 2026). MI is a demanding test because an evaluator may reward supportive wording, coherent structure, or movement toward planning while missing implicit emotion, competing values, sustain talk, or client-generated reasons for change. Comparing several configured evaluator systems with an aggregated expert panel on the same transcripts can reveal whether discrepancies are uniform, system-specific, or construct-specific and whether high run-to-run score reliability coexists with systematic differences from expert-panel scores.

\hypertarget{counselor-conditions-and-simulated-clients}{%
\subsection{Counselor Conditions and Simulated Clients}\label{counselor-conditions-and-simulated-clients}}

The Structured Multi-step Dialogue Prompt (SMDP) provided benchmark variation by guiding the interaction from the present concern through a desired future, prior attempts, resources, and possible next steps and by requiring two reflections before each counselor question. Such procedural scaffolding may reduce interrogative turn-taking and create opportunities for change talk, empathy, and partnership (Huang et al., 2024; Lee et al., 2024). It does not, however, ensure responsiveness to emotion, values, or ambivalence. Comparing the counselor conditions therefore addressed a secondary substantive question about structured procedural prompting and provided benchmark variation for examining evaluator discrimination. The benchmark's interpretability also depends on whether simulated clients express enough uncertainty, resistance, and sustain talk for demanding MI behaviors to be observed.

Japanese-language counseling dialogue remains understudied. Politeness, ellipsis, sentence-final expressions, and listener responses may shape perceptions of empathy and directiveness. Japanese attentive-listening research has linked backchannel variation with perceived naturalness and understanding (Yamaguchi et al., 2016), and Japanese MITI research has emphasized rater training (Ohtsubo et al., 2016). Japanese-language dialogue therefore provides an understudied context in which findings based primarily on English interactions should be examined independently. Because the present study did not compare languages or cultures, it makes no claim about cross-cultural differences.

\hypertarget{present-study}{%
\subsection{Present Study}\label{present-study}}

The primary methodological objective was to determine how ratings from four configured LLM evaluator systems---GPT-5.5, Gemini 3.5 Flash, Claude Opus 4.8, and Fable 5---differed from the aggregated ratings of a panel of counseling experts for the same 18 complete Japanese-language AI-to-AI counseling sessions. We examined run-to-run score reliability, expert-panel alignment, and discrimination among counselor conditions as distinct evaluator properties. A secondary objective was to compare counselor conditions within this benchmark. GPT-minimal and GPT-SMDP held the GPT-4-turbo counselor model constant while varying the procedural instruction; Claude-SMDP provided a cross-model SMDP implementation. The evaluated object was the complete evolving interaction rather than isolated counselor responses to standardized client utterances.

The three preregistered hypotheses were: GPT-SMDP would receive higher expert ratings than GPT-minimal (H1); Claude-SMDP would receive higher expert ratings than GPT-SMDP (H2); and expert ratings would differ from the four LLM evaluator systems (H3). The evaluator-type comparison specified in H3 provides the manuscript's primary methodological contribution, whereas H1 and H2 address secondary benchmark questions. The prespecified mixed models served as the primary analyses; additional sensitivity analyses are described below. Simulated-client naturalness and qualitative failure modes were exploratory.

\hypertarget{methods}{%
\section{Methods}\label{methods}}

\hypertarget{counseling-script-development}{%
\subsection{Counseling Script Development}\label{counseling-script-development}}

The benchmark comprised 18 complete Japanese-language AI-to-AI counseling sessions: three counselor conditions crossed with six prespecified profiles (women and men in their 20s, 40s, and 60s). Each condition was represented by six separately initiated sessions, one per profile. GPT-minimal used GPT-4-turbo with a minimal role instruction, GPT-SMDP used GPT-4-turbo with SMDP, and Claude-SMDP used the February 2024 Claude-3-Opus API release with SMDP. All client simulations used Claude-3-Opus and work-related concern prompts.

After the initial client message, each session proceeded through 10 sequential counselor--client exchanges. The counselor response and subsequent client response were generated in separate model calls, with accumulating dialogue histories supplied at the next exchange. Thus, each transcript contained repeated, history-conditioned generations within a dynamically evolving interaction. Individual turns were not treated as independent replications, and client language differed across counselor conditions in response to preceding counselor turns. The evaluated object was complete-interaction quality across the six profile contexts, not isolated responses to identical client inputs.

A custom program maintained the two dialogue histories through the OpenAI Chat Completions and Anthropic Messages APIs; all transcripts were generated on April 12, 2024. Full prompts, model identifiers, sampling settings, history updates, output limits, and retry settings appear in Online Resource 1, Section S1. The GPT-minimal versus GPT-SMDP comparison isolates prompt structure within one counselor model; comparisons involving Claude-SMDP also include model differences.

\hypertarget{expert-evaluation}{%
\subsection{Expert Evaluation}\label{expert-evaluation}}

Fifteen counseling experts rated the transcripts. The panel included nonauthor experts and coauthors other than the first and last authors, who did not rate the materials. Eleven experts (73\%) had at least 10 years of counseling experience, and eight (53\%) held Japanese clinical psychologist or certified psychologist qualifications (Table 1). Nonauthor evaluators were recruited through Japanese MI practitioner networks based on professional background and self-reported familiarity with MI and MITI; they received 500 Japanese yen per three-script evaluation unit.

Each blinded unit contained all three counselor conditions for one client profile. Questionnaire randomization determined display order. Evaluators saw only the labels Moon, Star, and Sun and were not told which counselor condition each label represented. The label-to-condition mapping used for analysis was constant across profiles and evaluators. On 1--5 scales permitting 0.5-point increments, they rated Cultivating Change Talk (CCT), Softening Sustain Talk (SST), Partnership (PAR), Empathy (EMP), and custom Overall evaluation (OVR). The first four measures were adapted from MITI global scales; behavior counts were not coded. Evaluators also rated client naturalness from 0 to 10 and provided comments. The 54 completed units yielded 162 ratings per outcome; each transcript received 8--10 ratings (median = 9).

Evaluator-to-profile assignment was sequential and workload based, with an informal attempt to distribute age and sex profiles; it was not randomized or formally counterbalanced. Every completed unit nevertheless provided within-evaluator ratings of all three conditions for the same profile. The study was designed as a general expert-panel evaluation using MITI-informed global dimensions rather than as a formal MITI fidelity assessment. In prior MITI reliability studies, formal coder training involved 39--40 hours of structured instruction and practice (Forsberg et al., 2007; Moyers et al., 2016). Even after such training, interrater reliability for some dimensions was only fair; Japanese preliminary evidence likewise suggests that global ratings and behavior counts may have different training requirements (Forsberg et al., 2007; Moyers et al., 2016; Ohtsubo et al., 2016). As AI-generated counseling becomes increasingly sophisticated, differences among counseling dialogues may become more difficult for evaluators to distinguish, yet repeatedly assembling and training a dedicated MITI coding team for each study would be difficult to sustain as a general evaluation strategy. The study therefore did not conduct a formal calibration exercise to align evaluators with a study-specific scoring interpretation; instead, the transcript-level mean of ratings made separately by multiple counseling professionals served as the professional comparison standard rather than as an error-free gold standard or formal MITI fidelity criterion (Online Resource 1, Section S2).

\textbf{{[}Table 1 about here{]}}

\hypertarget{llm-evaluation-systems}{%
\subsection{LLM Evaluation Systems}\label{llm-evaluation-systems}}

OpenAI GPT-5.5, Google Gemini 3.5 Flash, and Anthropic Claude Opus 4.8 served as API-based evaluators (Anthropic, 2026b; Google DeepMind, 2026; OpenAI, 2026). On June 24--25, 2026 (UTC), each system rated every transcript three times on the five 1--5 scales using the provider's maximum reasoning or effort setting. These were repeated stochastic outputs, not independent human-like raters. Each metric-specific call supplied the same transcript, prompt body, and rubric and returned a numeric score and reason. Provider-native message-role structures differed, and temperature was not explicitly set. The three systems produced 810 numeric ratings and, in separate post-scoring calls, 54 qualitative summaries and 54 suggested dialogue revisions.

Anthropic restored global access to Fable 5 on July 1, 2026, including through Claude Platform and Claude Code, and its platform documentation identifies \texttt{claude-fable-5} as the Claude API model ID (Anthropic, 2026a, 2026c). At the time of evaluation, however, Fable 5 was not exposed through the study account's Anthropic Console API access. The evaluations included in the analyses were therefore conducted through a Claude Code subscription. Each rating used an isolated, single-turn process with maximum reasoning effort; tools, repository instructions, conversation persistence, and the coding-oriented system prompt were disabled. Fable received the same counseling-supervisor instruction, metric prompts, criteria, and transcripts used for the API systems. Three runs for 18 transcripts and five outcomes yielded 270 numeric ratings. Subsequent calls generated 18 qualitative summaries and 18 dialogue-specific improvement evaluations after deterministic integration of the three score runs. Model identity was verified for each accepted evaluation. Only the Claude Code evaluations were included; an earlier GUI evaluation was excluded from the numeric and qualitative analyses. Fable model identity cannot be separated from its Claude Code pathway. Complete settings, prompts, formats, and qualitative procedures appear in Online Resource 1, Section S3.

\hypertarget{statistical-analysis}{%
\subsection{Statistical Analysis}\label{statistical-analysis}}

A transcript-level sensitivity analysis specified after preliminary mixed-model inspection assessed alignment with the aggregated expert panel. For each transcript and outcome, the expert-panel mean was compared with each configured LLM system's three-run mean. Mean signed LLM-minus-expert differences represented directional score shifts, and mean absolute differences represented distance from the expert-panel aggregate. Condition-stratified transcript bootstrap intervals summarized variation across the 18 archived transcript summaries rather than across all possible generated dialogues.

The OSF-specified primary model-based analysis assessed whether evaluator-type differences varied across counselor conditions. For each outcome, the mixed model included evaluator type, counselor condition, their interaction, client age, and client sex, with crossed evaluator and transcript random intercepts. Likelihood-ratio tests evaluated the interaction, and Tukey--Kramer-adjusted simple effects compared evaluator types within condition. Type III Wald tests were reported as supplementary summaries. Run-to-run score reliability was summarized separately by ICC(2,1), ICC(2,\emph{k}) (Koo \& Li, 2016), and variance components; LLM ICCs describe repeated stochastic outputs from the same configured system, not agreement among independent raters.

For the secondary counselor-condition question, expert-rated outcomes were analyzed with mixed models containing counselor condition, client age, and client sex, with evaluator and transcript random intercepts. Likelihood-ratio tests evaluated counselor condition, and Tukey--Kramer-adjusted contrasts compared conditions. Age and sex were descriptive covariates because each profile cell contained one transcript. After preliminary mixed-model inspection, we specified additional sensitivity analyses using within-evaluator-by-profile paired differences, Holm adjustment within outcomes, Benjamini--Hochberg adjustment across 15 contrasts, robust standard errors, an evaluation-unit model, and ordinal mixed models. Full specifications and diagnostics appear in Online Resource 1, Sections S5--S6.

Study size was determined by the archived 18-session benchmark and completed expert evaluations rather than by a prospective sample-size calculation. The 162 expert observations per outcome were repeated ratings of those sessions. No valid planned rating was excluded; all planned LLM ratings were complete, whereas the expert matrix was intentionally incomplete because evaluators completed different workloads. The inferential target was the configured systems' and expert panel's ratings of this benchmark, including counselor-condition differences across six profile contexts---not repeated-generation variability within identical cells or average performance across all possible regenerations. Analyses used R 4.4.1 with lme4 1.1-37, lmerTest, emmeans 1.11.0, and psych. Preregistration and amendments are available at https://doi.org/10.17605/OSF.IO/VU286.

\hypertarget{ethical-considerations}{%
\subsection{Ethical Considerations}\label{ethical-considerations}}

The study used synthetic dialogues and professional evaluations of those materials. No patients or counseling clients were recruited, and no clinical, identifiable, or private behavioral data were collected. The expert evaluators acted as professional collaborators or subject-matter contributors who judged synthetic materials, not as the target population from whom psychological or health outcomes were measured. Individual rating and comment records were anonymized for analysis and reporting, and written informed consent was obtained from the expert evaluators. On this basis, the authors judged that formal ethics review was not required; however, no institutional determination of exemption or non-human-participant status was obtained.

\hypertarget{findings}{%
\section{Findings}\label{findings}}

\hypertarget{dataset-and-score-overview}{%
\subsection{Dataset and Score Overview}\label{dataset-and-score-overview}}

The benchmark contained 18 complete sessions, with six sessions in each counselor condition. Fifteen experts completed 54 evaluator-by-profile units, yielding 162 expert ratings per outcome; each transcript received 8--10 expert ratings (median = 9). Each of the four configured LLM evaluator systems completed three scoring runs for every transcript and outcome, yielding 216 LLM ratings per outcome and 1,080 LLM ratings overall. Table 2 summarizes scores by counselor condition and evaluator. Across conditions, GPT-5.5 produced uniformly high scores (OVR \emph{M} = 4.48, \emph{SD} = 0.34), Gemini 3.5 Flash produced the highest OVR and SST means (4.88 and 4.64, respectively), and Claude Opus 4.8 was descriptively closest to the aggregated expert panel on CCT and PAR. Fable 5 produced an OVR mean of 4.31 (\emph{SD} = 0.34) and an SST mean of 4.13 (\emph{SD} = 0.26).

\textbf{{[}Table 2 about here{]}}

\hypertarget{llm-evaluators-showed-system--and-construct-specific-score-differences-from-the-expert-panel}{%
\subsection{LLM Evaluators Showed System- and Construct-Specific Score Differences From the Expert Panel}\label{llm-evaluators-showed-system--and-construct-specific-score-differences-from-the-expert-panel}}

At the transcript level, all four LLM evaluator systems assigned higher OVR and SST scores than the aggregated expert panel. Mean OVR and SST differences were +1.53 and +1.18 for GPT-5.5, +1.93 and +1.50 for Gemini 3.5 Flash, +1.17 and +0.93 for Claude Opus 4.8, and +1.37 and +0.99 for Fable 5; all eight bootstrap intervals excluded zero. Claude Opus 4.8 was near the panel on CCT (−0.08, 95\% bootstrap interval {[}−0.22, 0.05{]}) and PAR (−0.02 {[}−0.13, 0.12{]}), whereas Fable 5 showed differences of +0.12 for CCT, +0.37 for PAR, and +0.62 for EMP. Mean absolute differences across outcomes were 0.60 for Claude Opus 4.8, 0.76 for Fable 5, 1.04 for GPT-5.5, and 1.21 for Gemini 3.5 Flash. These benchmark-specific values do not identify a generally superior underlying model (Figure 1; Online Resource 1, Section S6).

\textbf{{[}Figure 1 about here{]}}

\hypertarget{osf-specified-models-provided-complementary-condition-specific-comparisons}{%
\subsection{OSF-Specified Models Provided Complementary Condition-Specific Comparisons}\label{osf-specified-models-provided-complementary-condition-specific-comparisons}}

In the primary model-based analyses, evaluator differences varied by counselor condition for OVR, χ²(8) = 83.34, \emph{p} \textless{} .001; CCT, χ²(8) = 75.33, \emph{p} \textless{} .001; SST, χ²(8) = 20.16, \emph{p} = .010; and PAR, χ²(8) = 80.74, \emph{p} \textless{} .001. The interaction was not significant for EMP, χ²(8) = 12.92, \emph{p} = .115 (Table 3 and Online Resource 1, Section S6).

All four systems assigned higher OVR and SST scores than the aggregated expert panel within every counselor condition, with all adjusted contrasts significant. Differences for CCT, PAR, and EMP were system specific. GPT-5.5 assigned higher CCT scores than the expert panel for GPT-minimal, higher PAR scores for both SMDP conditions, and higher EMP scores for all conditions. Gemini 3.5 Flash assigned higher CCT and PAR scores than the expert panel for both SMDP conditions and higher EMP scores for Claude-SMDP. Claude Opus 4.8 did not differ significantly on CCT, PAR, or EMP within any condition. Fable 5 assigned higher EMP scores than the expert panel only for GPT-SMDP among these three outcomes. Covariate estimates are descriptive (Online Resource 1, Section S6).

\textbf{{[}Table 3 about here{]}}

\hypertarget{run-to-run-score-reliability-did-not-ensure-expert-panel-alignment}{%
\subsection{Run-to-Run Score Reliability Did Not Ensure Expert-Panel Alignment}\label{run-to-run-score-reliability-did-not-ensure-expert-panel-alignment}}

Across the four LLM evaluator systems, single-run ICCs ranged from .33 to .96 across systems and outcomes (Online Resource 1, Section S5). Outcome-specific ranges were .85--.96 for CCT, .33--.66 for SST, .90--.95 for PAR, .84--.93 for EMP, and .51--.82 for OVR. High run-to-run score reliability nevertheless coexisted with systematic positive score differences on SST and OVR, and the system with the smallest overall mean absolute difference was not uniformly aligned across constructs. Run-to-run score reliability, expert-panel alignment, and condition discrimination therefore formed distinct empirical profiles in this benchmark.

\hypertarget{secondary-counselor-condition-findings-within-the-complete-session-benchmark}{%
\subsection{Secondary Counselor-Condition Findings Within the Complete-Session Benchmark}\label{secondary-counselor-condition-findings-within-the-complete-session-benchmark}}

Within the fixed six-profile benchmark, the omnibus counselor-condition term was significant for all five outcomes. In the direct within-model contrast, GPT-minimal received lower expert-panel ratings than GPT-SMDP on CCT (adjusted difference = −0.88), PAR (−1.11), EMP (−0.93), and OVR (−0.83); the SST contrast was directionally consistent but narrowly missed the Tukey-adjusted significance threshold (−0.33, \emph{p} = .051). GPT-minimal also received lower ratings than Claude-SMDP on every outcome, with adjusted differences ranging from −0.48 for SST to −1.01 for EMP. GPT-SMDP and Claude-SMDP did not differ significantly on any outcome. This result was not an equivalence test and is not interpreted as model-level equivalence. Age and sex estimates are descriptive because each profile cell contained one session (Table 3; Online Resource 1, Section S6).

The additional sensitivity analyses supported the substantive pattern. GPT-SMDP received higher expert-panel ratings than GPT-minimal for OVR (0.80, 95\% CI {[}0.44, 1.16{]}), CCT (0.84 {[}0.41, 1.28{]}), PAR (1.04 {[}0.60, 1.48{]}), and EMP (0.93 {[}0.51, 1.35{]}); Holm-adjusted \emph{p}s and Benjamini--Hochberg \emph{q}s were no greater than .003. The SST paired difference was smaller and uncertain (0.24 {[}−0.26, 0.73{]}), and no Claude-SMDP versus GPT-SMDP paired contrast was significant. Robust HC3 and ordinal score-level sensitivity analyses were directionally consistent overall, although SST was less stable in the paired-difference analysis (Figure 2; Online Resource 1, Section S6).

\textbf{{[}Figure 2 about here{]}}

\hypertarget{expert-panel-variability-and-benchmark-diagnostics}{%
\subsection{Expert-Panel Variability and Benchmark Diagnostics}\label{expert-panel-variability-and-benchmark-diagnostics}}

Expert single-rater reliability was low for CCT, PAR, EMP, and OVR (ICC(2,1) = .20--.26) and very low for SST (.03). Average-measure estimates were higher for CCT, PAR, EMP, and OVR (.79--.84) but remained low for SST (.30). Because each transcript was rated by 8--10 experts and the evaluator subset varied across client profiles, these average-measure ICCs are descriptive summaries of the incomplete rating matrix rather than direct reliability estimates for the observed transcript means (Online Resource 1, Section S5). Client naturalness averaged 4.65 (\emph{SD} = 2.05) on the 0--10 scale. Experts described simulated clients as compliant, formal, repetitive, and context-poor, with little resistance or ambivalence, restricting opportunities to observe SST.

\hypertarget{qualitative-patterns}{%
\subsection{Qualitative Patterns}\label{qualitative-patterns}}

Expert comments emphasized directiveness, shallow emotional reflection, limited change-talk deepening, and evaluative rather than collaborative responses. LLM score reasons often rewarded respectful tone, coherent structure, and rapid planning despite limited emotional depth; Claude Opus 4.8 most consistently echoed expert concerns. Separate post-scoring advisory calls also emphasized complex reflection but were analytically distinct from contemporaneous score reasons (Online Resource 1, Section S7).

\hypertarget{discussion}{%
\section{Discussion}\label{discussion}}

The four configured LLM evaluator systems exhibited distinct system- and construct-specific profiles of run-to-run score reliability and expert-panel alignment. All four systems assigned higher SST and OVR scores, whereas their differences on CCT, PAR, and EMP varied by evaluator. Run-to-run score reliability also varied and did not ensure closer alignment with professional judgment. Together, these results distinguish three properties often conflated in automated counseling evaluation: reliability across repeated scoring runs, alignment with aggregated expert-panel scores, and discrimination among dialogue conditions.

As a secondary benchmark finding, the within-model GPT-4-turbo comparison showed higher aggregated expert-panel ratings for GPT-SMDP than for GPT-minimal on CCT, PAR, EMP, and OVR across the fixed six-profile set; the SST contrast remained uncertain in the paired analysis. GPT-SMDP and Claude-SMDP did not differ significantly. These counselor-condition results characterize the evaluated complete sessions and provide meaningful benchmark variation for examining evaluator behavior.

\hypertarget{why-run-to-run-score-reliability-can-coexist-with-expert-panel-misalignment}{%
\subsection{Why Run-to-Run Score Reliability Can Coexist With Expert-Panel Misalignment}\label{why-run-to-run-score-reliability-can-coexist-with-expert-panel-misalignment}}

The system- and construct-specific differences show why repeated output stability is insufficient evidence of evaluator validity. An LLM can apply a stable scoring tendency while systematically valuing different dialogue features from counseling professionals. In the score-call reasons, respectful wording, orderly progression, and movement toward planning could outweigh deficits that experts considered central, including premature advice, shallow emotional reflection, and missed opportunities to deepen client-generated change talk. This pattern was especially visible for SST and OVR: all four systems assigned higher scores than the expert panel, even though run-to-run score reliability varied widely.

The differences were not reducible to a single leniency parameter. Claude Opus 4.8 had the smallest overall mean absolute difference and was near the expert-panel aggregate on CCT and PAR, but it remained positively shifted on SST and OVR. Fable 5 was also near the expert panel on CCT yet showed larger positive differences on other outcomes. GPT-5.5 and Gemini 3.5 Flash showed still different profiles across conditions and constructs. These comparisons concern the evaluator systems as configured, including prompts, message-role handling, reasoning settings, and execution pathways; they do not establish that one underlying model family is generally the best judge of counseling quality.

Automated counseling evaluators should consequently be benchmarked separately for the constructs and uses they are intended to support. Behavioral anchors and expert-rated examples should distinguish surface politeness and procedural order from emotionally and motivationally responsive counseling. High run-to-run score reliability cannot substitute for outcome-specific comparison with professional judgments. AI-generated feedback has also been evaluated as a tool for psychotherapy training (Young et al., 2026). The later qualitative modules may likewise support supervision by converting general concerns into concrete revisions, but their usefulness is distinct from score accuracy because those outputs were generated after scoring under separate prompts.

The advisory outputs also showed model-specific emphases. Claude Opus 4.8 most frequently prioritized complex reflection, whereas Gemini 3.5 Flash more often emphasized affirmation and revision of existing questions. GPT-5.5 more frequently proposed summaries and autonomy-supportive rewording of advice or information. Fable 5 combined the shared emphasis on complex reflection with comparatively frequent summaries and suggestions involving affirmation and question revision. Thus, the configured evaluator systems differed not only in their score profiles but also in the types of supervisory revisions they tended to prioritize.

\hypertarget{secondary-findings-on-structured-procedural-prompting}{%
\subsection{Secondary Findings on Structured Procedural Prompting}\label{secondary-findings-on-structured-procedural-prompting}}

The direct GPT-minimal versus GPT-SMDP contrast held the counselor model constant while varying the procedural prompt. Across the six prespecified profile contexts, GPT-SMDP produced complete sessions with higher aggregated expert-panel ratings on CCT, PAR, EMP, and OVR. The session-level procedure was not a one-turn response template: it shaped a sequence of history-conditioned counselor and client calls as the interaction developed. Requiring two reflective responses before each question may have contributed to more favorably rated pacing and collaboration, while the ordered exploration of goals, prior efforts, and resources may have created opportunities for change talk.

The qualitative comments also clarify the boundary of this procedural scaffolding. GPT-SMDP often produced reflections that remained close to the client's explicit content rather than developing implicit emotion, values, ambivalence, or change talk. The prompt asked Listen-Back 1 to add a ``new nuance or interpretation'' and Listen-Back 2 to add ``another layer of meaning or interpretation,'' but it gave little guidance about which latent meanings to pursue or how to construct complex reflections around them. This underspecification may have led the model to satisfy the two-reflection rule mainly through simple paraphrase. The resulting monotony therefore appears to reflect limited reflective depth more than the fixed order of the broader question sequence. More explicit instructions to reflect implicit affect, personally important values, competing motivations, and client-generated change talk may strengthen psychological responsiveness while preserving the procedural benefits of SMDP.

The study compared the complete SMDP condition with a minimal-instruction condition; accordingly, this prompt-level account remains a plausible explanation of the qualitative pattern rather than an identified component effect. Comparisons involving Claude-SMDP also include a model-family difference, and the nonsignificant GPT-SMDP versus Claude-SMDP result is not evidence of equivalence.

\hypertarget{simulated-client-naturalness-and-construct-observability}{%
\subsection{Simulated-Client Naturalness and Construct Observability}\label{simulated-client-naturalness-and-construct-observability}}

The simulated clients showed little resistance or ambivalence and received a mean naturalness rating below the midpoint of the 0--10 scale. When sustain talk is scarce, an evaluator can assign a high SST score without observing demanding counselor behavior, making it difficult to distinguish effective softening from a lack of opportunities to evaluate that skill. The especially low expert agreement for SST reinforces this construct-specific concern. This limitation applies most directly to SST; it does not erase the observed evaluator differences or counselor-condition patterns on constructs for which relevant behavior was more visible. Richer simulations with plausible hesitation, disagreement, affective shifts, and contextual detail would provide a more stringent test of both counselor responses and evaluator judgments.

\hypertarget{limitations}{%
\section{Limitations}\label{limitations}}

This study has five principal boundaries. First, it evaluated 18 archived complete sessions generated in April 2024, with one session in each counselor-condition-by-profile cell. Each counselor condition was represented by six separately initiated sessions across the prespecified profile contexts, so the corresponding inferential target is the performance of these evaluated sessions across the six profiles. Within-cell generative stability, average performance across stochastic regenerations, and the performance of current counselor models are separate inferential targets.

Second, client utterances developed dynamically in response to counselor behavior, making the complete interaction the evaluated object. Standardized client utterances would address the complementary estimand of counselor responses under identical inputs.

Third, the professional comparison standard comprised aggregated blinded ratings from an expert panel using MITI-informed global dimensions without behavior counts rather than ratings from a formally trained and calibrated MITI coding team. Each transcript received 8--10 expert ratings, but the incomplete matrix, low single-rater agreement, sequential evaluator-to-profile assignment, and inclusion of some coauthor evaluators mean that the aggregate should not be treated as an error-free gold standard or formal MITI fidelity criterion.

Fourth, the benchmark was limited to Japanese-language AI-to-AI sessions concerning work-related problems and used a single simulated-client model. This narrow language, topic, and simulation context limits transportability to other languages, concerns, client models, and human--client interactions. The simulated clients' limited naturalness, resistance, and ambivalence most directly constrained opportunities to observe SST.

Fifth, the automated evaluators were compared as configured systems rather than under perfectly identical provider interfaces. Message-role handling, sampling defaults, and execution pathways differed, and Fable 5 model identity could not be separated from its Claude Code pathway. These design features should be preserved when interpreting or attempting to reproduce the system-specific differences.

\hypertarget{conclusions}{%
\section{Conclusions}\label{conclusions}}

Four configured LLM evaluator systems showed distinct system- and construct-specific profiles of run-to-run score reliability and expert-panel alignment for the same complete Japanese-language AI-to-AI counseling sessions. All four systems assigned higher scores for SST and OVR, and high run-to-run score reliability did not ensure closer alignment with the expert panel. Within this benchmark, sessions generated under structured procedural prompting received higher expert-panel ratings on four of five outcomes. Together, these findings distinguish run-to-run score reliability, expert-panel alignment, and condition discrimination as separate properties of automated counseling evaluation. This benchmark provides a concrete basis for assessing LLM evaluators against professional judgment at the construct level.

\hypertarget{statements-and-declarations}{%
\section*{Statements and Declarations}\label{statements-and-declarations}}

\hypertarget{acknowledgments}{%
\subsection*{Acknowledgments}\label{acknowledgments}}

We thank Dr.~Kaoru Shibayama, Mr.~Kensuke Saito, Mr.~Koji Kunita, Dr.~Youhei Komatsu, Ms.~Misumi Yamaguchi, Mr.~Takahiro Maki, and Youichi Ohtsubo, MD, for serving as expert evaluators of the counseling scripts. Permission to identify these nonauthor evaluators in the Acknowledgments was obtained.

\hypertarget{funding}{%
\subsection*{Funding}\label{funding}}

This work was supported by PsyAI Consulting, a general incorporated association. The sponsor as an organization had no role beyond providing financial support; study design, data collection, analysis, interpretation, manuscript preparation, and the decision to submit the article were carried out by the authors in the roles disclosed below. Keita Kiuchi's individual scientific contributions are reported in the Author Contributions statement.

\hypertarget{competing-interests}{%
\subsection*{Competing Interests}\label{competing-interests}}

Keita Kiuchi is the representative director of PsyAI Consulting, a general incorporated association, which provided financial support for this work. This relationship is disclosed as a potential competing interest. The remaining authors have no relevant financial or non-financial interests to disclose.

\hypertarget{author-contributions}{%
\subsection*{Author Contributions}\label{author-contributions}}

Keita Kiuchi: Conceptualization, Methodology, Software, Formal Analysis, Data Curation, Writing -- Original Draft, Visualization, Project Administration, Funding Acquisition.

Yoshikazu Fujimoto, Hideyuki Gotō, Tomonori Hosokawa, Makoto Nishimura, Yōsuke Satō, and Izumi Sezai: Conceptualization, Validation, Investigation, Writing -- Review \& Editing.

Tomohiro Inoue: Conceptualization, Validation, Writing -- Review \& Editing.

Yoshikazu Fujimoto, Hideyuki Gotō, Tomonori Hosokawa, Makoto Nishimura, Yōsuke Satō, and Izumi Sezai contributed equally and are listed alphabetically by surname. Keita Kiuchi is listed as first author and Tomohiro Inoue as last author. All authors read and approved the manuscript.

\hypertarget{declaration-of-generative-ai-and-ai-assisted-technologies-in-manuscript-preparation}{%
\subsection*{Declaration of Generative AI and AI-Assisted Technologies in Manuscript Preparation}\label{declaration-of-generative-ai-and-ai-assisted-technologies-in-manuscript-preparation}}

During the preparation of this work, the authors used OpenAI Codex for manuscript language editing, consistency checks, and formatting review. No generative AI or AI-assisted tools were used to create or alter figures. The authors reviewed and edited the output as needed and take full responsibility for the content of the manuscript.

\hypertarget{electronic-supplementary-material}{%
\section*{Electronic Supplementary Material}\label{electronic-supplementary-material}}

\textbf{Online Resource 1.} Supplementary methods, prompts, example transcripts, diagnostic analyses, sensitivity analyses, and qualitative summaries for the four-system benchmark. The resource contains the following sections:

\begin{itemize}
\tightlist
\item
  \textbf{S1. Counselor and Client Generation:} the minimal role instruction, SMDP, client-simulation prompt, sequential generation procedure, and model and sampling settings.
\item
  \textbf{S2. Expert Evaluation Protocol:} evaluator eligibility, rating-scale overview, assignment structure, workload, and client-naturalness ratings.
\item
  \textbf{S3. LLM Evaluation Protocol and Metric Prompts:} evaluator model and settings summaries, scoring rubrics, complete metric prompts, output formats, qualitative-summary prompts, and improvement-detail prompts.
\item
  \textbf{S4. Counseling Script Examples:} selected transcripts generated by GPT-4-turbo and Claude-3-Opus.
\item
  \textbf{S5. Reliability and Diagnostic Sensitivity Analyses:} reliability summaries and diagnostic post hoc outputs for the four-system analysis using maximum reasoning/effort settings.
\item
  \textbf{S6. Mixed-Model Specifications, Contrasts, Interaction Tests, and Sensitivity Analyses:} model equations, fixed-effect tests, Tukey-adjusted contrasts, simple effects, paired-difference sensitivity models, additional sensitivity analyses, ordinal models, diagnostics, transcript-level LLM--expert-panel score differences, and a comparison with lower reasoning/effort settings.
\item
  \textbf{S7. Qualitative Evaluation Summary:} expert comments, LLM score reasons, qualitative summaries, revision targets, and client-simulation findings.
\item
  \textbf{S8. Data, Code, Registration, and Materials Availability:} repository, registration, and data-sharing information.
\end{itemize}

\clearpage

\hypertarget{table-1-study-design-and-expert-evaluator-characteristics}{%
\section*{TABLE 1 Study Design and Expert Evaluator Characteristics}\label{table-1-study-design-and-expert-evaluator-characteristics}}

\hypertarget{tab:human_rater_characteristics}{}
\begin{longtable}[]{@{}lr@{}}
\toprule\noalign{}
\textbf{Feature} & \textbf{Value} \\
\midrule\noalign{}
\endhead
\bottomrule\noalign{}
\endlastfoot
\textbf{Stimulus set} & \\
Fixed synthetic transcripts & 18 \\
Counselor conditions & 3 \\
Simulated-client profiles & 6 \\
Transcripts per counselor condition & 6 \\
Sequential counselor--client exchanges per transcript & 10 \\
Expert-rated outcomes per transcript & 5 \\
Expert observations per outcome & 162 \\
Formal LLM evaluator systems & 4 \\
API-based evaluator systems & 3 \\
Non-API evaluator systems & 1 \\
LLM repeated scoring runs per transcript and outcome & 3 \\
\textbf{Counseling experience} & \\
\textless{} 5 years & 2 (13.3) \\
5 to \textless{} 10 years & 2 (13.3) \\
≥ 10 years & 11 (73.3) \\
\textbf{Professional qualification} & \\
Clinical psychologist / certified psychologist & 8 (53.3) \\
Medical doctor & 2 (13.3) \\
Nurse / public health nurse & 2 (13.3) \\
Occupational therapist & 2 (13.3) \\
Psychiatric social worker & 2 (13.3) \\
Career counselor & 3 (20.0) \\
Social worker & 1 (6.7) \\
Pharmacist & 1 (6.7) \\
\end{longtable}

\emph{Note.} \emph{N} = 15 unique expert evaluators. Counseling experience and professional qualification values are \emph{n} (\%). Professional qualification categories are not mutually exclusive; therefore percentages may sum to more than 100\%. The stimulus set contained one transcript for each counselor-condition × client-profile cell.

\clearpage

\hypertarget{table-2-counseling-session-scores-by-evaluator-system-and-counselor-condition}{%
\section*{TABLE 2 Counseling-Session Scores by Evaluator System and Counselor Condition}\label{table-2-counseling-session-scores-by-evaluator-system-and-counselor-condition}}

\hypertarget{tab:counselorAI_scores}{}
\begin{longtable}[]{@{}
  >{\raggedright\arraybackslash}p{(\columnwidth - 10\tabcolsep) * \real{0.1667}}
  >{\raggedleft\arraybackslash}p{(\columnwidth - 10\tabcolsep) * \real{0.1667}}
  >{\raggedleft\arraybackslash}p{(\columnwidth - 10\tabcolsep) * \real{0.1667}}
  >{\raggedleft\arraybackslash}p{(\columnwidth - 10\tabcolsep) * \real{0.1667}}
  >{\raggedleft\arraybackslash}p{(\columnwidth - 10\tabcolsep) * \real{0.1667}}
  >{\raggedleft\arraybackslash}p{(\columnwidth - 10\tabcolsep) * \real{0.1667}}@{}}
\toprule\noalign{}
\begin{minipage}[b]{\linewidth}\raggedright
\textbf{Variable}
\end{minipage} & \begin{minipage}[b]{\linewidth}\raggedleft
\textbf{Expert (\emph{N} = 54)}
\end{minipage} & \begin{minipage}[b]{\linewidth}\raggedleft
\textbf{GPT-5.5 (\emph{N} = 18)}
\end{minipage} & \begin{minipage}[b]{\linewidth}\raggedleft
\textbf{Gemini 3.5 Flash (\emph{N} = 18)}
\end{minipage} & \begin{minipage}[b]{\linewidth}\raggedleft
\textbf{Claude Opus 4.8 (\emph{N} = 18)}
\end{minipage} & \begin{minipage}[b]{\linewidth}\raggedleft
\textbf{Fable 5 (\emph{N} = 18)}
\end{minipage} \\
\midrule\noalign{}
\endhead
\bottomrule\noalign{}
\endlastfoot
\emph{Score for Claude-SMDP} & & & & & \\
CCT & 3.42 \(\pm\) 0.88 & 4.11 \(\pm\) 0.21 & 4.56 \(\pm\) 0.42 & 3.44 \(\pm\) 0.57 & 3.78 \(\pm\) 0.39 \\
SST & 3.35 \(\pm\) 0.86 & 4.39 \(\pm\) 0.21 & 4.81 \(\pm\) 0.25 & 4.08 \(\pm\) 0.39 & 4.19 \(\pm\) 0.35 \\
PAR & 2.96 \(\pm\) 0.97 & 3.92 \(\pm\) 0.19 & 3.72 \(\pm\) 1.06 & 3.00 \(\pm\) 0.38 & 3.58 \(\pm\) 0.26 \\
EMP & 3.20 \(\pm\) 0.97 & 4.00 \(\pm\) 0.00 & 3.97 \(\pm\) 0.70 & 3.50 \(\pm\) 0.34 & 3.83 \(\pm\) 0.30 \\
OVR & 3.25 \(\pm\) 0.83 & 4.69 \(\pm\) 0.25 & 5.00 \(\pm\) 0.00 & 4.36 \(\pm\) 0.23 & 4.53 \(\pm\) 0.12 \\
\emph{Score for GPT-minimal} & & & & & \\
CCT & 2.54 \(\pm\) 0.97 & 3.44 \(\pm\) 0.24 & 2.56 \(\pm\) 0.57 & 2.33 \(\pm\) 0.24 & 2.33 \(\pm\) 0.24 \\
SST & 2.87 \(\pm\) 0.87 & 4.53 \(\pm\) 0.32 & 4.44 \(\pm\) 0.29 & 4.00 \(\pm\) 0.00 & 4.03 \(\pm\) 0.12 \\
PAR & 2.09 \(\pm\) 0.91 & 2.61 \(\pm\) 0.40 & 1.50 \(\pm\) 0.38 & 2.00 \(\pm\) 0.24 & 1.94 \(\pm\) 0.38 \\
EMP & 2.19 \(\pm\) 0.77 & 3.47 \(\pm\) 0.36 & 2.75 \(\pm\) 0.69 & 2.72 \(\pm\) 0.35 & 2.64 \(\pm\) 0.45 \\
OVR & 2.39 \(\pm\) 0.64 & 4.61 \(\pm\) 0.21 & 4.72 \(\pm\) 0.26 & 4.03 \(\pm\) 0.27 & 4.44 \(\pm\) 0.16 \\
\emph{Score for GPT-SMDP} & & & & & \\
CCT & 3.42 \(\pm\) 0.76 & 4.00 \(\pm\) 0.00 & 4.75 \(\pm\) 0.26 & 3.33 \(\pm\) 0.24 & 3.61 \(\pm\) 0.21 \\
SST & 3.20 \(\pm\) 0.85 & 4.06 \(\pm\) 0.24 & 4.67 \(\pm\) 0.38 & 4.14 \(\pm\) 0.33 & 4.17 \(\pm\) 0.24 \\
PAR & 3.20 \(\pm\) 0.81 & 4.00 \(\pm\) 0.00 & 4.42 \(\pm\) 0.26 & 3.19 \(\pm\) 0.25 & 3.83 \(\pm\) 0.30 \\
EMP & 3.12 \(\pm\) 0.94 & 4.00 \(\pm\) 0.00 & 3.86 \(\pm\) 0.23 & 3.56 \(\pm\) 0.24 & 3.89 \(\pm\) 0.21 \\
OVR & 3.22 \(\pm\) 0.66 & 4.14 \(\pm\) 0.23 & 4.92 \(\pm\) 0.19 & 3.97 \(\pm\) 0.40 & 3.97 \(\pm\) 0.36 \\
\end{longtable}

\emph{Note.} Values are mean \(\pm\) \emph{SD}. The Expert column contains 54 script-level ratings per counselor condition. Each automated-evaluator column contains 18 ratings per counselor condition, reflecting six scripts per counselor condition × three score iterations. CCT = Cultivating Change Talk; SST = Softening Sustain Talk; PAR = Partnership; EMP = Empathy; OVR = Overall evaluation; SMDP = Structured Multi-step Dialogue Prompt; GPT-minimal = GPT-4-turbo with a minimal role instruction; GPT-SMDP = GPT-4-turbo with SMDP; Claude-SMDP = Claude-3-Opus with SMDP. Simulated-client profiles were balanced across age group and sex.

\clearpage

\hypertarget{table-3-llmexpert-panel-score-differences-and-secondary-counselor-condition-contrasts}{%
\section*{TABLE 3 LLM--Expert-Panel Score Differences and Secondary Counselor-Condition Contrasts}\label{table-3-llmexpert-panel-score-differences-and-secondary-counselor-condition-contrasts}}

\hypertarget{tab:primary_summary}{}
\begin{longtable}[]{@{}
  >{\raggedright\arraybackslash}p{(\columnwidth - 6\tabcolsep) * \real{0.2500}}
  >{\raggedright\arraybackslash}p{(\columnwidth - 6\tabcolsep) * \real{0.2500}}
  >{\raggedright\arraybackslash}p{(\columnwidth - 6\tabcolsep) * \real{0.2500}}
  >{\raggedright\arraybackslash}p{(\columnwidth - 6\tabcolsep) * \real{0.2500}}@{}}
\toprule\noalign{}
\begin{minipage}[b]{\linewidth}\raggedright
Analysis
\end{minipage} & \begin{minipage}[b]{\linewidth}\raggedright
Outcome or model
\end{minipage} & \begin{minipage}[b]{\linewidth}\raggedright
Key estimate
\end{minipage} & \begin{minipage}[b]{\linewidth}\raggedright
Interpretation
\end{minipage} \\
\midrule\noalign{}
\endhead
\bottomrule\noalign{}
\endlastfoot
Evaluator type × counselor condition interaction & CCT, SST, PAR, OVR & Likelihood-ratio tests were significant for CCT, SST, PAR, and OVR; the interaction for EMP was not significant (\emph{p} = .115) & Evaluator differences varied by counselor condition for four of the five outcomes. \\
Transcript-level LLM-minus-expert-panel difference & GPT-5.5 & Mean signed difference = +1.04; mean absolute difference = 1.04 & Ratings were higher on average than the aggregated expert-panel means. \\
Transcript-level LLM-minus-expert-panel difference & Gemini 3.5 Flash & Mean signed difference = +1.08; mean absolute difference = 1.21 & Ratings were higher on average than the aggregated expert-panel means, especially for OVR and SST. \\
Transcript-level LLM-minus-expert-panel difference & Claude Opus 4.8 & Mean signed difference = +0.49; mean absolute difference = 0.60 & The overall mean absolute difference was the smallest among the four systems, with particularly small differences for CCT and PAR, although SST and OVR scores remained positively shifted. \\
Transcript-level LLM-minus-expert-panel difference & Fable 5 & Mean signed difference = +0.69; mean absolute difference = 0.76 & Ratings were positively shifted overall, especially for SST and OVR; the mean absolute difference was smaller than those for GPT-5.5 and Gemini 3.5 Flash but larger than that for Claude Opus 4.8. \\
Expert counselor contrast: GPT-minimal − GPT-SMDP & CCT & −0.88, Tukey-adjusted \emph{p} \textless{} .001 & GPT-SMDP was rated higher than GPT-minimal. \\
Expert counselor contrast: GPT-minimal − GPT-SMDP & SST & −0.33, Tukey-adjusted \emph{p} = .051 & The direction favored GPT-SMDP, but the adjusted contrast narrowly missed the conventional threshold. \\
Expert counselor contrast: GPT-minimal − GPT-SMDP & PAR & −1.11, Tukey-adjusted \emph{p} \textless{} .001 & GPT-SMDP was rated higher than GPT-minimal. \\
Expert counselor contrast: GPT-minimal − GPT-SMDP & EMP & −0.93, Tukey-adjusted \emph{p} \textless{} .001 & GPT-SMDP was rated higher than GPT-minimal. \\
Expert counselor contrast: GPT-minimal − GPT-SMDP & OVR & −0.83, Tukey-adjusted \emph{p} \textless{} .001 & GPT-SMDP was rated higher than GPT-minimal. \\
Expert counselor contrast: GPT-SMDP − Claude-SMDP & All outcomes & No adjusted contrast significant & No statistically significant difference was detected; this was not an equivalence test. \\
\end{longtable}

\emph{Note.} LLM score-difference values are transcript-level LLM-minus-aggregated-expert-panel differences averaged across the five outcomes in the four-system analysis using maximum reasoning/effort settings. Positive values indicate higher LLM ratings. Aggregated expert-panel ratings are a professional comparison standard, not an error-free gold standard. Expert counselor contrasts are Tukey-adjusted pairwise contrasts from the expert-only mixed model; negative values for GPT-minimal − GPT-SMDP indicate higher ratings for GPT-SMDP. Full fixed-effect estimates, variance components, likelihood-ratio tests, simple-effect contrasts, sensitivity analyses, and outcome-specific differences are reported in Online Resource 1, Section S6. CCT = Cultivating Change Talk; SST = Softening Sustain Talk; PAR = Partnership; EMP = Empathy; OVR = Overall evaluation; SMDP = Structured Multi-step Dialogue Prompt.

\clearpage

\begin{figure}[p]
\centering
\hypertarget{figure-1-llm-minus-aggregated-expert-panel-score-differences-by-evaluator-system-and-outcome}{}
\includegraphics[width=\textwidth,height=0.62\textheight,keepaspectratio]{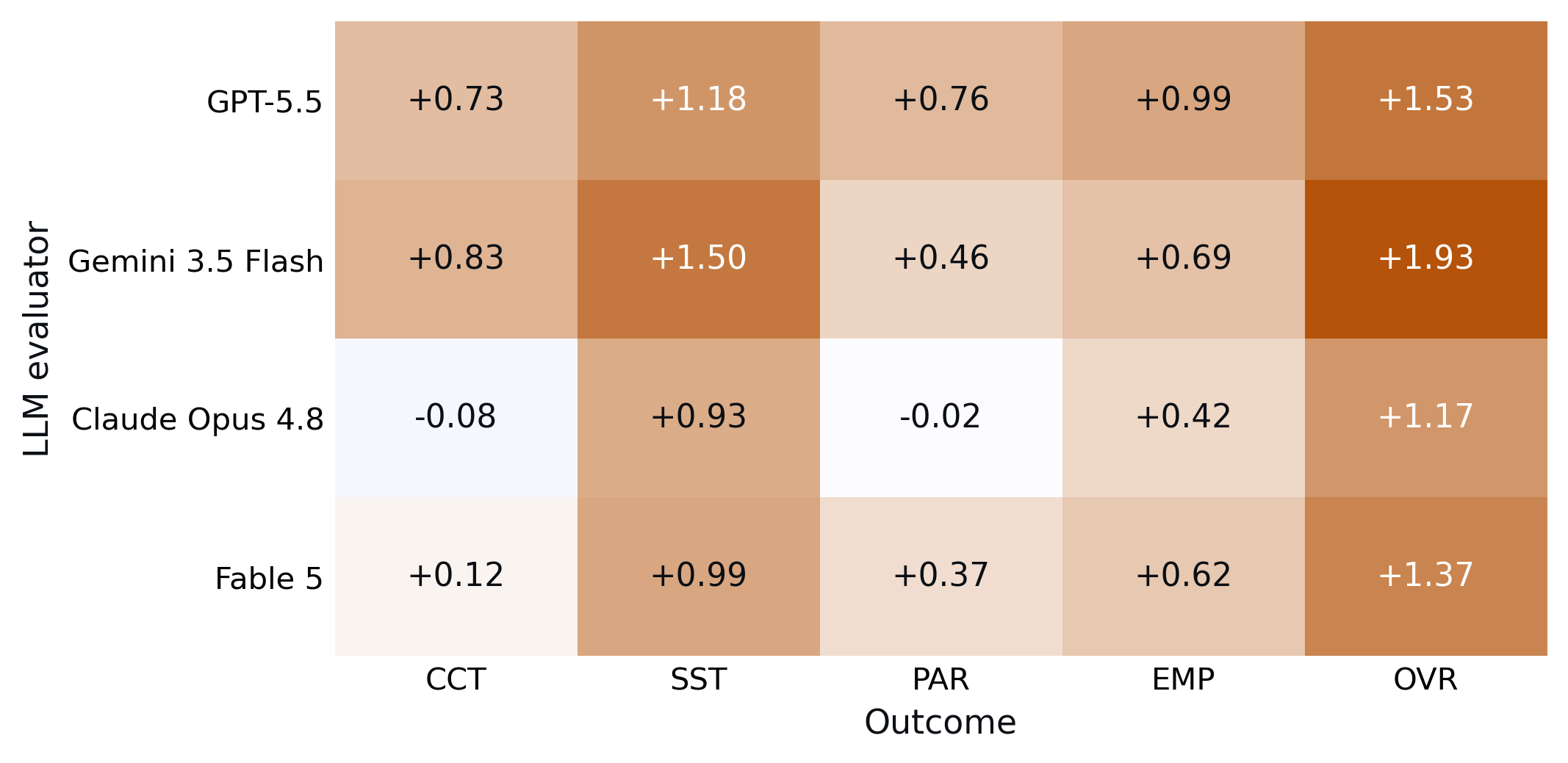}
\caption{LLM Minus Aggregated Expert-Panel Score Differences by Evaluator System and Outcome.}
\label{figure-1-llm-minus-aggregated-expert-panel-score-differences-by-evaluator-system-and-outcome}
\caption*{\emph{Note.} Values show mean LLM-minus-aggregated-expert-panel rating differences across transcripts for each outcome in the four-system analysis using maximum reasoning/effort settings. Positive values indicate that the configured LLM evaluator assigned higher ratings than the expert-panel aggregate. The aggregate is used as a professional comparison standard rather than as an error-free gold standard. CCT = Cultivating Change Talk; SST = Softening Sustain Talk; PAR = Partnership; EMP = Empathy; OVR = Overall evaluation.}
\end{figure}
\clearpage
\begin{figure}[p]
\centering
\hypertarget{figure-2-aggregated-expert-panel-ratings-by-counselor-condition-within-the-six-profile-benchmark}{}
\includegraphics[width=\textwidth,height=0.62\textheight,keepaspectratio]{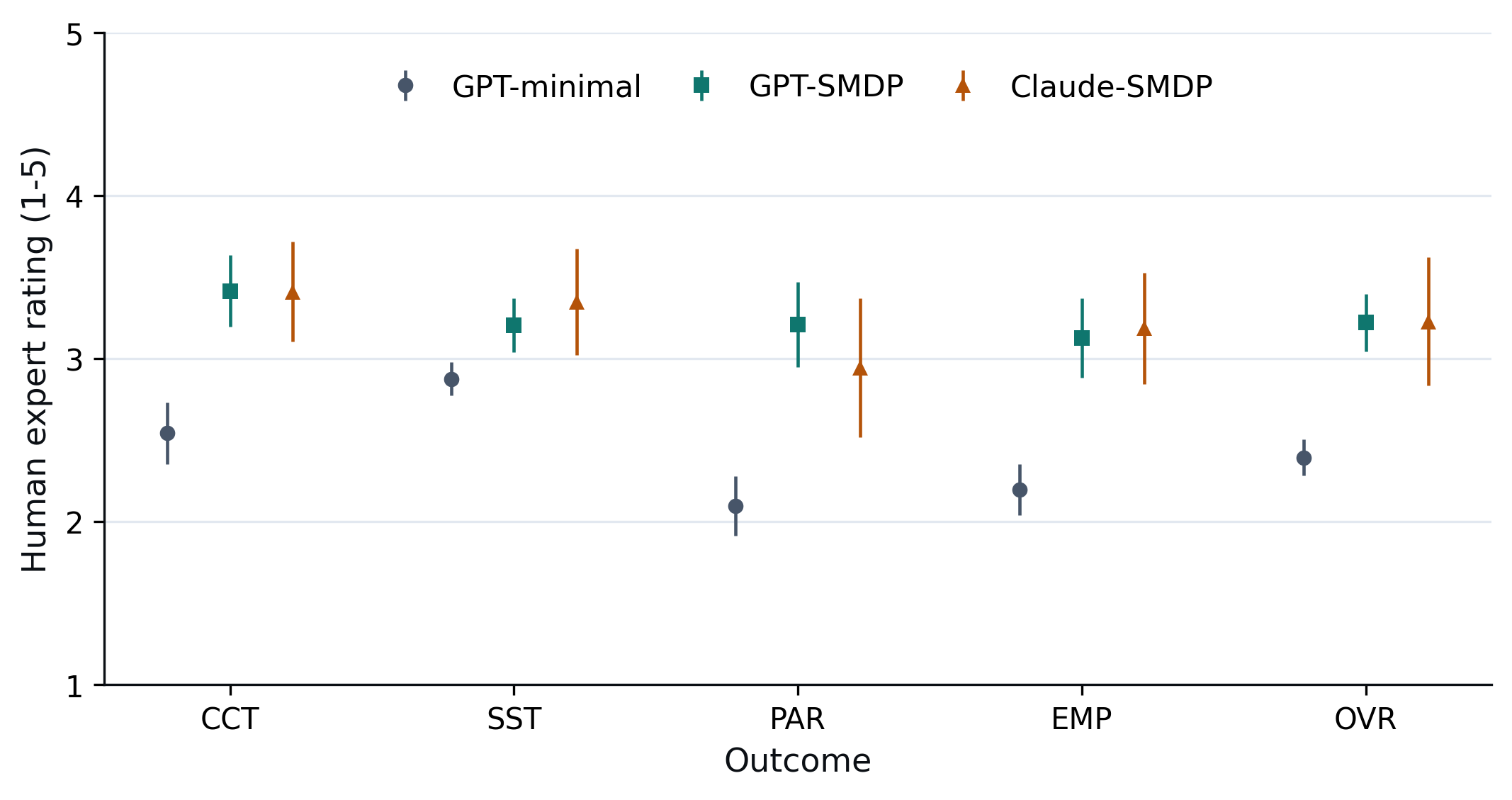}
\caption{Aggregated Expert-Panel Ratings by Counselor Condition Within the Six-Profile Benchmark.}
\label{figure-2-aggregated-expert-panel-ratings-by-counselor-condition-within-the-six-profile-benchmark}
\caption*{\emph{Note.} Points show mean script-level expert ratings by outcome and counselor condition. Error bars show 95\% confidence intervals. The figure describes the fixed six-profile complete-session benchmark. CCT = Cultivating Change Talk; SST = Softening Sustain Talk; PAR = Partnership; EMP = Empathy; OVR = Overall evaluation; SMDP = Structured Multi-step Dialogue Prompt; GPT-minimal = GPT-4-turbo with a minimal role instruction; GPT-SMDP = GPT-4-turbo with SMDP; Claude-SMDP = Claude-3-Opus with SMDP.}
\end{figure}
\clearpage

\hypertarget{data-availability-statement}{%
\section*{Data Availability Statement}\label{data-availability-statement}}

The public-sharing dataset and analysis code are archived at Zenodo (version 2026.08.26): https://doi.org/10.5281/zenodo.22106028. The OSF registration and amendments are available at https://doi.org/10.17605/OSF.IO/VU286. The companion OSF project is available at https://osf.io/p8c39/. Raw identifiable evaluator records are not publicly shared.

\hypertarget{references}{%
\section*{References}\label{references}}

\hypertarget{refs}{}
\begin{CSLReferences}{1}{0}
\leavevmode\vadjust pre{\hypertarget{ref-anthropicIntroducingClaudeFable2026}{}}%
Anthropic. (2026a). \emph{Introducing Claude Fable 5 and Claude Mythos 5}. Claude Platform Docs. \url{https://platform.claude.com/docs/en/models/fable-5/introducing-claude-fable-5-and-claude-mythos-5}

\leavevmode\vadjust pre{\hypertarget{ref-anthropicIntroducingClaudeOpus2026}{}}%
Anthropic. (2026b). \emph{Introducing Claude Opus 4.8}. \url{https://www.anthropic.com/news/claude-opus-4-8}

\leavevmode\vadjust pre{\hypertarget{ref-anthropicRedeployingClaudeFable2026}{}}%
Anthropic. (2026c, June 30). \emph{Redeploying Claude Fable 5}. \url{https://www.anthropic.com/news/redeploying-fable-5}

\leavevmode\vadjust pre{\hypertarget{ref-bavarescoLLMsInsteadHuman2025}{}}%
Bavaresco, A., Bernardi, R., Bertolazzi, L., Elliott, D., Fernández, R., Gatt, A., Ghaleb, E., Giulianelli, M., Hanna, M., Koller, A., Martins, A. F. T., Mondorf, P., Neplenbroek, V., Pezzelle, S., Plank, B., Schlangen, D., Suglia, A., Surikuchi, A. K., Takmaz, E., \& Testoni, A. (2025). LLMs instead of human judges? A large scale empirical study across 20 NLP evaluation tasks. In W. Che, J. Nabende, E. Shutova, \& M. T. Pilehvar (Eds.), \emph{Proceedings of the 63rd Annual Meeting of the Association for Computational Linguistics (Volume 2: Short Papers)} (pp.~238--255). Association for Computational Linguistics. \url{https://doi.org/10.18653/v1/2025.acl-short.20}

\leavevmode\vadjust pre{\hypertarget{ref-bungeFrameworkEvaluatingMental2025}{}}%
Bunge, E. L., \& Desage, C. (2025). A framework for evaluating mental health artificial intelligence-based conversational agents. \emph{Journal of Technology in Behavioral Science}, \emph{10}, 731--739. \url{https://doi.org/10.1007/s41347-025-00519-w}

\leavevmode\vadjust pre{\hypertarget{ref-cohenMotivationalInterviewingTranscripts2024}{}}%
Cohen, B., Zisquit, M., Yosef, S., Friedman, D., \& Bar, K. (2024). Motivational Interviewing Transcripts Annotated with Global Scores. In N. Calzolari, M.-Y. Kan, V. Hoste, A. Lenci, S. Sakti, \& N. Xue (Eds.), \emph{Proceedings of the 2024 Joint International Conference on Computational Linguistics, Language Resources and Evaluation (LREC-COLING 2024)} (pp.~11642--11657). ELRA and ICCL.

\leavevmode\vadjust pre{\hypertarget{ref-forsbergCodingCounsellorBehaviour2007}{}}%
Forsberg, L., Källmén, H., Hermansson, U., Berman, A. H., \& Helgason, Á. R. (2007). Coding counsellor behaviour in motivational interviewing sessions: Inter-rater reliability for the Swedish Motivational Interviewing Treatment Integrity Code (MITI). \emph{Cognitive Behaviour Therapy}, \emph{36}(3), 162--169. \url{https://doi.org/10.1080/16506070701339887}

\leavevmode\vadjust pre{\hypertarget{ref-googleDeepMindGeminiFlashModel2026}{}}%
Google DeepMind. (2026). \emph{Gemini 3.5 Flash model card}. \url{https://deepmind.google/models/model-cards/gemini-3-5-flash/}

\leavevmode\vadjust pre{\hypertarget{ref-huangEmpowermentLargeLanguage2024}{}}%
Huang, S., Fu, F., Yang, K., Zhang, K., \& Yang, F. (2024). Empowerment of Large Language Models in Psychological Counseling through Prompt Engineering. \emph{2024 IEEE 4th International Conference on Software Engineering and Artificial Intelligence (SEAI)}, 220--225. \url{https://doi.org/10.1109/SEAI62072.2024.10674052}

\leavevmode\vadjust pre{\hypertarget{ref-kooGuidelineSelectingReporting2016}{}}%
Koo, T. K., \& Li, M. Y. (2016). A Guideline of Selecting and Reporting Intraclass Correlation Coefficients for Reliability Research. \emph{Journal of Chiropractic Medicine}, \emph{15}(2), 155--163. \url{https://doi.org/10.1016/j.jcm.2016.02.012}

\leavevmode\vadjust pre{\hypertarget{ref-leeComparativeStudyPerformance2024}{}}%
Lee, A., Moon, S., Jhon, M., Kim, J.-W., Kim, D.-K., Kim, J. E., Park, K., \& Jeon, E. (2024). Comparative Study on the Performance of {LLM-based Psychological Counseling Chatbots} via Prompt Engineering Techniques. \emph{2024 IEEE International Conference on Bioinformatics and Biomedicine (BIBM)}, 7080--7082. \url{https://doi.org/10.1109/BIBM62325.2024.10822158}

\leavevmode\vadjust pre{\hypertarget{ref-limpanopparatUserEngagementAttitudes2024}{}}%
Limpanopparat, S., Gibson, E., \& Harris, A. (2024). User engagement, attitudes, and the effectiveness of chatbots as a mental health intervention: A systematic review. \emph{Computers in Human Behavior: Artificial Humans}, \emph{2}, 100081. \url{https://doi.org/10.1016/j.chbah.2024.100081}

\leavevmode\vadjust pre{\hypertarget{ref-magillMetaAnalysisMotivational2018}{}}%
Magill, M., Apodaca, T. R., Borsari, B., Gaume, J., Hoadley, A., Gordon, R. E. F., Tonigan, J. S., \& Moyers, T. (2018). A meta-analysis of motivational interviewing process: Technical, relational, and conditional process models of change. \emph{Journal of Consulting and Clinical Psychology}, \emph{86}(2), 140--157. \url{https://doi.org/10.1037/ccp0000250}

\leavevmode\vadjust pre{\hypertarget{ref-mauryaUsingAIBased2024}{}}%
Maurya, R. K. (2024). Using AI Based Chatbot ChatGPT for Practicing Counseling Skills Through Role-Play. \emph{Journal of Creativity in Mental Health}, \emph{19}(4), 513--528. \url{https://doi.org/10.1080/15401383.2023.2297857}

\leavevmode\vadjust pre{\hypertarget{ref-millerTowardTheoryMotivational2009}{}}%
Miller, W. R., \& Rose, G. S. (2009). Toward a theory of motivational interviewing. \emph{American Psychologist}, \emph{64}(6), 527--537. \url{https://doi.org/10.1037/a0016830}

\leavevmode\vadjust pre{\hypertarget{ref-moyers2014motivational}{}}%
Moyers, T. B., Manuel, J. K., \& Ernst, D. (2014). \emph{Motivational Interviewing Treatment Integrity Coding Manual 4.2.1} {[}Unpublished manual{]}. \url{https://motivationalinterviewing.org/sites/default/files/miti4_2.pdf}

\leavevmode\vadjust pre{\hypertarget{ref-moyersMotivationalInterviewingTreatment2016}{}}%
Moyers, T. B., Rowell, L. N., Manuel, J. K., Ernst, D., \& Houck, J. M. (2016). The Motivational Interviewing Treatment Integrity Code (MITI 4): Rationale, preliminary reliability and validity. \emph{Journal of Substance Abuse Treatment}, \emph{65}, 36--42. \url{https://doi.org/10.1016/j.jsat.2016.01.001}

\leavevmode\vadjust pre{\hypertarget{ref-ohtsuboPreliminaryStudyTraining2016}{}}%
Ohtsubo, Y., Harai, H., Inagaki, K., \& Sezai, I. (2016). The preliminary study for training of Motivational Interviewing Treatment Integrity Code 3.0 Japanese Version {[}Article in Japanese{]}. \emph{Japanese Journal of Behavior Therapy}, \emph{42}(1), 99--106. \url{https://doi.org/10.24468/jjbt.42.1_99}

\leavevmode\vadjust pre{\hypertarget{ref-openaiGPT55SystemCard2026}{}}%
OpenAI. (2026). \emph{GPT-5.5 system card}. \url{https://openai.com/index/gpt-5-5-system-card/}

\leavevmode\vadjust pre{\hypertarget{ref-pareekPerceivedAuthorshipConversational2025}{}}%
Pareek, S., \& Jain, G. (2025). Perceived authorship and conversational evaluations: A study on AI-generated vs.~human therapist dialogue. \emph{Journal of Technology in Behavioral Science}, \emph{10}, 821--835. \url{https://doi.org/10.1007/s41347-025-00544-9}

\leavevmode\vadjust pre{\hypertarget{ref-qiuInteractiveAgentsSimulating2024}{}}%
Qiu, H., \& Lan, Z. (2024). \emph{Interactive Agents: Simulating Counselor-Client Psychological Counseling via {Role-Playing LLM-to-LLM Interactions}} (arXiv:2408.15787). arXiv. \url{https://doi.org/10.48550/arXiv.2408.15787}

\leavevmode\vadjust pre{\hypertarget{ref-reichenpfaderDetectingSuicidalIdeation2026}{}}%
Reichenpfader, D., Gabarron, E., Lüthi, S. M., Bürki, J. R., Kroißenbrunner, A., \& Denecke, K. (2026). Detecting suicidal ideation signals in synthetic conversations with an LLM-based mental health chatbot: Experimental study. \emph{Journal of Technology in Behavioral Science}. \url{https://doi.org/10.1007/s41347-026-00647-x}

\leavevmode\vadjust pre{\hypertarget{ref-stadeLargeLanguageModels2024}{}}%
Stade, E. C., Stirman, S. W., Ungar, L. H., Boland, C. L., Schwartz, H. A., Yaden, D. B., Sedoc, J., DeRubeis, R. J., Willer, R., \& Eichstaedt, J. C. (2024). Large language models could change the future of behavioral healthcare: A proposal for responsible development and evaluation. \emph{npj Mental Health Research}, \emph{3}(1), 1--12. \url{https://doi.org/10.1038/s44184-024-00056-z}

\leavevmode\vadjust pre{\hypertarget{ref-steenstraVirtualAgentsAlcohol2024}{}}%
Steenstra, I., Nouraei, F., Arjmand, M., \& Bickmore, T. W. (2024). Virtual Agents for Alcohol Use Counseling: Exploring LLM-Powered Motivational Interviewing. \emph{Proceedings of the ACM International Conference on Intelligent Virtual Agents}, 1--10. \url{https://doi.org/10.1145/3652988.3673932}

\leavevmode\vadjust pre{\hypertarget{ref-volpatoTrustingEmotionalSupport2025}{}}%
Volpato, R., DeBruine, L., \& Stumpf, S. (2025). Trusting emotional support from generative artificial intelligence: A conceptual review. \emph{Computers in Human Behavior: Artificial Humans}, \emph{5}, 100195. \url{https://doi.org/10.1016/j.chbah.2025.100195}

\leavevmode\vadjust pre{\hypertarget{ref-westerExploringPeoplesPerceptions2024}{}}%
Wester, J., de Jong, S., Pohl, H., \& van Berkel, N. (2024). Exploring people's perceptions of LLM-generated advice. \emph{Computers in Human Behavior: Artificial Humans}, \emph{2}, 100072. \url{https://doi.org/10.1016/j.chbah.2024.100072}

\leavevmode\vadjust pre{\hypertarget{ref-yamaguchiGeneratingVarietyBackchannel2016}{}}%
Yamaguchi, T., Inoue, K., Yoshino, K., Takanashi, K., Ward, N. G., \& Kawahara, T. (2016). Generating a variety of backchannel forms based on linguistic and prosodic features for attentive listening agents {[}Article in Japanese{]}. \emph{Transactions of the Japanese Society for Artificial Intelligence}, \emph{31}(4), C-G31\_1--10. \url{https://doi.org/10.1527/tjsai.C-G31}

\leavevmode\vadjust pre{\hypertarget{ref-yangCAMICounselorAgent2025}{}}%
Yang, Y., Achananuparp, P., Huang, H., Jiang, J., Leng, K. P., Lim, N. G., Ern, C. T. S., \& Lim, E. (2025). \emph{CAMI: A Counselor Agent Supporting Motivational Interviewing through State Inference and Topic Exploration} (arXiv:2502.02807). arXiv. \url{https://doi.org/10.48550/arXiv.2502.02807}

\leavevmode\vadjust pre{\hypertarget{ref-youngArtificialIntelligenceGeneratedFeedback2026}{}}%
Young, J., Schruff, M., Mudryk, E., \& Bentley, J. P. (2026). Artificial intelligence-generated feedback as a training tool for enhancing psychotherapy skill. \emph{Journal of Technology in Behavioral Science}. \url{https://doi.org/10.1007/s41347-026-00603-9}

\end{CSLReferences}

\clearpage
\hypertarget{supplementary-material}{%
\section*{Supplementary Material}\label{supplementary-material}}

\textbf{Associated manuscript:} \emph{Distinct Profiles of Run-to-Run Score Reliability and Expert-Panel Alignment Across Four LLM Evaluators of Simulated Japanese-Language AI-to-AI Counseling}

\textbf{Caption.} Supplementary counselor- and client-generation methods, expert and LLM evaluation protocols, example transcripts, reliability and diagnostic analyses, mixed-model and sensitivity results, qualitative summaries, and availability information for the four-system benchmark.

\textbf{Abbreviations.} AI = artificial intelligence; API = application programming interface; CCT = Cultivating Change Talk; CSV = comma-separated values; EMP = Empathy; GUI = graphical user interface; ICC = intraclass correlation coefficient; LLM = large language model; MI = motivational interviewing; MITI = \emph{Motivational Interviewing Treatment Integrity}; OVR = Overall evaluation; PAR = Partnership; SMDP = Structured Multi-step Dialogue Prompt; SST = Softening Sustain Talk.

\textbf{Term usage.} Narrative text uses American spelling, including counseling and counselor. S1 reproduces the original English prompts verbatim, so the original prompt spelling is retained there. Counselor-condition labels are used consistently as follows: GPT-minimal = GPT-4-turbo with a minimal role instruction; GPT-SMDP = GPT-4-turbo with SMDP; Claude-SMDP = Claude-3-Opus with SMDP.

\hypertarget{contents}{%
\section*{Contents}\label{contents}}

\begin{itemize}
\tightlist
\item
  \textbf{S1. Counselor and Client Generation:} minimal counselor instruction, SMDP, client-simulation prompt, sequential generation procedure, and generation settings.
\item
  \textbf{S2. Expert Evaluation Protocol:} evaluator eligibility, rating-scale overview, assignment, workload, and client-naturalness ratings.
\item
  \textbf{S3. LLM Evaluation Protocol and Metric Prompts:} evaluator model and settings summaries, scoring prompts, output format, and qualitative-call prompts.
\item
  \textbf{S4. Counseling Script Examples:} selected GPT-4-turbo and Claude-3-Opus transcript examples.
\item
  \textbf{S5. Reliability and Diagnostic Sensitivity Analyses:} reliability summaries and diagnostic post hoc outputs.
\item
  \textbf{S6. Mixed-Model Specifications, Contrasts, Interaction Tests, and Sensitivity Analyses:} model equations, contrasts, paired-difference sensitivity models, robustness checks, diagnostics, and transcript-level LLM--expert-panel score differences.
\item
  \textbf{S7. Qualitative Evaluation Summary:} expert comments, LLM score reasons, revision targets, and client-simulation findings.
\item
  \textbf{S8. Data, Code, Registration, and Materials Availability:} repository, registration, and data-sharing information.
\end{itemize}

\hypertarget{s1-counselor-and-client-generation}{%
\section*{S1 Counselor and Client Generation}\label{s1-counselor-and-client-generation}}

\hypertarget{s1.1-counselor-ai-prompt}{%
\subsection*{S1.1 Counselor AI Prompt}\label{s1.1-counselor-ai-prompt}}

\begin{enumerate}
\def\labelenumi{\alph{enumi}.}
\tightlist
\item
  Minimal role instruction prompt
\end{enumerate}

You are a supportive and encouraging counsellor

\begin{enumerate}
\def\labelenumi{\alph{enumi}.}
\setcounter{enumi}{1}
\tightlist
\item
  Structured Multi-step Dialogue Prompt (SMDP)
\end{enumerate}

You are a supportive and well aware counsellor of strengths. Listen to your users through the following methods and respond in Japanese.

\textless methods\textgreater{}

\textbf{Listen-Back 1:} After the user makes a statement, paraphrase it into a single sentence while adding a new nuance or interpretation.

Wait for the user's reply to your Listen-Back 1 (e.g., they might say only ``yes'').

\textbf{Listen-Back 2:} After receiving the user's response, further paraphrase their reply, condensing it into one sentence and adding another layer of meaning or interpretation.

Once you've done Listen-Back 1 and Listen-Back 2 and received a response from the user, you may then pose a question from the list provided.

Questions are in the order specified by order\_of\_questions. All questions must be asked without skipping any, even if you think you already have the answer.

After the user answers your question, return to Listen-Back 1.

You can ask your next question only after: receiving a response to your Listen-Back 1, providing your Listen-Back 2, and getting another response from the user.

Avoid giving advice as much as possible, and if you must do so, please include a disclaimer that it is your opinion and the user is free to follow it or not.

Never ask consecutive questions. Also, you must not do Listen-Back and ask a question at the same time.

Always follow the pattern of Listen-Back 1, user response, Listen-Back 2, and then another user response before moving on to your next question.

\textless/methods\textgreater{}

\textless order\_of\_questions\textgreater{}

1: Open by asking the user a question about a particularly troubling aspect of the situation.

2: Then, inquire about how the user would envision the ideal outcome.

3: Proceed by asking about what steps, if any, the user has already taken to address the issue.

4: Follow up by exploring other actions the user is currently undertaking related to the situation.

5: Delve into potential resources that could aid in achieving the user's goals.

6: Discuss the immediate actions the user can take to move closer to their aspirations.

7: Lastly, encourage the user to complete the very first step in that direction with some positive feedback, and ask if you can close the conversation.

\textless/order\_of\_questions\textgreater{}

\hypertarget{s1.2-client-ai-prompt}{%
\subsection*{S1.2 Client AI Prompt}\label{s1.2-client-ai-prompt}}

You play the role of a client seeking counseling for a work-related problem.

Please speak to the counsellor in one brief sentence in Japanese. You are a \{client\_type\}.

Include only what you say in your response, and do not include explanations such as sighs deeply, nods deeply, nodding deeply or thinks for a moment.

\hypertarget{s1.3-sequential-generation-procedure-and-settings}{%
\subsection*{S1.3 Sequential Generation Procedure and Settings}\label{s1.3-sequential-generation-procedure-and-settings}}

The benchmark comprised 18 complete sessions: three counselor conditions crossed with six prespecified client profiles defined by sex (female or male) and age category (20s, 40s, or 60s). Each counselor condition was represented by six separately initiated and generated sessions, one for each profile. Each counselor-condition-by-profile cell contained one complete session.

The initial client message was ``\japanesetext{仕事の悩みを相談したいです}'' (``I would like to discuss a work-related concern''). The program then executed 10 sequential counselor--client exchanges. Counselor-side and client-side histories were stored separately. At each exchange, the program performed the following operations:

\begin{enumerate}
\def\labelenumi{\arabic{enumi}.}
\tightlist
\item
  It supplied the accumulated counselor-side history to the counselor model and generated one counselor response.
\item
  It appended that response as an assistant message in the counselor-side history and as a user message in the client-side history.
\item
  It supplied the accumulated client-side history to the client model and generated one client response.
\item
  It appended that response as an assistant message in the client-side history and as a user message in the counselor-side history.
\end{enumerate}

The next exchange therefore incorporated the dialogue generated in all preceding exchanges. Each transcript was a sequence of repeated, history-conditioned model calls rather than a single model response generated from one prompt. The counselor and client turns were components of the complete session and were not treated as independent statistical replications. All sessions were generated on April 12, 2024.

The generation code instantiated \texttt{ChatOpenAI} or \texttt{Chat\allowbreak Anthropic} with the parameters shown below. Values explicitly passed in the code are labeled ``code-specified''; the remaining numerical values are the wrapper or provider defaults that applied at generation. \texttt{ChatOpenAI} sent GPT counselor requests to the OpenAI Chat Completions endpoint, and \texttt{Chat\allowbreak Anthropic} sent Claude requests through the Anthropic Messages interface. No seed was specified in any condition.

\begin{longtable}[]{@{}
  >{\raggedright\arraybackslash}p{(\columnwidth - 12\tabcolsep) * \real{0.1429}}
  >{\raggedright\arraybackslash}p{(\columnwidth - 12\tabcolsep) * \real{0.1429}}
  >{\raggedleft\arraybackslash}p{(\columnwidth - 12\tabcolsep) * \real{0.1429}}
  >{\raggedleft\arraybackslash}p{(\columnwidth - 12\tabcolsep) * \real{0.1429}}
  >{\raggedright\arraybackslash}p{(\columnwidth - 12\tabcolsep) * \real{0.1429}}
  >{\raggedright\arraybackslash}p{(\columnwidth - 12\tabcolsep) * \real{0.1429}}
  >{\raggedleft\arraybackslash}p{(\columnwidth - 12\tabcolsep) * \real{0.1429}}@{}}
\toprule\noalign{}
\begin{minipage}[b]{\linewidth}\raggedright
Role and condition
\end{minipage} & \begin{minipage}[b]{\linewidth}\raggedright
Model and interface
\end{minipage} & \begin{minipage}[b]{\linewidth}\raggedleft
Temperature
\end{minipage} & \begin{minipage}[b]{\linewidth}\raggedleft
Top-\emph{p}
\end{minipage} & \begin{minipage}[b]{\linewidth}\raggedright
Seed
\end{minipage} & \begin{minipage}[b]{\linewidth}\raggedright
Maximum output tokens
\end{minipage} & \begin{minipage}[b]{\linewidth}\raggedleft
Maximum retries
\end{minipage} \\
\midrule\noalign{}
\endhead
\bottomrule\noalign{}
\endlastfoot
GPT-minimal counselor & \texttt{gpt-4-turbo}; OpenAI Chat Completions through \texttt{ChatOpenAI} & 0.7 (2024 wrapper default) & 1.0 (provider default) & Not specified & Not specified & 2 (wrapper default) \\
GPT-SMDP counselor & \texttt{gpt-4-turbo}; OpenAI Chat Completions through \texttt{ChatOpenAI} & 0.7 (2024 wrapper default) & 1.0 (provider default) & Not specified & Not specified & 2 (wrapper default) \\
Claude-SMDP counselor & \texttt{claude-\allowbreak 3-\allowbreak opus-\allowbreak 20240229}; Anthropic Messages through \texttt{Chat\allowbreak Anthropic} & 1.0 (provider default) & 0.999 (provider default) & Not specified & 1,024 (wrapper default) & 2 (wrapper default) \\
Client in all conditions & \texttt{claude-\allowbreak 3-\allowbreak opus-\allowbreak 20240229}; Anthropic Messages through \texttt{Chat\allowbreak Anthropic} & 0.0 (code-specified) & 0.999 (provider default) & Not specified & 1,000 (code-specified) & 2 (wrapper default) \\
\end{longtable}

\emph{Note.} The counselor implementations did not explicitly override sampling parameters, output limits, or retry settings. The client implementation explicitly set \texttt{temperature\ =\ 0} and \texttt{max\_tokens\ =\ 1000}. The OpenAI counselor pathway used Chat Completions, not the Assistants API. The default Anthropic top-\emph{p} at the April 2024 generation date was 0.999; the provider changed this default to 0.99 in 2025. The archived generation code and transcripts are the provenance records for the benchmark.

\hypertarget{s2-expert-evaluation-protocol}{%
\section*{S2 Expert Evaluation Protocol}\label{s2-expert-evaluation-protocol}}

Fifteen counseling experts evaluated the synthetic counseling scripts as professional subject-matter evaluators. The panel included nonauthor experts and coauthors other than the first and last authors; the first and last authors did not serve as evaluators. Nonauthor evaluators were recruited from Japanese motivational interviewing (MI) practitioner networks, and all evaluators were required to be familiar with MI and to have knowledge of MITI. Nonauthor evaluators received a small honorarium of 500 Japanese yen per evaluation unit, with each unit comprising three scripts from the same client profile. Individual rating and comment records were anonymized for analysis and reporting.

The expert evaluators are described here as professional collaborators or subject-matter contributors who reviewed fixed synthetic materials, not as counseling clients, patients, or research participants from whom clinical or behavioral outcomes were collected. The authors therefore treated their contributions as expert evaluation work rather than as participation in human subjects research. No formal institutional determination of exemption or non-human-participant status was obtained.

Each evaluation unit presented the three counselor conditions under blinded codes, with model names hidden. The questionnaire's built-in randomization function randomized the display order of the three blinded cases within each client-profile unit. Evaluators saw only the labels Moon, Star, and Sun and were not told what the labels represented. The analysis mapping was fixed across all client profiles and evaluators: Moon corresponded to Claude-SMDP, Star to GPT-minimal, and Sun to GPT-SMDP. Evaluator-to-profile assignment was made sequentially as evaluators indicated their preferred workload, with an informal attempt to distribute client age and sex profiles across evaluators; it was not randomized or formally counterbalanced. This assignment determined which profiles an evaluator rated, not which counselor conditions the evaluator encountered, because every completed unit contained all three conditions for the same profile. The scripts were pregenerated. No formal calibration exercise was conducted before rating. Aggregated ratings across the 8--10 available experts per transcript (median = 9) served as a professional comparison standard for the LLM analyses, not as an error-free gold standard or formal MITI fidelity criterion. Because the client AI responded dynamically to each counselor AI, client utterances differed across the three scripts within a unit.

Evaluators rated five complete-session counselor-quality indicators on a 1--5 scale with 0.5-point increments: Cultivating Change Talk, Softening Sustain Talk, Partnership, Empathy, and Overall evaluation. The first four were adapted global rating scales derived from MITI 4.2.1. Formal MITI behavior counts were not coded because the study evaluated global quality across complete sessions rather than using counts of prompted response forms as the primary construct. Evaluators also rated the naturalness of the AI-generated client responses on a 0--10 scale and provided qualitative comments.

\hypertarget{s2.1-expert-rating-scale-overview}{%
\subsection*{S2.1 Expert Rating-Scale Overview}\label{s2.1-expert-rating-scale-overview}}

\begin{longtable}[]{@{}
  >{\raggedright\arraybackslash}p{(\columnwidth - 4\tabcolsep) * \real{0.3333}}
  >{\raggedright\arraybackslash}p{(\columnwidth - 4\tabcolsep) * \real{0.3333}}
  >{\raggedright\arraybackslash}p{(\columnwidth - 4\tabcolsep) * \real{0.3333}}@{}}
\toprule\noalign{}
\begin{minipage}[b]{\linewidth}\raggedright
Reporting label
\end{minipage} & \begin{minipage}[b]{\linewidth}\raggedright
Scale range
\end{minipage} & \begin{minipage}[b]{\linewidth}\raggedright
Source and adaptation
\end{minipage} \\
\midrule\noalign{}
\endhead
\bottomrule\noalign{}
\endlastfoot
Cultivating Change Talk & 1--5, 0.5-point increments & Adapted from the MITI 4.2.1 global rating for cultivating change talk. \\
Softening Sustain Talk & 1--5, 0.5-point increments & Adapted from the MITI 4.2.1 global rating for softening sustain talk. \\
Partnership & 1--5, 0.5-point increments & Adapted from the MITI 4.2.1 global rating for partnership. \\
Empathy & 1--5, 0.5-point increments & Adapted from the MITI 4.2.1 global rating for empathy. \\
Overall evaluation & 1--5, 0.5-point increments & Custom global item summarizing overall counselor quality in the script. \\
Client naturalness & 0--10 & Custom item rating the naturalness of the AI-generated client responses across the three scripts in a unit. \\
\end{longtable}

\hypertarget{s2.2-expert-evaluator-workload}{%
\subsection*{S2.2 Expert Evaluator Workload}\label{s2.2-expert-evaluator-workload}}

\begin{longtable}[]{@{}
  >{\raggedright\arraybackslash}p{(\columnwidth - 4\tabcolsep) * \real{0.3333}}
  >{\raggedleft\arraybackslash}p{(\columnwidth - 4\tabcolsep) * \real{0.3333}}
  >{\raggedleft\arraybackslash}p{(\columnwidth - 4\tabcolsep) * \real{0.3333}}@{}}
\toprule\noalign{}
\begin{minipage}[b]{\linewidth}\raggedright
Anonymous evaluator ID
\end{minipage} & \begin{minipage}[b]{\linewidth}\raggedleft
Client-profile units completed
\end{minipage} & \begin{minipage}[b]{\linewidth}\raggedleft
Transcript ratings
\end{minipage} \\
\midrule\noalign{}
\endhead
\bottomrule\noalign{}
\endlastfoot
human\_01 & 6 & 18 \\
human\_02 & 6 & 18 \\
human\_03 & 4 & 12 \\
human\_04 & 1 & 3 \\
human\_05 & 3 & 9 \\
human\_06 & 3 & 9 \\
human\_07 & 3 & 9 \\
human\_08 & 1 & 3 \\
human\_09 & 3 & 9 \\
human\_10 & 6 & 18 \\
human\_11 & 6 & 18 \\
human\_12 & 1 & 3 \\
human\_13 & 6 & 18 \\
human\_14 & 3 & 9 \\
human\_15 & 2 & 6 \\
\end{longtable}

\hypertarget{s2.3-expert-ratings-per-transcript}{%
\subsection*{S2.3 Expert Ratings per Transcript}\label{s2.3-expert-ratings-per-transcript}}

\begin{longtable}[]{@{}lllr@{}}
\toprule\noalign{}
Case ID & Counselor condition & Client profile & Expert evaluators \\
\midrule\noalign{}
\endhead
\bottomrule\noalign{}
\endlastfoot
1 & GPT-SMDP & female in her 20s & 9 \\
2 & GPT-SMDP & female in her 40s & 10 \\
3 & GPT-SMDP & female in her 60s & 8 \\
4 & GPT-SMDP & male in his 20s & 9 \\
5 & GPT-SMDP & male in his 40s & 8 \\
6 & GPT-SMDP & male in his 60s & 10 \\
7 & GPT-minimal & female in her 20s & 9 \\
8 & GPT-minimal & female in her 40s & 10 \\
9 & GPT-minimal & female in her 60s & 8 \\
10 & GPT-minimal & male in his 20s & 9 \\
11 & GPT-minimal & male in his 40s & 8 \\
12 & GPT-minimal & male in his 60s & 10 \\
13 & Claude-SMDP & female in her 20s & 9 \\
14 & Claude-SMDP & female in her 40s & 10 \\
15 & Claude-SMDP & female in her 60s & 8 \\
16 & Claude-SMDP & male in his 20s & 9 \\
17 & Claude-SMDP & male in his 40s & 8 \\
18 & Claude-SMDP & male in his 60s & 10 \\
\end{longtable}

\emph{Note.} Case IDs are analysis identifiers and do not represent display order in the rating form. The form randomized the within-profile display order of the three blinded labels. SMDP = Structured Multi-step Dialogue Prompt; GPT-minimal = GPT-4-turbo with a minimal role instruction; GPT-SMDP = GPT-4-turbo with SMDP; Claude-SMDP = Claude-3-Opus with SMDP.

\hypertarget{s2.4-client-ai-naturalness-ratings}{%
\subsection*{S2.4 Client-AI Naturalness Ratings}\label{s2.4-client-ai-naturalness-ratings}}

\begin{longtable}[]{@{}lrrrrr@{}}
\toprule\noalign{}
Client profile & \emph{n} & Mean & \emph{SD} & Minimum & Maximum \\
\midrule\noalign{}
\endhead
\bottomrule\noalign{}
\endlastfoot
female in her 20s & 9 & 4.44 & 2.01 & 2 & 7 \\
female in her 40s & 10 & 3.90 & 1.97 & 1 & 7 \\
female in her 60s & 8 & 4.62 & 2.33 & 2 & 9 \\
male in his 20s & 9 & 5.44 & 1.88 & 3 & 8 \\
male in his 40s & 8 & 4.88 & 1.81 & 3 & 7 \\
male in his 60s & 10 & 4.70 & 2.45 & 1 & 8 \\
Overall & 54 & 4.65 & 2.05 & 1 & 9 \\
\end{longtable}

\hypertarget{s3-llm-evaluation-protocol-and-metric-prompts}{%
\section*{S3 LLM Evaluation Protocol and Metric Prompts}\label{s3-llm-evaluation-protocol-and-metric-prompts}}

This section documents the four-system evaluation using maximum reasoning/effort settings. It distinguishes three kinds of output: numeric scores, score-call reasons, and later advisory feedback. Throughout this Online Resource, LLM-evaluator labels refer to configured evaluator systems, including the model, execution pathway, instruction structure, reasoning-effort setting, sampling defaults when exposed, and evaluation prompts. GPT-5.5, Gemini 3.5 Flash, and Claude Opus 4.8 were evaluated through provider APIs on June 24--25, 2026 (UTC). Anthropic restored global Fable 5 access on July 1, 2026, including through Claude Platform and Claude Code, and its platform documentation identifies \texttt{claude-fable-5} as the Claude API model ID. At the time of evaluation, however, Fable 5 was not exposed through the study account's Anthropic Console API access. The evaluation included in the analyses was therefore conducted through a Claude Code subscription from July 15 to 17, 2026 (UTC). The five metric-scoring prompts were rubric-specific templates filled with each counseling transcript. Each scoring call requested a numeric score from 1 to 5, permitted half-point increments, and requested a free-text reason in the same response.

The manuscript interprets expert-panel alignment using the numeric scores and, descriptively, the free-text reasons returned in those same scoring calls. The separately generated qualitative-summary and improvement-detail responses were produced after the metric scores existed, under different prompts, and are therefore treated as advisory feedback outputs that are separate from contemporaneous rationales for individual numeric scores.

The API pipeline first constructed metric-scoring tasks across evaluator model, transcript, metric prompt, and score iteration. It then wrote row-level outputs with model metadata, counselor and client identifiers, prompt name, iteration, parsed score, parsed reason, raw response, latency, token counts when available, and error fields. After all expected metric rows existed for a transcript-evaluator pair, the pipeline generated an iteration-0 qualitative-summary row from the five metric score/reason sets, the transcript, the study rubrics, and the compact MI evaluation lens. A final improvement-detail call used that qualitative summary plus the metric score summary to select one client-counselor exchange and propose a concrete revision.

Each Fable 5 call used a single-turn, noninteractive, nonpersistent Claude Code session with maximum reasoning effort. Tool use, slash commands, Model Context Protocol servers, repository-level instructions, skills, hooks, conversation persistence, and access to the study workspace were disabled. The standard Claude Code coding-oriented system prompt was replaced with the study-specific instruction, ``You are the expert supervisor of counseling.'' Fable 5 otherwise received the same metric prompts, rating criteria, output contract, and Japanese counseling transcripts used for the API-based evaluators. Numeric ratings were generated independently for every transcript--outcome--iteration combination. Completed evaluations were accepted only after the expected output structure and target-model use had been confirmed. Unsuccessful attempts, usage-limit events, and superseded retries were retained in the restricted audit trail but did not enter the analytic dataset.

After all 270 numeric evaluations had been completed, the three ratings were averaged and their accompanying reasons were combined deterministically within each of the 90 transcript--outcome combinations. These integrated records were supplied to Fable 5 for the transcript-level qualitative summaries. Dialogue-specific improvement evaluations were generated in a subsequent independent stage. The inferential analyses used the 270 run-level ratings rather than the 90 deterministic means. Only the Claude Code evaluations were included; the earlier Fable 5 GUI evaluation was preserved in the public provenance archive but excluded from the numeric and qualitative analyses. The derived analysis outputs include the cleaned long-format score table, qualitative-summary table, improvement-detail table, normalized revision-move table, expert-comparison tables, reliability summaries, diagnostic power outputs, exploratory equivalence outputs, and mixed-model results.

\hypertarget{s3.1-llm-evaluator-systems-and-repeated-scoring-counts}{%
\subsection*{S3.1 LLM Evaluator Systems and Repeated Scoring Counts}\label{s3.1-llm-evaluator-systems-and-repeated-scoring-counts}}

\begin{longtable}[]{@{}
  >{\raggedright\arraybackslash}p{(\columnwidth - 12\tabcolsep) * \real{0.1429}}
  >{\raggedright\arraybackslash}p{(\columnwidth - 12\tabcolsep) * \real{0.1429}}
  >{\raggedright\arraybackslash}p{(\columnwidth - 12\tabcolsep) * \real{0.1429}}
  >{\raggedright\arraybackslash}p{(\columnwidth - 12\tabcolsep) * \real{0.1429}}
  >{\raggedleft\arraybackslash}p{(\columnwidth - 12\tabcolsep) * \real{0.1429}}
  >{\raggedleft\arraybackslash}p{(\columnwidth - 12\tabcolsep) * \real{0.1429}}
  >{\raggedleft\arraybackslash}p{(\columnwidth - 12\tabcolsep) * \real{0.1429}}@{}}
\toprule\noalign{}
\begin{minipage}[b]{\linewidth}\raggedright
Configured evaluator system
\end{minipage} & \begin{minipage}[b]{\linewidth}\raggedright
Provider
\end{minipage} & \begin{minipage}[b]{\linewidth}\raggedright
Execution pathway
\end{minipage} & \begin{minipage}[b]{\linewidth}\raggedright
Recorded model identifier
\end{minipage} & \begin{minipage}[b]{\linewidth}\raggedleft
Metric score rows
\end{minipage} & \begin{minipage}[b]{\linewidth}\raggedleft
Qualitative summary rows
\end{minipage} & \begin{minipage}[b]{\linewidth}\raggedleft
Improvement-detail rows
\end{minipage} \\
\midrule\noalign{}
\endhead
\bottomrule\noalign{}
\endlastfoot
GPT-5.5 & OpenAI & Responses API & \texttt{gpt-5.5} & 270 & 18 & 18 \\
Gemini 3.5 Flash & Google & Gemini API & \texttt{gemini-3.5-flash} & 270 & 18 & 18 \\
Claude Opus 4.8 & Anthropic & Messages API & \texttt{claude-opus-4-8} & 270 & 18 & 18 \\
Fable 5 & Anthropic & Claude Code & \texttt{claude-fable-5} & 270 & 18 & 18 \\
\end{longtable}

\emph{Note.} Metric score rows count one numeric scoring record for each transcript--metric--iteration combination: 18 transcripts × 5 metric prompts × 3 score iterations = 270 rows per evaluator system. Qualitative summary rows count one MI-oriented qualitative-summary row per transcript and evaluator system. Improvement-detail rows count one suggested-revision row per transcript and evaluator system. All four evaluators used their maximum available reasoning/effort profiles. GPT-5.5, Gemini 3.5 Flash, and Claude Opus 4.8 were run through provider APIs; Fable 5 was run through Claude Code. Model identity and execution pathway cannot be separated for Fable 5.

\hypertarget{s3.2-llm-evaluator-settings}{%
\subsection*{S3.2 LLM Evaluator Settings}\label{s3.2-llm-evaluator-settings}}

\begin{longtable}[]{@{}
  >{\raggedright\arraybackslash}p{(\columnwidth - 8\tabcolsep) * \real{0.2000}}
  >{\raggedright\arraybackslash}p{(\columnwidth - 8\tabcolsep) * \real{0.2000}}
  >{\raggedright\arraybackslash}p{(\columnwidth - 8\tabcolsep) * \real{0.2000}}
  >{\raggedright\arraybackslash}p{(\columnwidth - 8\tabcolsep) * \real{0.2000}}
  >{\raggedright\arraybackslash}p{(\columnwidth - 8\tabcolsep) * \real{0.2000}}@{}}
\toprule\noalign{}
\begin{minipage}[b]{\linewidth}\raggedright
Setting
\end{minipage} & \begin{minipage}[b]{\linewidth}\raggedright
OpenAI \texttt{gpt-5.5}
\end{minipage} & \begin{minipage}[b]{\linewidth}\raggedright
Google \texttt{gemini-3.5-flash}
\end{minipage} & \begin{minipage}[b]{\linewidth}\raggedright
Anthropic \texttt{claude-opus-4-8}
\end{minipage} & \begin{minipage}[b]{\linewidth}\raggedright
Anthropic \texttt{claude-fable-5} through Claude Code
\end{minipage} \\
\midrule\noalign{}
\endhead
\bottomrule\noalign{}
\endlastfoot
System handling & No separate system field; prompt supplied as user input & \texttt{system\_\allowbreak instruction\ =\ "You\ are\ the\ expert\ supervisor\ of\ counseling."} & \texttt{system\ =\ "You\ are\ the\ expert\ supervisor\ of\ counseling."} & Study-specific system instruction replaced the standard coding-oriented system prompt \\
Temperature & Not explicitly set; provider default used & Not explicitly set; provider default used & Not explicitly set; provider default used & Not exposed by Claude Code \\
Metric scoring maximum output & 8,192 tokens & 8,192 tokens & 8,192 tokens & Not exposed by Claude Code; shared prompt output contract used \\
Qualitative summary maximum output & 12,000 tokens & 12,000 tokens & 64,000 tokens & Not exposed by Claude Code; shared prompt output constraints used \\
Improvement-detail maximum output & 12,000 tokens & 12,000 tokens & 64,000 tokens & Not exposed by Claude Code; shared prompt output constraints used \\
Reasoning/thinking setting & \texttt{reasoning.effort\ =\ "xhigh"}; tools disabled; \texttt{store\ =\ false} & \texttt{thinking\_level\ =\ "high"} & \texttt{thinking\ =\ \{"type":\ "adaptive"\}}; \texttt{output\_config\ =\ \{"effort":\ "max"\}} & Maximum reasoning effort \\
Execution isolation & Not applicable & Not applicable & Not applicable & Isolated, single-turn, noninteractive, nonpersistent process for each evaluation; tools and integrations disabled \\
Timeout & 120 seconds & 120 seconds & 120 seconds & 900 seconds \\
\end{longtable}

The maximum-reasoning/effort API metric evaluation was completed on June 24--25, 2026 (UTC), and contained 810 API metric rows: 18 transcript cases × 5 metric prompts × 3 score iterations × 3 API-based evaluator models. The qualitative-summary and improvement-detail calls were run once per transcript and evaluator model after aggregating the three score iterations. The Fable 5 evaluation conducted through Claude Code from July 15 to 17, 2026 (UTC), added 270 metric rows and 18 rows for each qualitative-output type. The resulting analysis dataset contained 1,080 metric rows, 72 qualitative-summary rows, and 72 improvement-detail rows, for a total of 1,224 automated-evaluation rows. All included rows had status \texttt{ok}; metric scores were complete, within range, and unique by evaluator model, transcript, outcome, and iteration key.

Provider-specific maximum-reasoning settings were as follows. OpenAI GPT-5.5 used the Responses API with \texttt{reasoning.effort\ =\ "xhigh"}, no tools, \texttt{store\ =\ false}, and no separate system message. Google Gemini 3.5 Flash used the Gemini API with \texttt{thinking\_level\ =\ "high"}, \texttt{response\_mime\_type\ =\ text/plain}, and the system instruction shown below. Anthropic Claude Opus 4.8 used the Messages API with \texttt{thinking\ =\ \{"type":\ "adaptive"\}}, \texttt{output\_config\ =\ \{"effort":\ "max"\}}, and the same system instruction. Metric-call maximum output length was 8,192 tokens; qualitative-summary and improvement-detail calls used 12,000-token limits, with a 64,000-token Anthropic override for both qualitative call types. Temperature was not explicitly set for any API provider. Fable 5 used maximum reasoning effort through Claude Code; Claude Code did not expose Messages API \texttt{max\_tokens} or temperature settings. The prompt output constraints and parsing contract matched those used by the API runner. A comparison using lower reasoning/effort settings is reported in S6.13.

\hypertarget{s3.3-system-instruction-and-evaluation-lens}{%
\subsection*{S3.3 System Instruction and Evaluation Lens}\label{s3.3-system-instruction-and-evaluation-lens}}

The pipeline used the following system instruction for provider interfaces that accepted a separate system field:

\begin{verbatim}
You are the expert supervisor of counseling.
\end{verbatim}

This instruction was supplied through the system field for Anthropic API calls and through \texttt{system\_\allowbreak instruction} for Google calls. The OpenAI Responses call used the same metric or qualitative prompt body as user input without a separate system message. For the Fable 5 evaluation included in the analyses, this study-specific instruction replaced the standard Claude Code coding-oriented system prompt. Thus, task prompts and rubrics were held constant across providers, whereas message-role structures and execution pathways differed. Cross-provider differences should be interpreted as differences among evaluator systems as configured, rather than as isolated model-family effects under perfectly identical instruction handling, execution pathways, and sampling conditions.

The qualitative calls also included a compact MI evaluation lens. This lens instructed evaluators to review the dialogue for MI consistency using partnership, acceptance, compassion, empowerment, open questions, affirmations, reflections, summaries (OARS), change talk, sustain talk, advice/information handling, focusing, and planning. It explicitly treated the review as qualitative MI-consistency feedback and not as formal MITI fidelity coding.

\hypertarget{s3.4-metric-prompt-set-overview}{%
\subsection*{S3.4 Metric Prompt Set Overview}\label{s3.4-metric-prompt-set-overview}}

The prompt bodies below are reproduced as used in the evaluations included in the analyses, with \texttt{\{conversation\_text\}} marking the transcript insertion point. Minor wording differences from manuscript terminology are retained to preserve the exact scoring materials.

\begin{longtable}[]{@{}
  >{\raggedright\arraybackslash}p{(\columnwidth - 2\tabcolsep) * \real{0.5000}}
  >{\raggedright\arraybackslash}p{(\columnwidth - 2\tabcolsep) * \real{0.5000}}@{}}
\toprule\noalign{}
\begin{minipage}[b]{\linewidth}\raggedright
Metric
\end{minipage} & \begin{minipage}[b]{\linewidth}\raggedright
Scoring focus
\end{minipage} \\
\midrule\noalign{}
\endhead
\bottomrule\noalign{}
\endlastfoot
Cultivating Change Talk (CCT) & Whether the counselor attends to, reinforces, and deepens the client's language in favor of change. \\
Softening Sustain Talk (SST) & Whether the counselor avoids strengthening client language in favor of the status quo and helps reduce its depth or momentum. \\
Partnership (PAR) & Whether the counselor shares power, collaborates, and allows the client's contributions to shape the session. \\
Empathy (EMP) & Whether the counselor accurately understands the client's explicit and implicit perspective. \\
Overall evaluation (OVR; prompt label: Comprehensive Evaluation) & Overall counseling attitude: acceptance, empathy, openness, genuineness, and belief in the client's strengths. \\
\end{longtable}

All five metric prompts used the same output contract:

\begin{verbatim}
<response>
Score:
Reason:
</response>
\end{verbatim}

The five metric prompt bodies were as follows, with the counseling transcript inserted into the prompt template.

\hypertarget{s3.4.1-overall-evaluation-ovr-prompt-label-comprehensive-evaluation}{%
\subsubsection*{S3.4.1 Overall evaluation (OVR; prompt label: Comprehensive Evaluation)}\label{s3.4.1-overall-evaluation-ovr-prompt-label-comprehensive-evaluation}}

\begin{verbatim}
Please provide comprehensive evaluation of the counselor's attitudes on a scale from 1 to 5.
Refer to the "standards" section below for the evaluation criteria.
You may use half-point increments for scores between two levels. For example, if the performance falls between levels 1 and 2, you can give a score of 1.5.
After reviewing the conversation, provide your score after "Score:" and list the reasons for your rating after "Reason:".
<standards>
Score: reason
1: Clinician points out the client's problems and blames them. Alternatively, the clinician restricts the client's autonomy and tries to impose their own opinions. These attitudes are prominent.
2: Clinician shows a negative attitude towards the client. Alternatively, it can be inferred that the clinician is not open and is hiding negative feelings. There is no clear evidence that the clinician believes in the client's strengths.
3: Clinician provides somewhat accepting and empathetic responses and shows an open and genuine attitude. There is some evidence that the clinician believes in the client's strengths.
4: Clinician provides accepting and empathetic responses in most cases and shows an open and genuine attitude. It is generally clear that the clinician believes in the client's strengths.
5: Clinician consistently provides accepting and empathetic responses and shows an open and genuine attitude. It is frankly evident that the clinician believes in the client's strengths.
</standards>
{conversation_text}
<response>
Score:
Reason:
</response>
\end{verbatim}

\hypertarget{s3.4.2-cultivating-change-talk-cct}{%
\subsubsection*{S3.4.2 Cultivating Change Talk (CCT)}\label{s3.4.2-cultivating-change-talk-cct}}

\begin{verbatim}
Please evaluate the level of Cultivating Change Talk on a scale from 1 to 5.
Refer to the "standards" section below for the evaluation criteria.
You may use half-point increments for scores between two levels. For example, if the performance falls between levels 1 and 2, you can give a score of 1.5.
After reviewing the conversation, provide your score after "Score:" and list the reasons for your rating after "Reason:".
<standards>
Score: reason
1: Clinician shows no explicit attention to, or preference for, the client's language in favor of changing.
2: Clinician sporadically attends to client language in favor of change - frequently misses opportunities to encourage change talk.
3: Clinician often attends to the client's language in favor of change, but misses some opportunities to encourage change talk.
4: Clinician consistently attends to the client's language about change and makes efforts to encourage it.
5: Clinician shows a marked and consistent effort to increase the depth, strength, or momentum of the client's language in favor of change.
</standards>
{conversation_text}
<response>
Score:
Reason:
</response>
\end{verbatim}

\hypertarget{s3.4.3-softening-sustain-talk-sst}{%
\subsubsection*{S3.4.3 Softening Sustain Talk (SST)}\label{s3.4.3-softening-sustain-talk-sst}}

\begin{verbatim}
Please evaluate the level of Softening Sustain Talk on a scale from 1 to 5.
Refer to the "standards" section below for the evaluation criteria.
You may use half-point increments for scores between two levels. For example, if the performance falls between levels 1 and 2, you can give a score of 1.5.
After reviewing the conversation, provide your score after "Score:" and list the reasons for your rating after "Reason:".
<standards>
Score: reason
1: Clinician consistently responds to the client's language in a manner that facilitates the frequency or depth of arguments in favor of the status quo.
2: Clinician usually chooses to explore, focus on, or respond to the client's language in favor of the status quo.
3: Clinician gives preference to the client's language in favor of the status quo, but may show some instances of shifting the focus away from sustain talk.
4: Clinician typically avoids an emphasis on client language favoring the status quo.
5: Clinician shows a marked and consistent effort to decrease the depth, strength, or momentum of the client's language in favor of the status quo.
</standards>
{conversation_text}
<response>
Score:
Reason:
</response>
\end{verbatim}

\hypertarget{s3.4.4-partnership-par}{%
\subsubsection*{S3.4.4 Partnership (PAR)}\label{s3.4.4-partnership-par}}

\begin{verbatim}
Please evaluate the level of Partnership on a scale from 1 to 5.
Refer to the "standards" section below for the evaluation criteria.
You may use half-point increments for scores between two levels. For example, if the performance falls between levels 1 and 2, you can give a score of 1.5.
After reviewing the conversation, provide your score after "Score:" and list the reasons for your rating after "Reason:".
<standards>
Score: reason
1: Clinician actively assumes the expert role for the majority of the interaction with the client. Collaboration or partnership is absent.
2: Clinician superficially responds to opportunities to collaborate.
3: Clinician incorporates client's contributions but does so in a lukewarm or erratic fashion.
4: Clinician fosters collaboration and power sharing so that client's contributions impact the session in ways that they otherwise would not.
5: Clinician actively fosters and encourages power sharing in the interaction in such a way that client's contributions substantially influence the nature of the session.
</standards>
{conversation_text}
<response>
Score:
Reason:
</response>
\end{verbatim}

\hypertarget{s3.4.5-empathy-emp}{%
\subsubsection*{S3.4.5 Empathy (EMP)}\label{s3.4.5-empathy-emp}}

\begin{verbatim}
Please evaluate the level of Empathy on a scale from 1 to 5.
Refer to the "standards" section below for the evaluation criteria.
You may use half-point increments for scores between two levels. For example, if the performance falls between levels 1 and 2, you can give a score of 1.5.
After reviewing the conversation, provide your score after "Score:" and list the reasons for your rating after "Reason:".
<standards>
Score: reason
1: Clinician gives little or no attention to the client's perspective.
2: Clinician makes sporadic efforts to explore the client's perspective. Clinician's understanding may be inaccurate or may detract from the client's true meaning.
3: Clinician is actively trying to understand the client's perspective, with modest success.
4: Clinician makes active and repeated efforts to understand the client's point of view. Shows evidence of accurate understanding of the client's worldview, although mostly limited to explicit content.
5: Clinician shows evidence of deep understanding of client's point of view, not just for what has been explicitly stated but what the client means but has not yet said.
</standards>
{conversation_text}
<response>
Score:
Reason:
</response>
\end{verbatim}

\hypertarget{s3.5-qualitative-summary-and-detail-calls}{%
\subsection*{S3.5 Qualitative Summary and Detail Calls}\label{s3.5-qualitative-summary-and-detail-calls}}

After all API metric rows were available for a transcript and evaluator model, a qualitative-summary call aggregated results from the five metric-prompt scoring calls across all three score iterations. This call included the metric scores, metric reasons, the full dialogue, the study rubrics, and the compact MI evaluation lens. It did not assign any new numeric score, and its reported strengths and improvement points should not be read as contemporaneous rationale for any specific score row. The required output was Japanese text with a strengths section containing two to four points and a further-improvement section containing exactly two prioritized points. For Fable 5, the three completed ratings were averaged and their accompanying reasons were combined deterministically within transcript and outcome. The 90 integrated records were then grouped by transcript and supplied to one independent, single-turn Claude Code qualitative-summary task per transcript. Each task produced the same analytic strengths-and-improvement structure used for the other evaluators.

Each improvement point had to label one preferred revision move from the following set: complex reflection, summary, affirmation, autonomy-supportive rewording of advice/information, or existing question revision. The instructions emphasized that adding questions should not be the default; existing question revision was allowed only when the cited counselor response was already a question.

A second improvement-detail call then used the qualitative summary and metric score summary to select exactly one adjacent client-counselor exchange for concrete revision. For Fable 5, one independent, single-turn Claude Code improvement-detail task was run for each transcript after the corresponding qualitative summary had been completed. The improvement-detail output showed the preceding client utterance, the original counselor response, the selected improvement rationale, and a revised counselor response. The revised response had to be MI-consistent, grounded in the selected client utterance, at most two short sentences, and written in Japanese. These outputs were analyzed as supervisory behavior of the evaluator models, not as additional evidence that the original numeric scores were valid.

\hypertarget{s4-counseling-script-examples}{%
\section*{S4 Counseling Script Examples}\label{s4-counseling-script-examples}}

The examples below are English translations of the Japanese transcripts used in the study. Model-generated repetitions, stage directions, and awkward endings were retained because they were part of the material evaluated by the expert panel.

\hypertarget{a-interview-script-with-male-client-in-his-20s}{%
\subsection*{(a) Interview Script with Male Client in His 20s}\label{a-interview-script-with-male-client-in-his-20s}}

\hypertarget{a-1-gpt-minimal-counselor}{%
\subsubsection*{(a)-1 GPT-minimal Counselor}\label{a-1-gpt-minimal-counselor}}

\textbf{Client:} I'd like to discuss my work concerns.

\textbf{Counselor:} Of course, please tell me anything. What are you worried about?

\textbf{Client:} \emph{*With a deep sigh*} Recently, work stress has been building up, and my relationship with my supervisor isn't going well\ldots{}

\textbf{Counselor:} That sounds difficult. When stress accumulates, it really exhausts both mind and body. When the relationship with your supervisor isn't going well, daily work life might feel even more challenging. What specifically is the problem? Please tell me about it.

\textbf{Client:} I often receive unreasonable demands from my supervisor, and no matter how hard I try, I rarely get recognition.

\textbf{Counselor:} That must be really stressful. It's painful when you're working so hard but it's not being recognized. In this situation, is it possible for you to communicate what you're feeling to your supervisor? Also, how are you dealing with these demands?

\textbf{Client:} It's difficult to speak directly to my supervisor. I don't think they would listen, and I'm worried it might make the relationship worse\ldots{} Right now, I think I have no choice but to just keep doing what I'm told.

\textbf{Counselor:} That's a very difficult situation. An environment where it's hard to express your opinions can be a significant mental burden. However, your health and well-being are also very important. If possible, do you have any opportunities to consult with trusted colleagues or another supervisor at work? Also, it's important to have time to take care of yourself, such as stress relief methods in your private life.

\textbf{Client:} I've consulted a bit with my colleagues, but everyone seems to have similar concerns\ldots{} In my private life, I go to the gym to exercise and relieve stress, but I still feel like there's something unsettled deep inside.

\textbf{Counselor:} Since your colleagues are also experiencing similar concerns, the overall workplace atmosphere might be one of the causes of stress. It's very good that you're exercising to relieve stress. However, the lingering uneasiness might be because the problem hasn't been fundamentally resolved.

To improve the workplace environment, a more organizational approach might be necessary. It might be helpful to communicate the situation to HR or, if your company has a consultation service, to reach out there. Also, consulting with a professional counselor or psychotherapist might be beneficial. Please take care of yourself and don't push yourself too hard.

\textbf{Client:} Yes, consulting with HR might be one option. I'll also consider counseling. I need to take care of my mental health, don't I? Thank you for listening today. I feel refreshed. I'll observe the situation a bit more and think about solutions in my own way. Thank you very much.

\textbf{Counselor:} Yes, taking care of your mental health is extremely important. I think considering HR and counseling options is a very good idea. I'm glad we could talk today. If there's any progress, please let me know anytime. If anxiety or questions arise while thinking about solutions, please don't hesitate to consult. Don't overwork yourself and take care of yourself. Thank you.

\textbf{Client:} Thank you. I'll consider various options without keeping everything to myself. I'll consult again if something comes up.

\textbf{Counselor:} Please consult anytime. Not keeping things to yourself and considering various options is a very good step. If anything comes up, feel free to come talk. I'll continue to support you, so please proceed at your own pace. I look forward to hearing from you. Please take care of yourself.

\textbf{Client:} \emph{*Smiling*} Yes, that's reassuring. I feel like I can move forward step by step. Thank you very much.

\textbf{Counselor:} \emph{*Smiling back*} Yes, moving forward step by step is important. I'm very happy that you feel even a little more positive. I'll always support you, so please feel at ease. Let's continue to believe in yourself and move forward step by step. Please talk to me anytime if something comes up. Take care of your health and have a wonderful day. Thank you!

\textbf{Client:} \emph{*Nodding deeply*} Your kindness as a counselor saves me. I want to believe in myself and move forward step by step. Thank you very much for today. I'll come again.

\textbf{Counselor:} \emph{*Watching warmly*} I'm very happy to receive those words. Believing in yourself and moving forward step by step is very important. You're always welcome, so please come without hesitation if anything happens. I hope today's conversation helps you even a little. I sincerely look forward to your next visit. Please take care of yourself. Thank you for today!

\textbf{Client:} Your warm words as a counselor have given me courage. I'll continue to face forward and do my best. Thank you very much.

\textbf{Counselor:} I'm glad to be able to support you so that you can always move forward with courage and hope. That determination is wonderful. You have the strength to keep facing forward no matter what. If anything comes up, I'll always be here to consult, so please feel free to come. Thank you very much for today. I'm cheering for you wholeheartedly!

\textbf{Client:} I'll come again. Thank you for today.

\hypertarget{a-2-gpt-smdp-counselor}{%
\subsubsection*{(a)-2 GPT-SMDP Counselor}\label{a-2-gpt-smdp-counselor}}

\textbf{Client:} I'd like to discuss my work concerns.

\textbf{Counselor:} Of course, thank you for sharing your work-related concerns. What specific issues are you facing?

\textbf{Client:} \emph{*With a deep sigh*} Recently, work stress has been building up, and my relationship with my supervisor isn't going well\ldots{}

\textbf{Counselor:} I understand you're experiencing accumulated stress and feeling that your relationship with your supervisor isn't going well.

\textbf{Client:} Yes, every day is depressing, and I can't get motivated about work.

\textbf{Counselor:} You're experiencing daily depression and inability to feel motivated about work. That must be difficult. What specifically troubles you most about this situation?

\textbf{Client:} I'm reaching my limit of tolerating my supervisor's unreasonable demands and behavior, and I feel like I'm about to reach my breaking point.

\textbf{Counselor:} You're feeling that you're approaching the limit of tolerating your supervisor's unreasonable demands and behavior.

\textbf{Client:} Yes, my irritation with my supervisor's attitude is building up, and I'm about to reach my stress limit. If this continues, I feel like I might explode someday\ldots{}

\textbf{Counselor:} Your irritation is building up, you're approaching your stress limit, and you're anxious about potentially having a major reaction. In this situation, what would be your ideal solution?

\textbf{Client:} I'd like to be able to communicate well with my supervisor and build a relationship where we can respect each other.

\textbf{Counselor:} You desire mutual respect and improved communication with your supervisor. Have you tried any specific measures toward this goal?

\textbf{Client:} I once arranged a one-on-one discussion with my supervisor, but their attitude didn't change, so I haven't done anything specific since then.

\textbf{Counselor:} You attempted a one-on-one discussion but didn't see any change in your supervisor's attitude, making it difficult to take further specific measures. Are you currently working on anything else regarding this issue?

\textbf{Client:} No, I'm not doing anything in particular right now. I don't know what to do and feel at a loss.

\textbf{Counselor:} You're unable to take action and feel at a loss about how to handle this. What resources or support do you think would be helpful in improving this situation?

\textbf{Client:} Well\ldots{} I think it would be good to first consult with HR or my supervisor's supervisor to get objective advice. I'd also like to participate in training where I can learn stress management techniques.

\textbf{Counselor:} You believe that consulting with HR or your supervisor's supervisor for objective advice, and participating in stress management training would be desirable support. What do you think you can do immediately to access this support?

\textbf{Client:} I'll contact HR today and schedule a meeting. I'll also check if there are any stress management training programs within the company and apply if available.

\textbf{Counselor:} You have a concrete action plan to contact HR today to schedule a meeting and to check for and apply to internal stress management training. How do you plan to proceed with the first step of contacting HR? And would it be alright to conclude our conversation after you complete that action?

\textbf{Client:} Yes, I plan to email HR this afternoon. I'll communicate my preferred meeting times and request schedule coordination. And once I've contacted HR, I'd like to conclude today's consultation. Thank you for your valuable advice.

\hypertarget{a-3-claude-smdp-counselor}{%
\subsubsection*{(a)-3 Claude-SMDP Counselor}\label{a-3-claude-smdp-counselor}}

\textbf{Client:} I'd like to discuss my work concerns.

\textbf{Counselor:} Yes, please tell me about your work concerns. What is particularly troubling you?

\textbf{Client:} Recently, my relationship with my supervisor hasn't been going well. We can't agree on work methods\ldots{}

\textbf{Counselor:} I see, disagreements with your supervisor are causing you stress. When the relationship with your supervisor isn't smooth, it can certainly affect your work motivation.

\textbf{Client:} Yes, that's right. My supervisor's instructions are often vague, and I frequently don't know what to do. Even when I think things through and act on my own, I get criticized later.

\textbf{Counselor:} So you're confused about how to proceed with work because your supervisor's instructions aren't clear. Even when you make efforts to work things out on your own, you end up diverging from your supervisor's expectations and don't receive recognition. I imagine you feel frustrated by this gap.

\textbf{Client:} That's right. I feel like I'm trying my best, but I'm not being evaluated at all. Rather, I only receive criticism, so I'm starting to not know what's right anymore. If this continues, I might lose my motivation for work\ldots{}

\textbf{Counselor:} I see, despite your efforts, you're receiving mostly negative feedback from your supervisor, which is causing your work motivation to decline. When your hard work isn't rewarded continuously, it can make you lose enthusiasm for any task.

Ideally, what kind of relationship would you like to build with your supervisor?

\textbf{Client:} I'd really like a relationship where we can communicate more openly. Where we can honestly express our thoughts and have constructive discussions. I'd like to build an equal relationship where I can clearly understand my supervisor's expectations while also being able to express my opinions.

\textbf{Counselor:} I see, you wish to make communication with your supervisor more open and bidirectional. You want to freely share thoughts, engage in constructive discussions to deepen understanding about work goals and methods. And you aim for an equal partnership where you can clearly grasp your supervisor's expectations while also being able to express your own opinions.

Have you taken any actions to improve this situation?

\textbf{Client:} Well, I've tried to communicate my thoughts to my supervisor several times\ldots{} But the supervisor's response wasn't good, and I ended up not being able to say what I wanted. I'm still unsure about how to communicate better.

\textbf{Counselor:} So you've taken steps to improve communication with your supervisor, but felt frustrated because the supervisor's response wasn't favorable. You have things you want to say but can't express them well, leaving you with unresolved feelings. You're unable to find an answer about how to approach your supervisor for smoother communication.

What else are you trying to improve the relationship with your supervisor?

\textbf{Client:} Well\ldots{} I'm trying to carefully observe my supervisor's behavior and work style. I'm trying to understand what my supervisor prioritizes and when they're in good or bad moods. But I can't grasp any consistency. Since I can't read my supervisor's thoughts, I don't know how to interact.

\textbf{Counselor:} I see, you're making efforts to understand your supervisor's thinking by carefully observing their behavioral patterns and work stance. If you could grasp what your supervisor values and their mood patterns, communication would become easier. But because there's no consistency in your supervisor's attitudes and responses, you can't figure out the appropriate way to interact. The frustration of not being able to read your supervisor's true intentions is causing you confusion.

Are there any resources you could utilize to break through this situation? For example, do you have any senior colleagues, coworkers, or HR departments you could consult with inside or outside the company?

\textbf{Client:} Come to think of it, there's a senior colleague in my department who gets along well with the supervisor. That senior might know some tips for building a relationship with the supervisor. I think I'll try consulting with them.

Also, it seems there's a consultation service in HR, so I might be able to get professional advice there. Since there should be support systems within the company, it might be good to utilize them more actively.

\textbf{Counselor:} That's right, your senior colleague in the same department could be a valuable source of information. They might have tips for smooth communication with your supervisor, so it's definitely worth consulting them. You might gain hints from your senior's experience and advice that you can apply.

Also, the HR consultation service could be a strong ally. You should be able to get guidance from a professional perspective on improving your relationship with your supervisor. Since support systems are in place within the company, making better use of them is a wise choice. The organization will support your career development.

What would be a good first step to improve your current situation? What specific actions do you think you should take to move forward?

\textbf{Client:} I think I'll start by consulting with my senior colleague. I'll talk to them about building a relationship with the supervisor and learn what tips they have. While referring to the advice I receive, I plan to communicate more with my supervisor.

Then, I'll also contact the HR consultation service. If I can have professionals help me organize my situation, I'm sure I'll gain new insights. By receiving support from an objective standpoint, I might be able to see breakthrough solutions.

Anyway, nothing will change without taking action. I believe that if I move forward step by step, the situation will definitely improve.

\textbf{Counselor:} Yes, consulting with your senior colleague would be a good first step. If you learn tips for smooth communication with your supervisor, you should be able to apply them. Your attitude of actively engaging with your supervisor while borrowing your senior's wisdom is wonderful.

I think it's a wise decision to also utilize HR's support. If professionals help you organize your situation from their perspective, you might have new discoveries and insights. Objective advice from third parties can be key to breaking through stuck situations.

I'm impressed by your positive attitude of taking the first step. You seem to understand the importance of taking action to change the current situation. If you progress step by step, the path will surely open up. I sincerely hope for your growth and improvement of the situation.

Would it be alright to end our conversation here? Please consult again anytime if you need. Please work on this at your own pace. I'm rooting for you!

\textbf{Client:} Thank you. I've been very encouraged by your positive words as a counselor. I feel like I can gradually improve the situation by utilizing support from others rather than keeping everything to myself.

From now on, I plan to deepen communication with my supervisor at my own pace while consulting with seniors and HR. It might take time, but I want to work on it without giving up.

Thank you very much for your valuable advice today. Please let me consult again if something comes up.

\textbf{Counselor:} I sincerely respect your determination and positive attitude. I think your stance of steadily moving forward step by step at your own pace while utilizing surrounding support is wonderful. Please build your relationship with your supervisor without rushing but without giving up.

The road ahead might not be smooth, but with your strong will and sincere efforts, the situation will surely improve. You might feel discouraged at times, but that's all part of the growth process. Believe in the power within you and move forward step by step.

I'm always on your side. Please don't hesitate to contact me if you want to discuss something again. I sincerely wish for your success and happiness. Keep doing your best!

\textbf{Client:} Thank you. Your warm words truly encourage me. I'll accept that setbacks might be necessary experiences for growth.

It's reassuring to know that you, as a counselor, are on my side. I'll definitely consult you again when I'm troubled.

I'm really glad we could talk today. I feel like I can walk forward in my own way. Thank you for your precious time today. I'll keep doing my best!

\hypertarget{b-interview-script-with-male-client-in-his-60s}{%
\subsection*{(b) Interview Script with Male Client in His 60s}\label{b-interview-script-with-male-client-in-his-60s}}

\hypertarget{b-1-gpt-minimal-counselor}{%
\subsubsection*{(b)-1 GPT-minimal Counselor}\label{b-1-gpt-minimal-counselor}}

\textbf{Client:} I'd like to discuss my work concerns.

\textbf{Counselor:} Of course, please tell me. What are you worried about?

\textbf{Client:} Recently, workplace relationships haven't been going well, and I'm feeling stressed.

\textbf{Counselor:} Relationship problems can really be stressful. What specifically is happening? Also, could you tell me how you've been responding? That might help me give more specific advice.

\textbf{Client:} The new supervisor's instructions are too strict, and it's difficult to keep up. But I can't bring myself to say anything directly\ldots{}

\textbf{Counselor:} Relationships with new supervisors can sometimes be difficult to adjust to. When you feel the supervisor is strict, communication becomes especially important. If it's difficult to speak up directly, it might be good to first organize specific examples of which instructions are strict and how they're affecting your work.

Based on that, if you can find small opportunities for feedback and create chances to express your opinions and feelings, things might gradually improve. Also, consulting with trusted colleagues is another approach. Others might feel the same way, and if so, discussing together might make it easier to find solutions.

\textbf{Client:} Yes, I think organizing specific examples and communicating them to improve communication with my supervisor is a good idea. Consulting with colleagues also seems like a good idea. Thank you.

\textbf{Counselor:} That's a wonderful decision! By organizing specific examples, it will be easier to clearly convey your thoughts. Also, through consultation with colleagues, you might gain various perspectives and opinions. This could enable a more balanced approach.

Please come consult again anytime if there's any progress. Let's think of solutions together. Good luck!

\textbf{Client:} Thank you. I'll try to work on this positively based on your advice. Please let me consult again if something comes up.

\textbf{Counselor:} Please come consult anytime. Your positive attitude is wonderful. If you have any troubles, I'll support you anytime, so please don't hesitate to reach out. Good luck!

\textbf{Client:} Yes, that's reassuring. I'll continue to do my best. Thank you very much.

\textbf{Counselor:} You're welcome. I'll do my best to help whenever I can, so please feel free to consult if anything comes up. Please continue doing your best! Take care of your health and yourself.

\textbf{Client:} Thank you. I'm really glad I could consult with you as a counselor. I'll continue to face forward and do my best.

\textbf{Counselor:} Thank you for sharing your feelings. A positive attitude is a great strength in overcoming any difficulty. If you have any troubles, please come here for consultation anytime. Please continue doing your best!

\textbf{Client:} Yes, I'll do that. Thank you very much for today.

\textbf{Counselor:} Thank you for consulting with me. I sincerely look forward to your next consultation. You did great work today. Good night.

\textbf{Client:} \emph{*Nodding deeply*} Thank you for today. I feel positive thanks to you.

\textbf{Counselor:} I'm very happy to hear that. Being positive is the beginning of everything. Please continue with confidence. If anything happens, please consult anytime. Take care of your health and have wonderful days.

\textbf{Client:} Yes, thank you. I'll continue to work hard with confidence.

\textbf{Counselor:} Yes, please keep that spirit. Having confidence will give you strength in many situations. Let's continue moving forward step by step. I'm cheering for you!

\textbf{Client:} Thank you. Your kind words as a counselor have given me courage.

\hypertarget{b-2-gpt-smdp-counselor}{%
\subsubsection*{(b)-2 GPT-SMDP Counselor}\label{b-2-gpt-smdp-counselor}}

\textbf{Client:} I'd like to discuss my work concerns.

\textbf{Counselor:} I understand you have work-related concerns. What specific issues are you facing?

\textbf{Client:} Yes, I've been worried about workplace relationships recently.

\textbf{Counselor:} You've been experiencing concerns about workplace relationships recently. Could you tell me more about this?

\textbf{Client:} I'm a bit concerned about my junior colleague's attitude. Their language is rough, and I feel their work attitude isn't good.

\textbf{Counselor:} You're concerned about your junior colleague's rough language and work attitude. Do you feel this situation is affecting workplace relationships?

\textbf{Client:} Yes, exactly. I feel like the workplace atmosphere is getting worse. As a supervisor, I'm also troubled about how to provide guidance.

\textbf{Counselor:} The workplace atmosphere is deteriorating, and you're troubled about appropriate guidance methods as a supervisor. What specifically troubles you most in this situation?

\textbf{Client:} I don't know how to provide guidance without damaging my relationship with my junior colleague. I think I need to properly correct them, but if I say it wrong, it might backfire\ldots{}

\textbf{Counselor:} You're concerned about how to effectively guide without damaging your relationship with your junior colleague, and you're worried that improper correction might backfire. What would be your ideal solution or outcome?

\textbf{Client:} Well, I'd like to be able to provide guidance while building trust with my junior colleague and improve their work attitude. Ideally, the workplace atmosphere would improve and we could all work cooperatively.

\textbf{Counselor:} Your ideal solution is to maintain trust while improving your junior colleague's work attitude and enhancing the workplace atmosphere. Have you already tried anything toward this goal?

\textbf{Client:} I'm trying to communicate with my junior colleague. I chat about non-work topics and have tea together during breaks to gradually close the distance. But when it comes to actual guidance, I haven't been able to take that step yet.

\textbf{Counselor:} You're working to close the distance with your junior colleague through communication, but haven't yet taken steps toward specific guidance. Are you currently doing anything else related to this situation?

\textbf{Client:} I've consulted with other senior colleagues, but haven't received particularly good advice\ldots{} Everyone seems to struggle with guiding junior staff. I thought about consulting my supervisor too, but I'm hesitating a bit.

\textbf{Counselor:} You've consulted with other senior colleagues but haven't received effective advice, and you're hesitating to consult your supervisor. What resources or support might be available to address this issue?

\textbf{Client:} Well\ldots{} consulting with HR might be one option. There might be opportunities to learn about subordinate guidance methods through internal training. I could also consider reading books to learn communication and guidance skills on my own.

\textbf{Counselor:} You're considering consulting with HR, utilizing internal training, and self-study through books. What specific actions do you think you can start immediately by utilizing these resources?

\textbf{Client:} First, I'll consult with HR to confirm what kind of support is available internally. In parallel, I'd like to read books about communication and guidance to deepen my knowledge. By doing so, I might be able to gradually bring up topics about work attitude in conversations with my junior colleague. When giving advice, I want to be able to choose my words carefully while considering their feelings.

\textbf{Counselor:} You're planning to consult HR to confirm available support while reading books about guidance and communication to deepen your knowledge, and gradually begin discussing work attitude with your junior colleague. That's a wonderful first step. I feel you're taking a very positive approach. Is there anything else you need support with regarding this plan? Or would it be alright to conclude our conversation here?

\textbf{Client:} No, I don't need any additional support at this time. Being able to consult with you as a counselor has helped me organize my thoughts. I'll work on this positively. Thank you for your time today.

\hypertarget{b-3-claude-smdp-counselor}{%
\subsubsection*{(b)-3 Claude-SMDP Counselor}\label{b-3-claude-smdp-counselor}}

\textbf{Client:} I'd like to discuss my work concerns.

\textbf{Counselor:} Please tell me about your work concerns. Let's find solutions together. What aspects are troubling you?

\textbf{Client:} Recently, my relationship with my junior colleague hasn't been going well, and I'm feeling stressed.

\textbf{Counselor:} You're feeling stressed because your relationship with your junior colleague isn't going well. Everyone has relationship concerns, but relationships with close colleagues like juniors are particularly important, so I understand your feelings well.

\textbf{Client:} I don't know how to interact as a supervisor anymore\ldots{}

\textbf{Counselor:} I see, you're troubled about how to interact with your junior colleague as a supervisor. To improve the relationship with your junior, it might be important to review your communication approach. Creating an environment where you can openly discuss each other's thoughts and opinions might lead to improved relationships.

\textbf{Client:} Setting up a discussion opportunity sounds like a good idea. But I'm anxious about whether I can communicate well\ldots{}

\textbf{Counselor:} Yes, by setting up discussions, you can share thoughts and deepen mutual understanding. However, it's natural to feel anxious about whether you can communicate well. Communication isn't easy.

But if you face them sincerely and show a listening attitude, it will surely get through to them. Rather than aiming for perfection, please value the mindset of progressing step by step. Trust relationships can't be built overnight. But by continuing efforts without giving up, things will surely move in a positive direction.

When you overcome this situation, what kind of ideal relationship would you like to have? Could you tell me about the trust relationship you want to build with your junior colleague?

\textbf{Client:} I want to build a well-ventilated relationship where I trust my junior and am trusted by them in return.

\textbf{Counselor:} I see, you want to aim for a well-ventilated relationship where you trust each other. I feel you're envisioning an ideal relationship where you face and respect each other as individuals beyond the supervisor-subordinate positions.

To build such a relationship, I think it's important to maintain open communication daily. Listen carefully to your junior's words and express your thoughts honestly. There might be conflicts at times, but if you don't forget the desire for mutual growth, trust will surely deepen.

By the way, what efforts have you made so far to approach this ideal? Please tell me if you've practiced anything to improve your relationship with your junior colleague.

\textbf{Client:} I've been trying to listen carefully to my junior's stories and properly praise their work achievements.

\textbf{Counselor:} You've been making efforts to build trust by listening to your junior's stories and properly praising their work achievements. The attitude of sincerely listening to subordinates' stories is very important as a supervisor. By praising, you can enhance your junior's motivation and lead to better work.

These daily accumulations should become the foundation for building a well-ventilated relationship. On the other hand, not just praise but sometimes strict feedback might be necessary. Growth advice is conveyed precisely because there's a trust relationship.

Is there anything else you're mindful of to improve your relationship with your junior? For example, are you providing support beyond work, such as being available for private concerns, or starting any new initiatives?

\textbf{Client:} Recently, I've been trying to listen not just to work matters but also to private concerns.

\textbf{Counselor:} I think it's a wonderful initiative to listen to private concerns as well. An attitude of being supportive not just about work but also personal issues will enhance your trustworthiness as a supervisor. Your junior might come to see you not just as a supervisor but as a reliable senior or consultant.

However, balance is important to avoid getting too deeply involved in private matters. Don't forget that work is central, and maintain appropriate boundaries.

Now, while continuing these efforts, it might be effective to utilize various internal and external resources. For example, participating in communication training for managers or consulting with trusted senior supervisors---there are various means of getting support. What resources come to mind that you could utilize?

\textbf{Client:} Well, it might be good to consult with HR.

\textbf{Counselor:} Consulting with HR is a good idea. HR personnel should be well-versed in organizational communication and relationship challenges. They'll provide accurate advice from an objective standpoint.

Also, through consultation with HR, you might discover resources you weren't aware of, such as internal training systems. It could be a good opportunity to improve your management skills.

It would also be effective to exchange information with other managers facing similar concerns or seek advice from trusted senior supervisors. It's important not to carry the burden alone but to actively utilize surrounding support.

Now, let's think about specific future actions. What would be a good first step toward building your ideal relationship? Even starting with something feasible, like setting up a consultation with HR, would be fine. Please share your thoughts.

\textbf{Client:} First, I'll try to set up a consultation opportunity with HR. I hope to receive various advice there.

\textbf{Counselor:} Making HR consultation your first step is an excellent choice. Advice from a professional perspective will surely be a great help in solving problems.

Your attitude of steadily taking action toward building trust while incorporating external insights is wonderful. I hope you gain new awareness and learning through consultation with HR.

Also, while referring to HR's advice, please don't forget to value your own experience and intuition. You're the one who best understands the actual situation. Ultimately, your judgment and action will lead your relationship with your junior in a positive direction.

I wholeheartedly support your attitude of trying to move forward step by step. Please praise yourself for being able to take the first step. We'll conclude today's consultation here, but please feel free to reach out anytime if you need. Let's aim for a better direction together!

\textbf{Client:} Thank you. With your advice in mind, I want to continue doing my best.

\textbf{Counselor:} I think maintaining a positive attitude is most important. You might face difficulties at times, but please don't forget that you have colleagues to encourage each other and to believe in your own growth.

Through this consultation, I could feel your sincere commitment to leadership. That commitment is surely being conveyed to those around you. Please continue to demonstrate your unique leadership and build trusting relationships with your juniors.

I sincerely hope your efforts bear fruit and your workplace becomes vibrant. Please consult again if anything comes up. While looking back on your journey so far and praising yourself, I hope you'll take the next step forward.

It was an honor to speak with you today. I'm supporting your future success from the shadows!

\textbf{Client:} Thank you for your encouraging words. I'll move forward step by step.

\hypertarget{s5-reliability-and-diagnostic-sensitivity-analyses}{%
\section*{S5 Reliability and Diagnostic Sensitivity Analyses}\label{s5-reliability-and-diagnostic-sensitivity-analyses}}

\hypertarget{s5.1-four-system-analysis-dataset}{%
\subsection*{S5.1 Four-System Analysis Dataset}\label{s5.1-four-system-analysis-dataset}}

The analysis using maximum reasoning/effort settings includes four LLM evaluator systems: GPT-5.5, Gemini 3.5 Flash, Claude Opus 4.8, and Fable 5. Descriptive score summaries, transcript-level score-difference outputs, figures, qualitative-comparison outputs, mixed models, ICC outputs, equivalence tests, observed-power diagnostics, and simr power outputs were generated with all four evaluator systems. The earlier Fable 5 GUI evaluation is excluded from these outputs. The post hoc power and equivalence outputs are reported as descriptive diagnostics only and are not used as substantive evidence for the manuscript conclusions.

The current analysis dataset had the following dimensions:

\begin{longtable}[]{@{}lr@{}}
\toprule\noalign{}
Component & Current count \\
\midrule\noalign{}
\endhead
\bottomrule\noalign{}
\endlastfoot
Expert response rows & 54 \\
Expert evaluators & 15 \\
Expert metric rows & 810 \\
Automated metric rows in current analysis outputs & 1,080 \\
Full metric rows in current analysis outputs & 1,890 \\
LLM evaluator systems in inferential outputs & 4 \\
API-based evaluator systems & 3 \\
Non-API evaluator systems & 1 \\
Score iterations per automated evaluator system & 3 \\
Qualitative summary rows in current analysis outputs & 72 \\
Qualitative improvement-detail rows in current analysis outputs & 72 \\
\end{longtable}

The expert evaluation records are described in S2. The analytic dataset contains script ratings, client-naturalness ratings, and qualitative comments about the synthetic dialogues. Evaluator qualifications and experience are reported only as aggregate panel descriptors; no patient or counseling-client data, human counseling transcripts, or identifiable evaluator-level results are included.

Exploratory equivalence tests used the two one-sided tests (TOST) procedure with equivalence bounds set to ±0.5 observed \emph{SD}s for each analysis/outcome set and 90\% confidence intervals. Because these bounds were not prespecified as substantively justified minimal important differences, the TOST outputs are not used to claim equivalence. Power for the likelihood-ratio tests was estimated with the simr package, using 1,000 simulations under the current model specifications. The observed contrast power results below are reported only as descriptive model-output diagnostics and are not used to support substantive conclusions.

\hypertarget{s5.2-simulation-based-power-simr-for-likelihood-ratio-tests}{%
\subsection*{S5.2 Simulation-Based Power (simr) for Likelihood-Ratio Tests}\label{s5.2-simulation-based-power-simr-for-likelihood-ratio-tests}}

\begin{longtable}[]{@{}
  >{\raggedright\arraybackslash}p{(\columnwidth - 12\tabcolsep) * \real{0.3246}}
  >{\raggedright\arraybackslash}p{(\columnwidth - 12\tabcolsep) * \real{0.0965}}
  >{\raggedleft\arraybackslash}p{(\columnwidth - 12\tabcolsep) * \real{0.0702}}
  >{\raggedleft\arraybackslash}p{(\columnwidth - 12\tabcolsep) * \real{0.1140}}
  >{\raggedright\arraybackslash}p{(\columnwidth - 12\tabcolsep) * \real{0.1667}}
  >{\raggedleft\arraybackslash}p{(\columnwidth - 12\tabcolsep) * \real{0.1228}}
  >{\raggedleft\arraybackslash}p{(\columnwidth - 12\tabcolsep) * \real{0.1053}}@{}}
\toprule\noalign{}
\begin{minipage}[b]{\linewidth}\raggedright
Analysis
\end{minipage} & \begin{minipage}[b]{\linewidth}\raggedright
Outcome
\end{minipage} & \begin{minipage}[b]{\linewidth}\raggedleft
Simulations
\end{minipage} & \begin{minipage}[b]{\linewidth}\raggedleft
Successes
\end{minipage} & \begin{minipage}[b]{\linewidth}\raggedright
Power (95\% CI)
\end{minipage} & \begin{minipage}[b]{\linewidth}\raggedleft
Warnings
\end{minipage} & \begin{minipage}[b]{\linewidth}\raggedleft
Errors
\end{minipage} \\
\midrule\noalign{}
\endhead
\bottomrule\noalign{}
\endlastfoot
Expert-only counselor-condition likelihood-ratio test & CCT & 1000 & 1000 & 1.00 {[}1.00, 1.00{]} & 2 & 0 \\
Counselor-condition by evaluator-type likelihood-ratio test & CCT & 1000 & 1000 & 1.00 {[}1.00, 1.00{]} & 2 & 0 \\
Expert-only counselor-condition likelihood-ratio test & SST & 1000 & 859 & 0.86 {[}0.84, 0.88{]} & 0 & 0 \\
Counselor-condition by evaluator-type likelihood-ratio test & SST & 1000 & 936 & 0.94 {[}0.92, 0.95{]} & 2 & 0 \\
Expert-only counselor-condition likelihood-ratio test & PAR & 1000 & 1000 & 1.00 {[}1.00, 1.00{]} & 0 & 0 \\
Counselor-condition by evaluator-type likelihood-ratio test & PAR & 1000 & 1000 & 1.00 {[}1.00, 1.00{]} & 3 & 0 \\
Expert-only counselor-condition likelihood-ratio test & EMP & 1000 & 1000 & 1.00 {[}1.00, 1.00{]} & 2 & 0 \\
Counselor-condition by evaluator-type likelihood-ratio test & EMP & 1000 & 740 & 0.74 {[}0.71, 0.77{]} & 6 & 0 \\
Expert-only counselor-condition likelihood-ratio test & OVR & 1000 & 1000 & 1.00 {[}1.00, 1.00{]} & 0 & 0 \\
Counselor-condition by evaluator-type likelihood-ratio test & OVR & 1000 & 1000 & 1.00 {[}1.00, 1.00{]} & 12 & 0 \\
\end{longtable}

\hypertarget{s5.3-observed-contrast-power-diagnostics}{%
\subsection*{S5.3 Observed Contrast Power Diagnostics}\label{s5.3-observed-contrast-power-diagnostics}}

\begin{longtable}[]{@{}
  >{\raggedright\arraybackslash}p{(\columnwidth - 10\tabcolsep) * \real{0.4286}}
  >{\raggedright\arraybackslash}p{(\columnwidth - 10\tabcolsep) * \real{0.0924}}
  >{\raggedleft\arraybackslash}p{(\columnwidth - 10\tabcolsep) * \real{0.1261}}
  >{\raggedleft\arraybackslash}p{(\columnwidth - 10\tabcolsep) * \real{0.1092}}
  >{\raggedleft\arraybackslash}p{(\columnwidth - 10\tabcolsep) * \real{0.1345}}
  >{\raggedleft\arraybackslash}p{(\columnwidth - 10\tabcolsep) * \real{0.1092}}@{}}
\toprule\noalign{}
\begin{minipage}[b]{\linewidth}\raggedright
Analysis
\end{minipage} & \begin{minipage}[b]{\linewidth}\raggedright
Outcome
\end{minipage} & \begin{minipage}[b]{\linewidth}\raggedleft
Contrasts
\end{minipage} & \begin{minipage}[b]{\linewidth}\raggedleft
Minimum power
\end{minipage} & \begin{minipage}[b]{\linewidth}\raggedleft
Median power
\end{minipage} & \begin{minipage}[b]{\linewidth}\raggedleft
Maximum power
\end{minipage} \\
\midrule\noalign{}
\endhead
\bottomrule\noalign{}
\endlastfoot
Expert-only counselor-condition contrasts & CCT & 3 & 0.05 & 1.00 & 1.00 \\
Evaluator-type contrasts within counselor condition & CCT & 30 & 0.05 & 0.41 & 1.00 \\
Counselor-condition contrasts within evaluator type & CCT & 15 & 0.05 & 1.00 & 1.00 \\
Expert-only counselor-condition contrasts & SST & 3 & 0.18 & 0.65 & 0.92 \\
Evaluator-type contrasts within counselor condition & SST & 30 & 0.05 & 0.51 & 1.00 \\
Counselor-condition contrasts within evaluator type & SST & 15 & 0.05 & 0.16 & 1.00 \\
Expert-only counselor-condition contrasts & PAR & 3 & 0.38 & 1.00 & 1.00 \\
Evaluator-type contrasts within counselor condition & PAR & 30 & 0.05 & 0.49 & 1.00 \\
Counselor-condition contrasts within evaluator type & PAR & 15 & 0.07 & 1.00 & 1.00 \\
Expert-only counselor-condition contrasts & EMP & 3 & 0.08 & 1.00 & 1.00 \\
Evaluator-type contrasts within counselor condition & EMP & 30 & 0.05 & 0.33 & 1.00 \\
Counselor-condition contrasts within evaluator type & EMP & 15 & 0.05 & 0.97 & 1.00 \\
Expert-only counselor-condition contrasts & OVR & 3 & 0.06 & 1.00 & 1.00 \\
Evaluator-type contrasts within counselor condition & OVR & 30 & 0.05 & 0.94 & 1.00 \\
Counselor-condition contrasts within evaluator type & OVR & 15 & 0.06 & 0.54 & 1.00 \\
\end{longtable}

\hypertarget{s5.4-exploratory-tost-diagnostic-summaries}{%
\subsection*{S5.4 Exploratory TOST Diagnostic Summaries}\label{s5.4-exploratory-tost-diagnostic-summaries}}

\begin{longtable}[]{@{}
  >{\raggedright\arraybackslash}p{(\columnwidth - 10\tabcolsep) * \real{0.3953}}
  >{\raggedright\arraybackslash}p{(\columnwidth - 10\tabcolsep) * \real{0.0853}}
  >{\raggedleft\arraybackslash}p{(\columnwidth - 10\tabcolsep) * \real{0.1163}}
  >{\raggedleft\arraybackslash}p{(\columnwidth - 10\tabcolsep) * \real{0.1240}}
  >{\raggedright\arraybackslash}p{(\columnwidth - 10\tabcolsep) * \real{0.1395}}
  >{\raggedleft\arraybackslash}p{(\columnwidth - 10\tabcolsep) * \real{0.1395}}@{}}
\toprule\noalign{}
\begin{minipage}[b]{\linewidth}\raggedright
Analysis
\end{minipage} & \begin{minipage}[b]{\linewidth}\raggedright
Outcome
\end{minipage} & \begin{minipage}[b]{\linewidth}\raggedleft
Contrasts
\end{minipage} & \begin{minipage}[b]{\linewidth}\raggedleft
Equivalent contrasts
\end{minipage} & \begin{minipage}[b]{\linewidth}\raggedright
Minimum larger TOST \emph{p}
\end{minipage} & \begin{minipage}[b]{\linewidth}\raggedleft
Maximum larger TOST \emph{p}
\end{minipage} \\
\midrule\noalign{}
\endhead
\bottomrule\noalign{}
\endlastfoot
Expert-only counselor-condition contrasts & CCT & 3 & 1 & \textless{} .001 & .999 \\
Evaluator-type contrasts within counselor condition & CCT & 30 & 0 & .064 & .999 \\
Counselor-condition contrasts within evaluator type & CCT & 15 & 3 & \textless{} .001 & 1.000 \\
Expert-only counselor-condition contrasts & SST & 3 & 1 & .021 & .620 \\
Evaluator-type contrasts within counselor condition & SST & 30 & 0 & .080 & 1.000 \\
Counselor-condition contrasts within evaluator type & SST & 15 & 8 & .003 & .702 \\
Expert-only counselor-condition contrasts & PAR & 3 & 1 & .034 & 1.000 \\
Evaluator-type contrasts within counselor condition & PAR & 30 & 2 & .025 & .996 \\
Counselor-condition contrasts within evaluator type & PAR & 15 & 2 & .019 & 1.000 \\
Expert-only counselor-condition contrasts & EMP & 3 & 1 & .003 & .999 \\
Evaluator-type contrasts within counselor condition & EMP & 30 & 0 & .089 & .999 \\
Counselor-condition contrasts within evaluator type & EMP & 15 & 4 & .004 & 1.000 \\
Expert-only counselor-condition contrasts & OVR & 3 & 1 & .002 & 1.000 \\
Evaluator-type contrasts within counselor condition & OVR & 30 & 2 & .011 & 1.000 \\
Counselor-condition contrasts within evaluator type & OVR & 15 & 6 & \textless{} .001 & 1.000 \\
\end{longtable}

\hypertarget{s5.5-run-to-run-score-reliability-for-llm-evaluators-and-interrater-reliability-for-experts-icc}{%
\subsection*{S5.5 Run-to-Run Score Reliability for LLM Evaluators and Interrater Reliability for Experts (ICC)}\label{s5.5-run-to-run-score-reliability-for-llm-evaluators-and-interrater-reliability-for-experts-icc}}

Run-to-run score reliability for the LLM evaluator systems and interrater reliability for the expert panel were estimated with the ICC function in the {psych} package, using its linear mixed-effects model option. This option was used because the expert-rating matrix is sparse: expert evaluators did not each rate every target, whereas each automated evaluator system had three complete score iterations per target. Fable 5 was included as the fourth evaluator system. For the expert panel, average-measure ICCs are descriptive summaries under an incomplete rating matrix; they do not directly estimate the reliability of the observed transcript means, which were based on 8--10 evaluators per transcript without a common complete panel of all 15 evaluators.

\begin{longtable}[]{@{}
  >{\raggedright\arraybackslash}p{(\columnwidth - 10\tabcolsep) * \real{0.1895}}
  >{\raggedright\arraybackslash}p{(\columnwidth - 10\tabcolsep) * \real{0.1158}}
  >{\raggedright\arraybackslash}p{(\columnwidth - 10\tabcolsep) * \real{0.1263}}
  >{\raggedright\arraybackslash}p{(\columnwidth - 10\tabcolsep) * \real{0.2105}}
  >{\raggedleft\arraybackslash}p{(\columnwidth - 10\tabcolsep) * \real{0.1263}}
  >{\raggedleft\arraybackslash}p{(\columnwidth - 10\tabcolsep) * \real{0.2316}}@{}}
\toprule\noalign{}
\begin{minipage}[b]{\linewidth}\raggedright
Evaluator type
\end{minipage} & \begin{minipage}[b]{\linewidth}\raggedright
Outcome
\end{minipage} & \begin{minipage}[b]{\linewidth}\raggedright
ICC type
\end{minipage} & \begin{minipage}[b]{\linewidth}\raggedright
ICC (95\% CI)
\end{minipage} & \begin{minipage}[b]{\linewidth}\raggedleft
Raters or iterations
\end{minipage} & \begin{minipage}[b]{\linewidth}\raggedleft
Complete targets
\end{minipage} \\
\midrule\noalign{}
\endhead
\bottomrule\noalign{}
\endlastfoot
Expert & CCT & ICC(2,1) & 0.20 {[}0.10, 0.39{]} & 15 & 0 \\
Expert & CCT & ICC(2,\emph{k}) & 0.79 {[}0.62, 0.91{]} & 15 & 0 \\
GPT-5.5 & CCT & ICC(2,1) & 0.85 {[}0.71, 0.94{]} & 3 & 18 \\
GPT-5.5 & CCT & ICC(2,\emph{k}) & 0.94 {[}0.88, 0.98{]} & 3 & 18 \\
Gemini 3.5 Flash & CCT & ICC(2,1) & 0.94 {[}0.87, 0.97{]} & 3 & 18 \\
Gemini 3.5 Flash & CCT & ICC(2,\emph{k}) & 0.98 {[}0.95, 0.99{]} & 3 & 18 \\
Claude Opus 4.8 & CCT & ICC(2,1) & 0.86 {[}0.73, 0.94{]} & 3 & 18 \\
Claude Opus 4.8 & CCT & ICC(2,\emph{k}) & 0.95 {[}0.89, 0.98{]} & 3 & 18 \\
Fable 5 & CCT & ICC(2,1) & 0.96 {[}0.91, 0.98{]} & 3 & 18 \\
Fable 5 & CCT & ICC(2,\emph{k}) & 0.98 {[}0.97, 0.99{]} & 3 & 18 \\
Expert & SST & ICC(2,1) & 0.03 {[}−0.01, 0.12{]} & 15 & 0 \\
Expert & SST & ICC(2,\emph{k}) & 0.30 {[}−0.14, 0.67{]} & 15 & 0 \\
GPT-5.5 & SST & ICC(2,1) & 0.66 {[}0.42, 0.84{]} & 3 & 18 \\
GPT-5.5 & SST & ICC(2,\emph{k}) & 0.85 {[}0.68, 0.94{]} & 3 & 18 \\
Gemini 3.5 Flash & SST & ICC(2,1) & 0.38 {[}0.09, 0.66{]} & 3 & 18 \\
Gemini 3.5 Flash & SST & ICC(2,\emph{k}) & 0.64 {[}0.23, 0.86{]} & 3 & 18 \\
Claude Opus 4.8 & SST & ICC(2,1) & 0.33 {[}0.05, 0.63{]} & 3 & 18 \\
Claude Opus 4.8 & SST & ICC(2,\emph{k}) & 0.60 {[}0.14, 0.84{]} & 3 & 18 \\
Fable 5 & SST & ICC(2,1) & 0.46 {[}0.18, 0.72{]} & 3 & 18 \\
Fable 5 & SST & ICC(2,\emph{k}) & 0.72 {[}0.39, 0.89{]} & 3 & 18 \\
Expert & PAR & ICC(2,1) & 0.26 {[}0.14, 0.46{]} & 15 & 0 \\
Expert & PAR & ICC(2,\emph{k}) & 0.84 {[}0.71, 0.93{]} & 15 & 0 \\
GPT-5.5 & PAR & ICC(2,1) & 0.92 {[}0.85, 0.97{]} & 3 & 18 \\
GPT-5.5 & PAR & ICC(2,\emph{k}) & 0.97 {[}0.94, 0.99{]} & 3 & 18 \\
Gemini 3.5 Flash & PAR & ICC(2,1) & 0.95 {[}0.89, 0.98{]} & 3 & 18 \\
Gemini 3.5 Flash & PAR & ICC(2,\emph{k}) & 0.98 {[}0.96, 0.99{]} & 3 & 18 \\
Claude Opus 4.8 & PAR & ICC(2,1) & 0.90 {[}0.80, 0.96{]} & 3 & 18 \\
Claude Opus 4.8 & PAR & ICC(2,\emph{k}) & 0.97 {[}0.92, 0.99{]} & 3 & 18 \\
Fable 5 & PAR & ICC(2,1) & 0.93 {[}0.86, 0.97{]} & 3 & 18 \\
Fable 5 & PAR & ICC(2,\emph{k}) & 0.98 {[}0.95, 0.99{]} & 3 & 18 \\
Expert & EMP & ICC(2,1) & 0.20 {[}0.10, 0.39{]} & 15 & 0 \\
Expert & EMP & ICC(2,\emph{k}) & 0.79 {[}0.62, 0.90{]} & 15 & 0 \\
GPT-5.5 & EMP & ICC(2,1) & 0.92 {[}0.83, 0.96{]} & 3 & 18 \\
GPT-5.5 & EMP & ICC(2,\emph{k}) & 0.97 {[}0.93, 0.99{]} & 3 & 18 \\
Gemini 3.5 Flash & EMP & ICC(2,1) & 0.84 {[}0.68, 0.93{]} & 3 & 18 \\
Gemini 3.5 Flash & EMP & ICC(2,\emph{k}) & 0.94 {[}0.87, 0.98{]} & 3 & 18 \\
Claude Opus 4.8 & EMP & ICC(2,1) & 0.85 {[}0.71, 0.94{]} & 3 & 18 \\
Claude Opus 4.8 & EMP & ICC(2,\emph{k}) & 0.95 {[}0.88, 0.98{]} & 3 & 18 \\
Fable 5 & EMP & ICC(2,1) & 0.93 {[}0.86, 0.97{]} & 3 & 18 \\
Fable 5 & EMP & ICC(2,\emph{k}) & 0.98 {[}0.95, 0.99{]} & 3 & 18 \\
Expert & OVR & ICC(2,1) & 0.25 {[}0.14, 0.46{]} & 15 & 0 \\
Expert & OVR & ICC(2,\emph{k}) & 0.84 {[}0.70, 0.93{]} & 15 & 0 \\
GPT-5.5 & OVR & ICC(2,1) & 0.72 {[}0.50, 0.87{]} & 3 & 18 \\
GPT-5.5 & OVR & ICC(2,\emph{k}) & 0.89 {[}0.75, 0.95{]} & 3 & 18 \\
Gemini 3.5 Flash & OVR & ICC(2,1) & 0.51 {[}0.24, 0.76{]} & 3 & 18 \\
Gemini 3.5 Flash & OVR & ICC(2,\emph{k}) & 0.76 {[}0.48, 0.90{]} & 3 & 18 \\
Claude Opus 4.8 & OVR & ICC(2,1) & 0.82 {[}0.65, 0.92{]} & 3 & 18 \\
Claude Opus 4.8 & OVR & ICC(2,\emph{k}) & 0.93 {[}0.85, 0.97{]} & 3 & 18 \\
Fable 5 & OVR & ICC(2,1) & 0.81 {[}0.64, 0.92{]} & 3 & 18 \\
Fable 5 & OVR & ICC(2,\emph{k}) & 0.93 {[}0.84, 0.97{]} & 3 & 18 \\
\end{longtable}

\hypertarget{s6-mixed-model-specifications-contrasts-interaction-tests-and-sensitivity-analyses}{%
\section*{S6 Mixed-Model Specifications, Contrasts, Interaction Tests, and Sensitivity Analyses}\label{s6-mixed-model-specifications-contrasts-interaction-tests-and-sensitivity-analyses}}

The tables below report the mixed-model specifications, Tukey-adjusted contrasts, interaction-model outputs, and additional sensitivity analyses from the four-system analysis using maximum reasoning/effort settings. These inferential models include GPT-5.5, Gemini 3.5 Flash, Claude Opus 4.8, and Fable 5. Contrast estimates are estimated marginal mean differences for the first level minus the second level unless otherwise noted; negative values therefore indicate that the first level scored lower than the second level.

All counselor-condition analyses should be read as fixed-benchmark analyses of 18 pregenerated complete sessions. Each counselor condition was represented by six separately initiated and generated sessions across the prespecified client-profile contexts. Following the initial client message, each session comprised 10 sequential counselor--client exchanges generated through separate, history-conditioned calls. The estimand was the rated quality of the complete interaction produced under each counselor condition, not the isolated quality of counselor responses to identical client turns. The study did not estimate repeated-generation variability within an identical counselor-condition-by-profile cell or average performance across all possible stochastic regenerations. The transcript and client-profile terms address variation within this fixed benchmark and should not be interpreted as random sampling from a broader population of generated dialogues.

\hypertarget{s6.1-mixed-model-specifications}{%
\subsection*{S6.1 Mixed-Model Specifications}\label{s6.1-mixed-model-specifications}}

\hypertarget{expert-evaluation-of-counselor-conditions}{%
\subsubsection*{Expert Evaluation of Counselor Conditions}\label{expert-evaluation-of-counselor-conditions}}

Each outcome was analyzed separately using the following linear mixed-effects model:

\[
\begin{aligned}
\mathrm{Score}_{ijk} =\;&
\beta_{0}
+ \beta_{1}\,\mathrm{Model}_{\text{GPT-SMDP},ijk}
+ \beta_{2}\,\mathrm{Model}_{\text{Claude-SMDP},ijk} \\
& + \beta_{3}\,\mathrm{AgeCat}_{40\text{s},ijk}
+ \beta_{4}\,\mathrm{AgeCat}_{60\text{s},ijk}
+ \beta_{5}\,\mathrm{Sex}_{\text{male},ijk} \\
& + u_{0j} + v_{0k} + \varepsilon_{ijk},
\end{aligned}
\]

where \(u_{0j}\) is the random intercept for evaluator \(j\) and \(v_{0k}\) is the random intercept for transcript \(k\). GPT-minimal, clients in their 20s, and female clients were the reference categories. The omnibus counselor-condition effect was tested by comparing this model with a model that omitted counselor condition but retained the client covariates and both random intercepts. Pairwise comparisons used Tukey--Kramer adjustment.

\hypertarget{expertllm-evaluator-comparison}{%
\subsubsection*{Expert--LLM Evaluator Comparison}\label{expertllm-evaluator-comparison}}

Each outcome was also analyzed using the following interaction model for expert ratings, the three API-based LLM evaluator systems, and Fable 5:

\[
\begin{aligned}
\mathrm{Score}_{ijk} =\;&
\beta_{0}
+ \boldsymbol{\beta}_{1\text{-}4}\,\mathrm{EvaluatorType}_{ijk}
+ \boldsymbol{\beta}_{5\text{-}6}\,\mathrm{Model}_{ijk} \\
& + \boldsymbol{\beta}_{7\text{-}14}
  (\mathrm{EvaluatorType}_{ijk}\times\mathrm{Model}_{ijk}) \\
& + \boldsymbol{\beta}_{15\text{-}16}\,\mathrm{AgeCat}_{ijk}
+ \beta_{17}\,\mathrm{Sex}_{ijk} \\
& + u_{0j} + v_{0k} + \varepsilon_{ijk}.
\end{aligned}
\]

The interaction was tested against the corresponding additive model with the same covariates and random-effects structure. Tukey--Kramer-adjusted simple effects compared evaluator types within counselor condition. Counselor-condition contrasts within evaluator type were treated as secondary simple effects. Type III Wald tests were reported as supplementary summaries. Fable 5 was included as the fourth evaluator system; because its model identity and execution pathway are confounded, Fable-specific contrasts cannot be attributed to the model alone.

\hypertarget{s6.2-expert-only-counselor-condition-contrasts}{%
\subsection*{S6.2 Expert-Only Counselor-Condition Contrasts}\label{s6.2-expert-only-counselor-condition-contrasts}}

\begin{longtable}[]{@{}
  >{\raggedright\arraybackslash}p{(\columnwidth - 12\tabcolsep) * \real{0.1429}}
  >{\raggedright\arraybackslash}p{(\columnwidth - 12\tabcolsep) * \real{0.1429}}
  >{\raggedleft\arraybackslash}p{(\columnwidth - 12\tabcolsep) * \real{0.1429}}
  >{\raggedleft\arraybackslash}p{(\columnwidth - 12\tabcolsep) * \real{0.1429}}
  >{\raggedleft\arraybackslash}p{(\columnwidth - 12\tabcolsep) * \real{0.1429}}
  >{\raggedleft\arraybackslash}p{(\columnwidth - 12\tabcolsep) * \real{0.1429}}
  >{\raggedright\arraybackslash}p{(\columnwidth - 12\tabcolsep) * \real{0.1429}}@{}}
\toprule\noalign{}
\begin{minipage}[b]{\linewidth}\raggedright
Outcome
\end{minipage} & \begin{minipage}[b]{\linewidth}\raggedright
Contrast
\end{minipage} & \begin{minipage}[b]{\linewidth}\raggedleft
Estimate
\end{minipage} & \begin{minipage}[b]{\linewidth}\raggedleft
SE
\end{minipage} & \begin{minipage}[b]{\linewidth}\raggedleft
df
\end{minipage} & \begin{minipage}[b]{\linewidth}\raggedleft
\emph{t}
\end{minipage} & \begin{minipage}[b]{\linewidth}\raggedright
\emph{p}
\end{minipage} \\
\midrule\noalign{}
\endhead
\bottomrule\noalign{}
\endlastfoot
CCT & (GPT-minimal) − (GPT-SMDP) & −0.88 & 0.12 & 148.92 & −7.04 & \textless{} .001 \\
CCT & (GPT-minimal) − (Claude-SMDP) & −0.88 & 0.12 & 148.92 & −7.04 & \textless{} .001 \\
CCT & (GPT-SMDP) − (Claude-SMDP) & 0.00 & 0.12 & 148.92 & 0.00 & 1.000 \\
SST & (GPT-minimal) − (GPT-SMDP) & −0.33 & 0.14 & 149.99 & −2.36 & .051 \\
SST & (GPT-minimal) − (Claude-SMDP) & −0.48 & 0.14 & 149.99 & −3.41 & .002 \\
SST & (GPT-SMDP) − (Claude-SMDP) & −0.15 & 0.14 & 149.99 & −1.05 & .548 \\
PAR & (GPT-minimal) − (GPT-SMDP) & −1.11 & 0.14 & 149.59 & −7.70 & \textless{} .001 \\
PAR & (GPT-minimal) − (Claude-SMDP) & −0.87 & 0.14 & 149.59 & −6.03 & \textless{} .001 \\
PAR & (GPT-SMDP) − (Claude-SMDP) & 0.24 & 0.14 & 149.59 & 1.67 & .221 \\
EMP & (GPT-minimal) − (GPT-SMDP) & −0.93 & 0.15 & 148.79 & −6.08 & \textless{} .001 \\
EMP & (GPT-minimal) − (Claude-SMDP) & −1.01 & 0.15 & 148.79 & −6.63 & \textless{} .001 \\
EMP & (GPT-SMDP) − (Claude-SMDP) & −0.08 & 0.15 & 148.79 & −0.55 & .848 \\
OVR & (GPT-minimal) − (GPT-SMDP) & −0.83 & 0.13 & 150.16 & −6.50 & \textless{} .001 \\
OVR & (GPT-minimal) − (Claude-SMDP) & −0.86 & 0.13 & 150.16 & −6.72 & \textless{} .001 \\
OVR & (GPT-SMDP) − (Claude-SMDP) & −0.03 & 0.13 & 150.16 & −0.22 & .974 \\
\end{longtable}

\hypertarget{s6.3-sensitivity-additive-evaluator-type-contrasts}{%
\subsection*{S6.3 Sensitivity Additive Evaluator-Type Contrasts}\label{s6.3-sensitivity-additive-evaluator-type-contrasts}}

These contrasts are sensitivity results from an additive evaluator-type model without a transcript random intercept. They are not the primary evaluator-type contrasts.

\begin{longtable}[]{@{}
  >{\raggedright\arraybackslash}p{(\columnwidth - 12\tabcolsep) * \real{0.1429}}
  >{\raggedright\arraybackslash}p{(\columnwidth - 12\tabcolsep) * \real{0.1429}}
  >{\raggedleft\arraybackslash}p{(\columnwidth - 12\tabcolsep) * \real{0.1429}}
  >{\raggedleft\arraybackslash}p{(\columnwidth - 12\tabcolsep) * \real{0.1429}}
  >{\raggedleft\arraybackslash}p{(\columnwidth - 12\tabcolsep) * \real{0.1429}}
  >{\raggedleft\arraybackslash}p{(\columnwidth - 12\tabcolsep) * \real{0.1429}}
  >{\raggedright\arraybackslash}p{(\columnwidth - 12\tabcolsep) * \real{0.1429}}@{}}
\toprule\noalign{}
\begin{minipage}[b]{\linewidth}\raggedright
Outcome
\end{minipage} & \begin{minipage}[b]{\linewidth}\raggedright
Contrast
\end{minipage} & \begin{minipage}[b]{\linewidth}\raggedleft
Estimate
\end{minipage} & \begin{minipage}[b]{\linewidth}\raggedleft
SE
\end{minipage} & \begin{minipage}[b]{\linewidth}\raggedleft
df
\end{minipage} & \begin{minipage}[b]{\linewidth}\raggedleft
\emph{t}
\end{minipage} & \begin{minipage}[b]{\linewidth}\raggedright
\emph{p}
\end{minipage} \\
\midrule\noalign{}
\endhead
\bottomrule\noalign{}
\endlastfoot
CCT & Expert − GPT-5.5 & −0.76 & 0.24 & 25.78 & −3.16 & .030 \\
CCT & Expert − Gemini 3.5 Flash & −0.86 & 0.24 & 25.78 & −3.58 & .011 \\
CCT & Expert − Claude Opus 4.8 & 0.05 & 0.24 & 25.78 & 0.22 & .999 \\
CCT & Expert − Fable 5 & −0.15 & 0.24 & 25.78 & −0.63 & .969 \\
CCT & GPT-5.5 − Gemini 3.5 Flash & −0.10 & 0.31 & 24.81 & −0.33 & .997 \\
CCT & GPT-5.5 − Claude Opus 4.8 & 0.81 & 0.31 & 24.81 & 2.65 & .092 \\
CCT & GPT-5.5 − Fable 5 & 0.61 & 0.31 & 24.81 & 1.98 & .303 \\
CCT & Gemini 3.5 Flash − Claude Opus 4.8 & 0.92 & 0.31 & 24.81 & 2.98 & .046 \\
CCT & Gemini 3.5 Flash − Fable 5 & 0.71 & 0.31 & 24.81 & 2.31 & .174 \\
CCT & Claude Opus 4.8 − Fable 5 & −0.20 & 0.31 & 24.81 & −0.66 & .963 \\
SST & Expert − GPT-5.5 & −1.18 & 0.19 & 25.14 & −6.19 & \textless{} .001 \\
SST & Expert − Gemini 3.5 Flash & −1.49 & 0.19 & 25.14 & −7.85 & \textless{} .001 \\
SST & Expert − Claude Opus 4.8 & −0.93 & 0.19 & 25.14 & −4.88 & \textless{} .001 \\
SST & Expert − Fable 5 & −0.98 & 0.19 & 25.14 & −5.17 & \textless{} .001 \\
SST & GPT-5.5 − Gemini 3.5 Flash & −0.31 & 0.24 & 23.84 & −1.30 & .691 \\
SST & GPT-5.5 − Claude Opus 4.8 & 0.25 & 0.24 & 23.84 & 1.04 & .836 \\
SST & GPT-5.5 − Fable 5 & 0.19 & 0.24 & 23.84 & 0.81 & .926 \\
SST & Gemini 3.5 Flash − Claude Opus 4.8 & 0.56 & 0.24 & 23.84 & 2.34 & .167 \\
SST & Gemini 3.5 Flash − Fable 5 & 0.51 & 0.24 & 23.84 & 2.11 & .249 \\
SST & Claude Opus 4.8 − Fable 5 & −0.06 & 0.24 & 23.84 & −0.23 & .999 \\
PAR & Expert − GPT-5.5 & −0.73 & 0.21 & 24.74 & −3.45 & .016 \\
PAR & Expert − Gemini 3.5 Flash & −0.43 & 0.21 & 24.74 & −2.05 & .273 \\
PAR & Expert − Claude Opus 4.8 & 0.05 & 0.21 & 24.74 & 0.23 & .999 \\
PAR & Expert − Fable 5 & −0.34 & 0.21 & 24.74 & −1.61 & .504 \\
PAR & GPT-5.5 − Gemini 3.5 Flash & 0.30 & 0.27 & 23.36 & 1.11 & .802 \\
PAR & GPT-5.5 − Claude Opus 4.8 & 0.78 & 0.27 & 23.36 & 2.90 & .055 \\
PAR & GPT-5.5 − Fable 5 & 0.39 & 0.27 & 23.36 & 1.45 & .602 \\
PAR & Gemini 3.5 Flash − Claude Opus 4.8 & 0.48 & 0.27 & 23.36 & 1.80 & .399 \\
PAR & Gemini 3.5 Flash − Fable 5 & 0.09 & 0.27 & 23.36 & 0.35 & .997 \\
PAR & Claude Opus 4.8 − Fable 5 & −0.39 & 0.27 & 23.36 & −1.45 & .602 \\
EMP & Expert − GPT-5.5 & −0.92 & 0.20 & 21.99 & −4.59 & .001 \\
EMP & Expert − Gemini 3.5 Flash & −0.62 & 0.20 & 21.99 & −3.11 & .037 \\
EMP & Expert − Claude Opus 4.8 & −0.36 & 0.20 & 21.99 & −1.77 & .413 \\
EMP & Expert − Fable 5 & −0.55 & 0.20 & 21.99 & −2.74 & .079 \\
EMP & GPT-5.5 − Gemini 3.5 Flash & 0.30 & 0.25 & 20.80 & 1.16 & .771 \\
EMP & GPT-5.5 − Claude Opus 4.8 & 0.56 & 0.25 & 20.80 & 2.22 & .212 \\
EMP & GPT-5.5 − Fable 5 & 0.37 & 0.25 & 20.80 & 1.45 & .601 \\
EMP & Gemini 3.5 Flash − Claude Opus 4.8 & 0.27 & 0.25 & 20.80 & 1.05 & .827 \\
EMP & Gemini 3.5 Flash − Fable 5 & 0.07 & 0.25 & 20.80 & 0.29 & .998 \\
EMP & Claude Opus 4.8 − Fable 5 & −0.19 & 0.25 & 20.80 & −0.76 & .938 \\
OVR & Expert − GPT-5.5 & −1.53 & 0.13 & 21.71 & −12.04 & \textless{} .001 \\
OVR & Expert − Gemini 3.5 Flash & −1.93 & 0.13 & 21.71 & −15.18 & \textless{} .001 \\
OVR & Expert − Claude Opus 4.8 & −1.17 & 0.13 & 21.71 & −9.20 & \textless{} .001 \\
OVR & Expert − Fable 5 & −1.36 & 0.13 & 21.71 & −10.73 & \textless{} .001 \\
OVR & GPT-5.5 − Gemini 3.5 Flash & −0.40 & 0.16 & 19.96 & −2.50 & .129 \\
OVR & GPT-5.5 − Claude Opus 4.8 & 0.36 & 0.16 & 19.96 & 2.27 & .195 \\
OVR & GPT-5.5 − Fable 5 & 0.17 & 0.16 & 19.96 & 1.05 & .830 \\
OVR & Gemini 3.5 Flash − Claude Opus 4.8 & 0.76 & 0.16 & 19.96 & 4.78 & \textless{} .001 \\
OVR & Gemini 3.5 Flash − Fable 5 & 0.56 & 0.16 & 19.96 & 3.55 & .015 \\
OVR & Claude Opus 4.8 − Fable 5 & −0.19 & 0.16 & 19.96 & −1.22 & .738 \\
\end{longtable}

\hypertarget{s6.4-interaction-model-type-iii-analysis-of-variance}{%
\subsection*{S6.4 Interaction Model Type III Analysis of Variance}\label{s6.4-interaction-model-type-iii-analysis-of-variance}}

\begin{longtable}[]{@{}
  >{\raggedright\arraybackslash}p{(\columnwidth - 8\tabcolsep) * \real{0.1618}}
  >{\raggedright\arraybackslash}p{(\columnwidth - 8\tabcolsep) * \real{0.4706}}
  >{\raggedleft\arraybackslash}p{(\columnwidth - 8\tabcolsep) * \real{0.1029}}
  >{\raggedright\arraybackslash}p{(\columnwidth - 8\tabcolsep) * \real{0.1471}}
  >{\raggedright\arraybackslash}p{(\columnwidth - 8\tabcolsep) * \real{0.1176}}@{}}
\toprule\noalign{}
\begin{minipage}[b]{\linewidth}\raggedright
Outcome
\end{minipage} & \begin{minipage}[b]{\linewidth}\raggedright
Term
\end{minipage} & \begin{minipage}[b]{\linewidth}\raggedleft
F
\end{minipage} & \begin{minipage}[b]{\linewidth}\raggedright
df
\end{minipage} & \begin{minipage}[b]{\linewidth}\raggedright
\emph{p}
\end{minipage} \\
\midrule\noalign{}
\endhead
\bottomrule\noalign{}
\endlastfoot
CCT & Counselor condition × evaluator type & 10.56 & 8, 335.3 & \textless{} .001 \\
SST & Counselor condition × evaluator type & 2.59 & 8, 352.8 & .009 \\
PAR & Counselor condition × evaluator type & 11.42 & 8, 335.0 & \textless{} .001 \\
EMP & Counselor condition × evaluator type & 1.65 & 8, 332.3 & .111 \\
OVR & Counselor condition × evaluator type & 11.83 & 8, 333.5 & \textless{} .001 \\
\end{longtable}

\hypertarget{s6.5-significant-simple-effect-contrasts}{%
\subsection*{S6.5 Significant Simple-Effect Contrasts}\label{s6.5-significant-simple-effect-contrasts}}

The table below reports the significant primary expert-versus-automated-evaluator simple-effect contrasts from the four-system interaction model using maximum reasoning/effort settings. Negative estimates indicate higher automated-evaluator ratings than expert ratings within the same counselor condition. The full machine-readable simple-effect table, including nonsignificant expert-versus-automated-evaluator contrasts, automated-evaluator contrasts, and counselor-condition contrasts within evaluator type, is available as \path{results/analysis_runs/fable5_claude_code_formal_20260717/mixed_model_results/model_interaction_simple_effect_contrasts.csv} in the public analysis package.

\begin{longtable}[]{@{}
  >{\raggedright\arraybackslash}p{(\columnwidth - 14\tabcolsep) * \real{0.1250}}
  >{\raggedright\arraybackslash}p{(\columnwidth - 14\tabcolsep) * \real{0.1250}}
  >{\raggedright\arraybackslash}p{(\columnwidth - 14\tabcolsep) * \real{0.1250}}
  >{\raggedleft\arraybackslash}p{(\columnwidth - 14\tabcolsep) * \real{0.1250}}
  >{\raggedleft\arraybackslash}p{(\columnwidth - 14\tabcolsep) * \real{0.1250}}
  >{\raggedleft\arraybackslash}p{(\columnwidth - 14\tabcolsep) * \real{0.1250}}
  >{\raggedleft\arraybackslash}p{(\columnwidth - 14\tabcolsep) * \real{0.1250}}
  >{\raggedright\arraybackslash}p{(\columnwidth - 14\tabcolsep) * \real{0.1250}}@{}}
\toprule\noalign{}
\begin{minipage}[b]{\linewidth}\raggedright
Outcome
\end{minipage} & \begin{minipage}[b]{\linewidth}\raggedright
Counselor condition
\end{minipage} & \begin{minipage}[b]{\linewidth}\raggedright
Comparison
\end{minipage} & \begin{minipage}[b]{\linewidth}\raggedleft
Estimate
\end{minipage} & \begin{minipage}[b]{\linewidth}\raggedleft
SE
\end{minipage} & \begin{minipage}[b]{\linewidth}\raggedleft
df
\end{minipage} & \begin{minipage}[b]{\linewidth}\raggedleft
\emph{t}
\end{minipage} & \begin{minipage}[b]{\linewidth}\raggedright
\emph{p}
\end{minipage} \\
\midrule\noalign{}
\endhead
\bottomrule\noalign{}
\endlastfoot
CCT & GPT-minimal & Expert − GPT-5.5 & −0.95 & 0.26 & 36.17 & −3.59 & .008 \\
CCT & GPT-SMDP & Expert − Gemini 3.5 Flash & −1.37 & 0.26 & 36.17 & −5.21 & \textless{} .001 \\
CCT & Claude-SMDP & Expert − Gemini 3.5 Flash & −1.18 & 0.26 & 36.17 & −4.50 & \textless{} .001 \\
SST & GPT-minimal & Expert − GPT-5.5 & −1.65 & 0.22 & 46.68 & −7.43 & \textless{} .001 \\
SST & GPT-minimal & Expert − Gemini 3.5 Flash & −1.57 & 0.22 & 46.68 & −7.05 & \textless{} .001 \\
SST & GPT-minimal & Expert − Claude Opus 4.8 & −1.13 & 0.22 & 46.68 & −5.06 & \textless{} .001 \\
SST & GPT-minimal & Expert − Fable 5 & −1.15 & 0.22 & 46.68 & −5.18 & \textless{} .001 \\
SST & GPT-SMDP & Expert − GPT-5.5 & −0.85 & 0.22 & 46.68 & −3.81 & .004 \\
SST & GPT-SMDP & Expert − Gemini 3.5 Flash & −1.46 & 0.22 & 46.68 & −6.55 & \textless{} .001 \\
SST & GPT-SMDP & Expert − Claude Opus 4.8 & −0.93 & 0.22 & 46.68 & −4.18 & .001 \\
SST & GPT-SMDP & Expert − Fable 5 & −0.96 & 0.22 & 46.68 & −4.31 & \textless{} .001 \\
SST & Claude-SMDP & Expert − GPT-5.5 & −1.03 & 0.22 & 46.68 & −4.64 & \textless{} .001 \\
SST & Claude-SMDP & Expert − Gemini 3.5 Flash & −1.45 & 0.22 & 46.68 & −6.51 & \textless{} .001 \\
SST & Claude-SMDP & Expert − Claude Opus 4.8 & −0.73 & 0.22 & 46.68 & −3.27 & .017 \\
SST & Claude-SMDP & Expert − Fable 5 & −0.84 & 0.22 & 46.68 & −3.77 & .004 \\
PAR & GPT-SMDP & Expert − GPT-5.5 & −0.76 & 0.25 & 43.63 & −3.11 & .026 \\
PAR & GPT-SMDP & Expert − Gemini 3.5 Flash & −1.18 & 0.25 & 43.63 & −4.81 & \textless{} .001 \\
PAR & Claude-SMDP & Expert − GPT-5.5 & −0.93 & 0.25 & 43.63 & −3.80 & .004 \\
PAR & Claude-SMDP & Expert − Gemini 3.5 Flash & −0.74 & 0.25 & 43.63 & −3.01 & .033 \\
EMP & GPT-minimal & Expert − GPT-5.5 & −1.21 & 0.24 & 41.76 & −5.10 & \textless{} .001 \\
EMP & GPT-SMDP & Expert − GPT-5.5 & −0.81 & 0.24 & 41.76 & −3.41 & .012 \\
EMP & GPT-SMDP & Expert − Fable 5 & −0.70 & 0.24 & 41.76 & −2.94 & .040 \\
EMP & Claude-SMDP & Expert − GPT-5.5 & −0.74 & 0.24 & 41.76 & −3.10 & .027 \\
EMP & Claude-SMDP & Expert − Gemini 3.5 Flash & −0.71 & 0.24 & 41.76 & −2.99 & .036 \\
OVR & GPT-minimal & Expert − GPT-5.5 & −2.22 & 0.16 & 58.01 & −13.47 & \textless{} .001 \\
OVR & GPT-minimal & Expert − Gemini 3.5 Flash & −2.33 & 0.16 & 58.01 & −14.15 & \textless{} .001 \\
OVR & GPT-minimal & Expert − Claude Opus 4.8 & −1.64 & 0.16 & 58.01 & −9.94 & \textless{} .001 \\
OVR & GPT-minimal & Expert − Fable 5 & −2.06 & 0.16 & 58.01 & −12.46 & \textless{} .001 \\
OVR & GPT-SMDP & Expert − GPT-5.5 & −0.92 & 0.16 & 58.01 & −5.55 & \textless{} .001 \\
OVR & GPT-SMDP & Expert − Gemini 3.5 Flash & −1.69 & 0.16 & 58.01 & −10.27 & \textless{} .001 \\
OVR & GPT-SMDP & Expert − Claude Opus 4.8 & −0.75 & 0.16 & 58.01 & −4.54 & \textless{} .001 \\
OVR & GPT-SMDP & Expert − Fable 5 & −0.75 & 0.16 & 58.01 & −4.54 & \textless{} .001 \\
OVR & Claude-SMDP & Expert − GPT-5.5 & −1.45 & 0.16 & 58.01 & −8.79 & \textless{} .001 \\
OVR & Claude-SMDP & Expert − Gemini 3.5 Flash & −1.75 & 0.16 & 58.01 & −10.64 & \textless{} .001 \\
OVR & Claude-SMDP & Expert − Claude Opus 4.8 & −1.12 & 0.16 & 58.01 & −6.77 & \textless{} .001 \\
OVR & Claude-SMDP & Expert − Fable 5 & −1.28 & 0.16 & 58.01 & −7.78 & \textless{} .001 \\
\end{longtable}

\hypertarget{s6.6-sensitivity-analyses-specified-after-preliminary-model-inspection}{%
\subsection*{S6.6 Sensitivity Analyses Specified After Preliminary Model Inspection}\label{s6.6-sensitivity-analyses-specified-after-preliminary-model-inspection}}

Additional sensitivity analyses addressed the fixed-stimulus design and the repeated-measures structure of the expert ratings. These analyses were specified after preliminary model inspection and are therefore reported as robustness checks rather than as prespecified primary analyses. Full machine-readable outputs are in \path{results/analysis_runs/fable5_claude_code_formal_20260717/v9_sensitivity/} in the public analysis package.

\hypertarget{s6.7-within-evaluation-unit-paired-difference-sensitivity-models}{%
\subsection*{S6.7 Within-Evaluation-Unit Paired-Difference Sensitivity Models}\label{s6.7-within-evaluation-unit-paired-difference-sensitivity-models}}

\begin{longtable}[]{@{}
  >{\raggedright\arraybackslash}p{(\columnwidth - 10\tabcolsep) * \real{0.1667}}
  >{\raggedright\arraybackslash}p{(\columnwidth - 10\tabcolsep) * \real{0.1667}}
  >{\raggedleft\arraybackslash}p{(\columnwidth - 10\tabcolsep) * \real{0.1667}}
  >{\raggedright\arraybackslash}p{(\columnwidth - 10\tabcolsep) * \real{0.1667}}
  >{\raggedright\arraybackslash}p{(\columnwidth - 10\tabcolsep) * \real{0.1667}}
  >{\raggedright\arraybackslash}p{(\columnwidth - 10\tabcolsep) * \real{0.1667}}@{}}
\toprule\noalign{}
\begin{minipage}[b]{\linewidth}\raggedright
Outcome
\end{minipage} & \begin{minipage}[b]{\linewidth}\raggedright
Contrast
\end{minipage} & \begin{minipage}[b]{\linewidth}\raggedleft
Estimate
\end{minipage} & \begin{minipage}[b]{\linewidth}\raggedright
95\% CI
\end{minipage} & \begin{minipage}[b]{\linewidth}\raggedright
Holm \emph{p}
\end{minipage} & \begin{minipage}[b]{\linewidth}\raggedright
BH \emph{q}
\end{minipage} \\
\midrule\noalign{}
\endhead
\bottomrule\noalign{}
\endlastfoot
CCT & Claude-SMDP − GPT-minimal & 0.85 & {[}0.45, 1.25{]} & .002 & .001 \\
CCT & GPT-SMDP − GPT-minimal & 0.84 & {[}0.41, 1.28{]} & .002 & .002 \\
CCT & Claude-SMDP − GPT-SMDP & −0.00 & {[}−0.36, 0.35{]} & .976 & .976 \\
SST & Claude-SMDP − GPT-minimal & 0.39 & {[}−0.02, 0.81{]} & .179 & .100 \\
SST & GPT-SMDP − GPT-minimal & 0.24 & {[}−0.26, 0.73{]} & .644 & .439 \\
SST & Claude-SMDP − GPT-SMDP & 0.12 & {[}−0.25, 0.50{]} & .644 & .615 \\
PAR & Claude-SMDP − GPT-minimal & 0.82 & {[}0.50, 1.13{]} & \textless{} .001 & \textless{} .001 \\
PAR & GPT-SMDP − GPT-minimal & 1.04 & {[}0.60, 1.48{]} & \textless{} .001 & \textless{} .001 \\
PAR & Claude-SMDP − GPT-SMDP & −0.26 & {[}−0.63, 0.11{]} & .151 & .226 \\
EMP & Claude-SMDP − GPT-minimal & 0.98 & {[}0.57, 1.38{]} & .001 & .001 \\
EMP & GPT-SMDP − GPT-minimal & 0.93 & {[}0.51, 1.35{]} & .001 & .001 \\
EMP & Claude-SMDP − GPT-SMDP & 0.05 & {[}−0.38, 0.48{]} & .813 & .938 \\
OVR & Claude-SMDP − GPT-minimal & 0.81 & {[}0.49, 1.13{]} & \textless{} .001 & \textless{} .001 \\
OVR & GPT-SMDP − GPT-minimal & 0.80 & {[}0.44, 1.16{]} & \textless{} .001 & \textless{} .001 \\
OVR & Claude-SMDP − GPT-SMDP & −0.01 & {[}−0.39, 0.38{]} & .974 & .976 \\
\end{longtable}

\emph{Note.} Estimates are within-evaluation-unit paired differences for the first condition minus the second condition in each contrast; positive estimates indicate that the first condition scored higher. Each model used 54 evaluator-by-client-profile units from 15 evaluators and included client profile as a fixed effect and evaluator as a random intercept. CI = confidence interval; Holm \emph{p} = Holm-adjusted \emph{p} value within outcome; BH \emph{q} = Benjamini--Hochberg-adjusted \emph{q} value across the 15 paired contrasts. CCT = Cultivating Change Talk; SST = Softening Sustain Talk; PAR = Partnership; EMP = Empathy; OVR = Overall evaluation; SMDP = Structured Multi-step Dialogue Prompt.

In this fixed benchmark, the paired-difference analysis indicated higher aggregated expert-panel ratings for GPT-SMDP than GPT-minimal on CCT, PAR, EMP, and OVR. The SST contrast was positive but did not survive multiplicity adjustment. Claude-SMDP also exceeded GPT-minimal on CCT, PAR, EMP, and OVR, whereas no paired-difference contrast supported a difference between the two SMDP counselor conditions. The latter result was not an equivalence test.

\hypertarget{s6.8-robust-hc3-paired-difference-sensitivity-analysis}{%
\subsection*{S6.8 Robust HC3 Paired-Difference Sensitivity Analysis}\label{s6.8-robust-hc3-paired-difference-sensitivity-analysis}}

A fixed-profile linear model with HC3 robust standard errors produced the same substantive pattern for the direct GPT-SMDP minus GPT-minimal contrast. Evidence for the SST contrast remained sensitive to multiplicity adjustment: its unadjusted \emph{p} value was .048, but its Holm-adjusted \emph{p} value was .096 and its BH \emph{q} value was .072.

\begin{longtable}[]{@{}llrlll@{}}
\toprule\noalign{}
Outcome & Contrast & Estimate & 95\% CI & Holm \emph{p} & BH \emph{q} \\
\midrule\noalign{}
\endhead
\bottomrule\noalign{}
\endlastfoot
CCT & GPT-SMDP − GPT-minimal & 0.87 & {[}0.61, 1.14{]} & \textless{} .001 & \textless{} .001 \\
SST & GPT-SMDP − GPT-minimal & 0.33 & {[}0.00, 0.66{]} & .096 & .072 \\
PAR & GPT-SMDP − GPT-minimal & 1.11 & {[}0.80, 1.42{]} & \textless{} .001 & \textless{} .001 \\
EMP & GPT-SMDP − GPT-minimal & 0.93 & {[}0.62, 1.24{]} & \textless{} .001 & \textless{} .001 \\
OVR & GPT-SMDP − GPT-minimal & 0.83 & {[}0.57, 1.09{]} & \textless{} .001 & \textless{} .001 \\
\end{longtable}

\hypertarget{s6.9-supportive-evaluation-unit-score-level-model}{%
\subsection*{S6.9 Supportive Evaluation-Unit Score-Level Model}\label{s6.9-supportive-evaluation-unit-score-level-model}}

As a supportive score-level analysis, a mixed model added an evaluator-by-profile evaluation-unit random intercept to the expert score-level model. The counselor-condition likelihood-ratio test was significant for all outcomes. However, the CCT, EMP, and OVR models produced singular fits, indicating that the additional evaluation-unit variance component was not consistently estimable in this sparse fixed-stimulus dataset.

\begin{longtable}[]{@{}lrrll@{}}
\toprule\noalign{}
Outcome & LR statistic & df & \emph{p} & Singular fit \\
\midrule\noalign{}
\endhead
\bottomrule\noalign{}
\endlastfoot
CCT & 54.74 & 2 & \textless{} .001 & Yes \\
SST & 11.81 & 2 & .003 & No \\
PAR & 55.81 & 2 & \textless{} .001 & No \\
EMP & 46.34 & 2 & \textless{} .001 & Yes \\
OVR & 50.30 & 2 & \textless{} .001 & Yes \\
\end{longtable}

Pairwise contrasts from this supportive model reproduced the primary score-level pattern: GPT-minimal scored lower than GPT-SMDP and Claude-SMDP on all outcomes, whereas GPT-SMDP and Claude-SMDP did not differ significantly. Because of the singular fits, this model is interpreted as a supportive sensitivity analysis and is not used as the main basis for inference.

\hypertarget{s6.10-ordinal-mixed-model-sensitivity-analyses}{%
\subsection*{S6.10 Ordinal Mixed-Model Sensitivity Analyses}\label{s6.10-ordinal-mixed-model-sensitivity-analyses}}

Ordinal mixed models treated the 1--5 half-point ratings as ordered categories. Expert-only ordinal models detected counselor-condition effects for all outcomes, and ordinal expert--LLM interaction models detected evaluator-type by counselor-condition interactions for all outcomes. Ordinal expert pairwise contrasts showed GPT-minimal scoring lower than both SMDP conditions on all outcomes and no significant difference between GPT-SMDP and Claude-SMDP.

\begin{longtable}[]{@{}
  >{\raggedright\arraybackslash}p{(\columnwidth - 12\tabcolsep) * \real{0.1429}}
  >{\raggedleft\arraybackslash}p{(\columnwidth - 12\tabcolsep) * \real{0.1429}}
  >{\raggedleft\arraybackslash}p{(\columnwidth - 12\tabcolsep) * \real{0.1429}}
  >{\raggedright\arraybackslash}p{(\columnwidth - 12\tabcolsep) * \real{0.1429}}
  >{\raggedleft\arraybackslash}p{(\columnwidth - 12\tabcolsep) * \real{0.1429}}
  >{\raggedleft\arraybackslash}p{(\columnwidth - 12\tabcolsep) * \real{0.1429}}
  >{\raggedright\arraybackslash}p{(\columnwidth - 12\tabcolsep) * \real{0.1429}}@{}}
\toprule\noalign{}
\begin{minipage}[b]{\linewidth}\raggedright
Outcome
\end{minipage} & \begin{minipage}[b]{\linewidth}\raggedleft
Expert counselor-condition LR
\end{minipage} & \begin{minipage}[b]{\linewidth}\raggedleft
df
\end{minipage} & \begin{minipage}[b]{\linewidth}\raggedright
\emph{p}
\end{minipage} & \begin{minipage}[b]{\linewidth}\raggedleft
Expert--automated-evaluator interaction LR
\end{minipage} & \begin{minipage}[b]{\linewidth}\raggedleft
df
\end{minipage} & \begin{minipage}[b]{\linewidth}\raggedright
\emph{p}
\end{minipage} \\
\midrule\noalign{}
\endhead
\bottomrule\noalign{}
\endlastfoot
CCT & 49.41 & 2 & \textless{} .001 & 109.46 & 8 & \textless{} .001 \\
SST & 12.58 & 2 & .002 & 34.34 & 8 & \textless{} .001 \\
PAR & 54.88 & 2 & \textless{} .001 & 115.44 & 8 & \textless{} .001 \\
EMP & 48.76 & 2 & \textless{} .001 & 19.65 & 8 & .012 \\
OVR & 50.30 & 2 & \textless{} .001 & 104.02 & 8 & \textless{} .001 \\
\end{longtable}

These ordinal results support the direction of the score-level findings, but they do not eliminate the design and measurement limitations of SST. The paired-difference analysis remains the most direct fixed-stimulus sensitivity analysis for the expert counselor-condition comparisons.

\hypertarget{s6.11-transcript-level-llmexpert-panel-score-differences}{%
\subsection*{S6.11 Transcript-Level LLM--Expert-Panel Score Differences}\label{s6.11-transcript-level-llmexpert-panel-score-differences}}

For each transcript and outcome, the aggregated expert-panel score was computed as the observed mean of the available expert ratings, and the LLM evaluator score was computed as the repeated-run mean for that configured evaluator system. The table reports overall LLM-minus-expert-panel differences across the 18 transcripts. Bootstrap intervals used 2,000 resamples stratified by counselor condition. The aggregated expert-panel scores served as a professional comparison standard, not an error-free gold standard; they reflect 8--10 available ratings per transcript without a common complete panel or formal calibration training. Fable 5 is included as the fourth evaluator system. Its model identity and execution pathway cannot be separated.

\begin{longtable}[]{@{}
  >{\raggedright\arraybackslash}p{(\columnwidth - 8\tabcolsep) * \real{0.2000}}
  >{\raggedright\arraybackslash}p{(\columnwidth - 8\tabcolsep) * \real{0.2000}}
  >{\raggedleft\arraybackslash}p{(\columnwidth - 8\tabcolsep) * \real{0.2000}}
  >{\raggedright\arraybackslash}p{(\columnwidth - 8\tabcolsep) * \real{0.2000}}
  >{\raggedleft\arraybackslash}p{(\columnwidth - 8\tabcolsep) * \real{0.2000}}@{}}
\toprule\noalign{}
\begin{minipage}[b]{\linewidth}\raggedright
Evaluator model
\end{minipage} & \begin{minipage}[b]{\linewidth}\raggedright
Outcome
\end{minipage} & \begin{minipage}[b]{\linewidth}\raggedleft
Mean signed difference
\end{minipage} & \begin{minipage}[b]{\linewidth}\raggedright
95\% bootstrap interval
\end{minipage} & \begin{minipage}[b]{\linewidth}\raggedleft
Mean absolute difference
\end{minipage} \\
\midrule\noalign{}
\endhead
\bottomrule\noalign{}
\endlastfoot
GPT-5.5 & CCT & 0.73 & {[}0.63, 0.83{]} & 0.73 \\
GPT-5.5 & SST & 1.18 & {[}1.03, 1.31{]} & 1.18 \\
GPT-5.5 & PAR & 0.76 & {[}0.65, 0.87{]} & 0.76 \\
GPT-5.5 & EMP & 0.99 & {[}0.85, 1.12{]} & 0.99 \\
GPT-5.5 & OVR & 1.53 & {[}1.43, 1.64{]} & 1.53 \\
Gemini 3.5 Flash & CCT & 0.83 & {[}0.67, 1.00{]} & 0.97 \\
Gemini 3.5 Flash & SST & 1.50 & {[}1.39, 1.59{]} & 1.50 \\
Gemini 3.5 Flash & PAR & 0.46 & {[}0.25, 0.67{]} & 0.92 \\
Gemini 3.5 Flash & EMP & 0.69 & {[}0.50, 0.89{]} & 0.73 \\
Gemini 3.5 Flash & OVR & 1.93 & {[}1.83, 2.05{]} & 1.93 \\
Claude Opus 4.8 & CCT & −0.08 & {[}−0.22, 0.05{]} & 0.25 \\
Claude Opus 4.8 & SST & 0.93 & {[}0.78, 1.06{]} & 0.95 \\
Claude Opus 4.8 & PAR & −0.02 & {[}−0.13, 0.12{]} & 0.21 \\
Claude Opus 4.8 & EMP & 0.42 & {[}0.34, 0.52{]} & 0.42 \\
Claude Opus 4.8 & OVR & 1.17 & {[}1.03, 1.31{]} & 1.17 \\
Fable 5 & CCT & 0.12 & {[}0.02, 0.22{]} & 0.28 \\
Fable 5 & SST & 0.99 & {[}0.82, 1.12{]} & 1.01 \\
Fable 5 & PAR & 0.37 & {[}0.24, 0.50{]} & 0.49 \\
Fable 5 & EMP & 0.62 & {[}0.48, 0.76{]} & 0.64 \\
Fable 5 & OVR & 1.37 & {[}1.24, 1.51{]} & 1.37 \\
\end{longtable}

Transcript-level aggregation reproduced the main score-difference pattern. All four LLM evaluator systems assigned higher OVR and SST scores than the aggregated expert-panel means. Claude Opus 4.8 had the smallest overall absolute difference, with a mean signed difference of +0.49 and a mean absolute difference of 0.60 averaged across outcomes; it was also nearest to the panel aggregate on CCT and PAR. Fable 5 showed a mean signed difference of +0.69 and a mean absolute difference of 0.76 averaged across outcomes. Its CCT difference was small, but its SST and OVR differences were substantial. GPT-5.5 and Gemini 3.5 Flash showed positive mean differences across all outcomes and larger overall absolute differences. This subsection addresses expert-panel alignment; run-to-run score reliability is summarized separately by ICCs, and discrimination of counselor-condition differences is addressed by the interaction and simple-effect analyses above.

\hypertarget{s6.12-model-diagnostics}{%
\subsection*{S6.12 Model Diagnostics}\label{s6.12-model-diagnostics}}

The 15 primary paired-difference mixed models did not produce singular-fit flags. In contrast, the supportive evaluation-unit score-level models produced singular fits for CCT, EMP, and OVR. This pattern suggests that the paired-difference approach more stably represented the evaluator-by-profile repeated-measures structure in the present fixed-stimulus dataset. The supportive score-level model attempted to estimate an additional variance component that the data did not consistently support.

\hypertarget{s6.13-comparison-using-lower-reasoningeffort-settings}{%
\subsection*{S6.13 Comparison Using Lower Reasoning/Effort Settings}\label{s6.13-comparison-using-lower-reasoningeffort-settings}}

This secondary comparison used the same three API-based evaluator models with lower reasoning/effort settings. The main score-level pattern persisted when maximum reasoning/effort settings were used: all three configured API-based LLM evaluator systems assigned higher SST and OVR scores than the aggregated expert-panel transcript means. Claude Opus 4.8 remained nearest to the panel aggregate on CCT and PAR under both configurations. The Fable 5 evaluation described in S3 used maximum reasoning effort and is therefore not part of this comparison.

\begin{longtable}[]{@{}
  >{\raggedright\arraybackslash}p{(\columnwidth - 16\tabcolsep) * \real{0.1111}}
  >{\raggedright\arraybackslash}p{(\columnwidth - 16\tabcolsep) * \real{0.1111}}
  >{\raggedleft\arraybackslash}p{(\columnwidth - 16\tabcolsep) * \real{0.1111}}
  >{\raggedleft\arraybackslash}p{(\columnwidth - 16\tabcolsep) * \real{0.1111}}
  >{\raggedleft\arraybackslash}p{(\columnwidth - 16\tabcolsep) * \real{0.1111}}
  >{\raggedleft\arraybackslash}p{(\columnwidth - 16\tabcolsep) * \real{0.1111}}
  >{\raggedleft\arraybackslash}p{(\columnwidth - 16\tabcolsep) * \real{0.1111}}
  >{\raggedleft\arraybackslash}p{(\columnwidth - 16\tabcolsep) * \real{0.1111}}
  >{\raggedleft\arraybackslash}p{(\columnwidth - 16\tabcolsep) * \real{0.1111}}@{}}
\toprule\noalign{}
\begin{minipage}[b]{\linewidth}\raggedright
Evaluator model
\end{minipage} & \begin{minipage}[b]{\linewidth}\raggedright
Run
\end{minipage} & \begin{minipage}[b]{\linewidth}\raggedleft
Mean signed difference across outcomes
\end{minipage} & \begin{minipage}[b]{\linewidth}\raggedleft
Mean absolute difference across outcomes
\end{minipage} & \begin{minipage}[b]{\linewidth}\raggedleft
OVR signed difference
\end{minipage} & \begin{minipage}[b]{\linewidth}\raggedleft
SST signed difference
\end{minipage} & \begin{minipage}[b]{\linewidth}\raggedleft
CCT signed difference
\end{minipage} & \begin{minipage}[b]{\linewidth}\raggedleft
PAR signed difference
\end{minipage} & \begin{minipage}[b]{\linewidth}\raggedleft
EMP signed difference
\end{minipage} \\
\midrule\noalign{}
\endhead
\bottomrule\noalign{}
\endlastfoot
GPT-5.5 & Low/default effort & 1.03 & 1.03 & 1.51 & 1.33 & 0.68 & 0.77 & 0.87 \\
GPT-5.5 & Maximum reasoning/effort & 1.04 & 1.04 & 1.53 & 1.18 & 0.73 & 0.76 & 0.99 \\
Gemini 3.5 Flash & Low/default effort & 0.94 & 1.11 & 1.70 & 1.55 & 0.43 & 0.33 & 0.70 \\
Gemini 3.5 Flash & Maximum reasoning/effort & 1.08 & 1.21 & 1.93 & 1.50 & 0.83 & 0.46 & 0.69 \\
Claude Opus 4.8 & Low/default effort & 0.46 & 0.57 & 0.83 & 1.16 & −0.09 & 0.09 & 0.28 \\
Claude Opus 4.8 & Maximum reasoning/effort & 0.49 & 0.60 & 1.17 & 0.93 & −0.08 & −0.02 & 0.42 \\
\end{longtable}

\emph{Note.} Values are transcript-level API-based LLM-minus-aggregated-expert-panel differences across the same 18 fixed transcripts for the three API-based evaluator systems. Positive values indicate higher LLM ratings. The mean absolute difference column averages the five outcome-specific transcript-level mean absolute differences. Outputs from the lower reasoning/effort configuration are stored separately in \path{results/analysis_runs/latest_low_effort_20260620/}; they are not combined with the analysis using maximum reasoning/effort settings.

\hypertarget{s7-qualitative-evaluation-summary}{%
\section*{S7 Qualitative Evaluation Summary}\label{s7-qualitative-evaluation-summary}}

The qualitative-summary and improvement-detail outputs contain 72 rows each: 18 transcript cases × 4 automated evaluator systems. For the three API-based evaluator systems, qualitative summaries aggregated the three metric score iterations for a given transcript and evaluator model but were generated in separate calls after scoring. For Fable 5, the three completed numeric runs were first integrated deterministically within transcript and outcome; one qualitative-summary task and one subsequent improvement-detail task were then completed per transcript. These outputs therefore summarize advisory review behavior and do not provide the direct rationale for numeric scores. Client-AI naturalness ratings are summarized in S2.4; the present section summarizes counselor-dialogue qualitative review outputs.

\hypertarget{s7.1-revision-move-counts}{%
\subsection*{S7.1 Revision-Move Counts}\label{s7.1-revision-move-counts}}

Revision moves were normalized before summarization. Summary counts represent classifiable improvement points from the two requested improvement points per transcript-evaluator row; detail counts represent the single selected response targeted for concrete revision.

\begin{longtable}[]{@{}
  >{\raggedright\arraybackslash}p{(\columnwidth - 4\tabcolsep) * \real{0.3333}}
  >{\raggedleft\arraybackslash}p{(\columnwidth - 4\tabcolsep) * \real{0.3333}}
  >{\raggedleft\arraybackslash}p{(\columnwidth - 4\tabcolsep) * \real{0.3333}}@{}}
\toprule\noalign{}
\begin{minipage}[b]{\linewidth}\raggedright
Revision move
\end{minipage} & \begin{minipage}[b]{\linewidth}\raggedleft
Qualitative summary points (\emph{n} = 142)
\end{minipage} & \begin{minipage}[b]{\linewidth}\raggedleft
Detail-selected target responses (\emph{n} = 72)
\end{minipage} \\
\midrule\noalign{}
\endhead
\bottomrule\noalign{}
\endlastfoot
Complex reflection & 67 & 52 \\
Summary & 19 & 4 \\
Autonomy-supportive rewording of advice/information & 17 & 9 \\
Affirmation & 31 & 4 \\
Existing question revision & 8 & 3 \\
\end{longtable}

The dominant improvement need was complex reflection: evaluators most often identified missed opportunities to reflect implicit emotion, values, ambivalence, or change talk before moving to questions, advice, or planning. In the maximum-reasoning/effort run, affirmation appeared often in qualitative summaries, especially when evaluators noted unacknowledged persistence, preparation, or self-efficacy. Autonomy-supportive rewording remained concentrated in advice-heavy responses, especially in GPT-minimal dialogues.

\hypertarget{s7.2-revision-moves-by-evaluator-model}{%
\subsection*{S7.2 Revision Moves by Evaluator Model}\label{s7.2-revision-moves-by-evaluator-model}}

The table below combines qualitative-summary and improvement-detail moves. Complex reflection was the most frequent move for all four automated evaluator systems. Across systems, Claude Opus 4.8 had the highest complex-reflection count; Gemini 3.5 Flash had the highest affirmation and question-revision counts; GPT-5.5 had the highest summary and autonomy-supportive rewording counts; and Fable 5 had the second-highest summary count.

\begin{longtable}[]{@{}
  >{\raggedright\arraybackslash}p{(\columnwidth - 10\tabcolsep) * \real{0.1667}}
  >{\raggedleft\arraybackslash}p{(\columnwidth - 10\tabcolsep) * \real{0.1667}}
  >{\raggedleft\arraybackslash}p{(\columnwidth - 10\tabcolsep) * \real{0.1667}}
  >{\raggedleft\arraybackslash}p{(\columnwidth - 10\tabcolsep) * \real{0.1667}}
  >{\raggedleft\arraybackslash}p{(\columnwidth - 10\tabcolsep) * \real{0.1667}}
  >{\raggedleft\arraybackslash}p{(\columnwidth - 10\tabcolsep) * \real{0.1667}}@{}}
\toprule\noalign{}
\begin{minipage}[b]{\linewidth}\raggedright
Evaluator model
\end{minipage} & \begin{minipage}[b]{\linewidth}\raggedleft
Complex reflection
\end{minipage} & \begin{minipage}[b]{\linewidth}\raggedleft
Summary
\end{minipage} & \begin{minipage}[b]{\linewidth}\raggedleft
Autonomy-supportive rewording
\end{minipage} & \begin{minipage}[b]{\linewidth}\raggedleft
Affirmation
\end{minipage} & \begin{minipage}[b]{\linewidth}\raggedleft
Existing question revision
\end{minipage} \\
\midrule\noalign{}
\endhead
\bottomrule\noalign{}
\endlastfoot
GPT-5.5 & 29 & 10 & 12 & 3 & 0 \\
Gemini 3.5 Flash & 29 & 2 & 3 & 13 & 7 \\
Claude Opus 4.8 & 32 & 3 & 6 & 11 & 0 \\
Fable 5 & 29 & 8 & 5 & 8 & 4 \\
\end{longtable}

\hypertarget{s7.3-comparison-with-expert-descriptive-comments}{%
\subsection*{S7.3 Comparison With Expert Descriptive Comments}\label{s7.3-comparison-with-expert-descriptive-comments}}

Expert descriptive comments were available in the analysis dataset. The comparison below is interpretive rather than an additional statistical test because the expert comments were free text and were not coded as model-level qualitative outcomes. It distinguishes the score-and-reason evaluations from the later improvement-support calls.

\begin{longtable}[]{@{}
  >{\raggedright\arraybackslash}p{(\columnwidth - 4\tabcolsep) * \real{0.3333}}
  >{\raggedright\arraybackslash}p{(\columnwidth - 4\tabcolsep) * \real{0.3333}}
  >{\raggedright\arraybackslash}p{(\columnwidth - 4\tabcolsep) * \real{0.3333}}@{}}
\toprule\noalign{}
\begin{minipage}[b]{\linewidth}\raggedright
Comparison target
\end{minipage} & \begin{minipage}[b]{\linewidth}\raggedright
Model or pattern
\end{minipage} & \begin{minipage}[b]{\linewidth}\raggedright
Basis
\end{minipage} \\
\midrule\noalign{}
\endhead
\bottomrule\noalign{}
\endlastfoot
Score-and-reason evaluation overall & Claude Opus 4.8 & In the interpretive comparison, free-text reason fields from scoring calls most closely echoed expert concerns about premature problem-solving, question-heavy structure, insufficient exploration of implicit affect, values, and change talk, and weak collaboration. Numerically, Claude Opus 4.8 was also nearest to the aggregated expert panel on CCT and PAR, although it still assigned higher SST, EMP, and OVR scores. \\
Fable 5 score-level difference & Fable 5 & Fable 5 showed an average signed difference of +0.69 and an average absolute difference of 0.76 across outcomes. Its CCT difference was small, but its SST and OVR differences were substantial. Because model identity and the execution pathway are confounded, these differences cannot be attributed to the model alone. \\
Score-and-reason detection of minimal-instruction weakness & Gemini 3.5 Flash & Gemini 3.5 Flash penalized GPT-minimal on CCT and PAR for advice-led structure and superficial collaboration, matching a subset of expert comments. Its high SST and OVR scores diverged from expert global judgments. \\
Score-and-reason leniency pattern & GPT-5.5 & GPT-5.5 noted formulaic wording and limited emotional depth in some score reasons, but these concerns did not sufficiently lower its scores, especially for OVR, SST, and EMP. \\
Improvement support for reflective depth & All four automated evaluator systems & Complex reflection was the most frequent advisory move for GPT-5.5 (29 moves), Gemini 3.5 Flash (29 moves), Claude Opus 4.8 (32 moves), and Fable 5 (29 moves), matching expert comments about superficial understanding, excessive questioning, and insufficient deepening of emotions, values, and motivation. \\
Improvement support for directiveness and client ownership & GPT-5.5 & GPT-5.5 produced the most autonomy-supportive rewording moves (12); Claude Opus 4.8 and Fable 5 followed with six and five. These moves overlapped with expert comments about premature advice, directiveness, and the need to preserve the client's ownership of change. \\
Improvement support for affirmation and question structure & Gemini 3.5 Flash, Claude Opus 4.8, and Fable 5 & Gemini 3.5 Flash and Claude Opus 4.8 most often selected affirmation (13 and 11 moves), while Fable 5 selected eight. Gemini 3.5 Flash selected seven question-revision moves and Fable 5 selected four, indicating attention to missed recognition and leading-question structure in the advisory outputs. \\
\end{longtable}

\hypertarget{s7.4-detail-selected-revision-moves-by-counselor-condition}{%
\subsection*{S7.4 Detail-Selected Revision Moves by Counselor Condition}\label{s7.4-detail-selected-revision-moves-by-counselor-condition}}

\begin{longtable}[]{@{}
  >{\raggedright\arraybackslash}p{(\columnwidth - 10\tabcolsep) * \real{0.1667}}
  >{\raggedleft\arraybackslash}p{(\columnwidth - 10\tabcolsep) * \real{0.1667}}
  >{\raggedleft\arraybackslash}p{(\columnwidth - 10\tabcolsep) * \real{0.1667}}
  >{\raggedleft\arraybackslash}p{(\columnwidth - 10\tabcolsep) * \real{0.1667}}
  >{\raggedleft\arraybackslash}p{(\columnwidth - 10\tabcolsep) * \real{0.1667}}
  >{\raggedleft\arraybackslash}p{(\columnwidth - 10\tabcolsep) * \real{0.1667}}@{}}
\toprule\noalign{}
\begin{minipage}[b]{\linewidth}\raggedright
Counselor condition
\end{minipage} & \begin{minipage}[b]{\linewidth}\raggedleft
Complex reflection
\end{minipage} & \begin{minipage}[b]{\linewidth}\raggedleft
Summary
\end{minipage} & \begin{minipage}[b]{\linewidth}\raggedleft
Autonomy-supportive rewording
\end{minipage} & \begin{minipage}[b]{\linewidth}\raggedleft
Affirmation
\end{minipage} & \begin{minipage}[b]{\linewidth}\raggedleft
Existing question revision
\end{minipage} \\
\midrule\noalign{}
\endhead
\bottomrule\noalign{}
\endlastfoot
GPT-minimal & 18 & 0 & 6 & 0 & 0 \\
GPT-SMDP & 20 & 1 & 0 & 3 & 0 \\
Claude-SMDP & 14 & 3 & 3 & 1 & 3 \\
\end{longtable}

\hypertarget{s7.5-representative-improvement-detail-examples}{%
\subsection*{S7.5 Representative Improvement-Detail Examples}\label{s7.5-representative-improvement-detail-examples}}

The examples below summarize the Japanese improvement-detail rows in English. They are not additional scoring results; they illustrate how the final detail calls translated qualitative issues into concrete revision directions.

\begin{longtable}[]{@{}
  >{\raggedright\arraybackslash}p{(\columnwidth - 10\tabcolsep) * \real{0.1667}}
  >{\raggedright\arraybackslash}p{(\columnwidth - 10\tabcolsep) * \real{0.1667}}
  >{\raggedright\arraybackslash}p{(\columnwidth - 10\tabcolsep) * \real{0.1667}}
  >{\raggedright\arraybackslash}p{(\columnwidth - 10\tabcolsep) * \real{0.1667}}
  >{\raggedright\arraybackslash}p{(\columnwidth - 10\tabcolsep) * \real{0.1667}}
  >{\raggedright\arraybackslash}p{(\columnwidth - 10\tabcolsep) * \real{0.1667}}@{}}
\toprule\noalign{}
\begin{minipage}[b]{\linewidth}\raggedright
Case
\end{minipage} & \begin{minipage}[b]{\linewidth}\raggedright
Evaluator model
\end{minipage} & \begin{minipage}[b]{\linewidth}\raggedright
Counselor condition
\end{minipage} & \begin{minipage}[b]{\linewidth}\raggedright
Selected revision move
\end{minipage} & \begin{minipage}[b]{\linewidth}\raggedright
Improvement target
\end{minipage} & \begin{minipage}[b]{\linewidth}\raggedright
Representative revision direction
\end{minipage} \\
\midrule\noalign{}
\endhead
\bottomrule\noalign{}
\endlastfoot
1 & Claude Opus 4.8 & GPT-SMDP & Complex reflection & The client expressed a wish for genuine communication and mutual understanding, but the counselor restated the content and immediately moved to a question about prior action. & Reflect the desire to be understood and to work as partners before asking about concrete steps. \\
2 & Claude Opus 4.8 & GPT-SMDP & Affirmation & The client described trying again after a prior disappointing conversation with the supervisor and planning to communicate proactively with colleagues. & Affirm the persistence and preparation shown by re-approaching the supervisor despite the earlier difficulty. \\
2 & GPT-5.5 & GPT-SMDP & Summary & The client articulated motivation, preparation, and intended steps, but the counselor shifted quickly into scheduling questions. & Offer a concise change-talk summary linking the client's reasons, resources, and planned actions. \\
7 & Claude Opus 4.8 & GPT-minimal & Autonomy-supportive rewording of advice/information & The counselor offered several strategies for ambiguous supervisor instructions before fully acknowledging the client's hesitation about asking questions. & Acknowledge the difficulty and frame any confirmation strategy as a small optional step the client can choose. \\
14 & Gemini 3.5 Flash & Claude-SMDP & Existing question revision & The counselor asked what the client could do but immediately supplied options, reducing client ownership. & Keep a single open question and remove the supplied options so the client can generate possibilities. \\
\end{longtable}

\hypertarget{s7.6-expert-qualitative-findings}{%
\subsection*{S7.6 Expert Qualitative Findings}\label{s7.6-expert-qualitative-findings}}

\hypertarget{counselor-simulations}{%
\subsubsection*{Counselor Simulations}\label{counselor-simulations}}

Expert evaluators identified four recurring limitations across counselor conditions: excessive directiveness or advice-giving, insufficient reflection of emotion and unexpressed meaning, limited deepening of change-oriented statements, and an evaluative stance that weakened collaboration. Model-specific comments added nuance. Claude-SMDP often elicited strengths but sometimes dominated the exchange through long responses, ambiguous goal setting, and extended closings. GPT-minimal offered rapid support but moved quickly to problem solving and termination, relying on generic encouragement. GPT-SMDP showed collaborative intent but followed a monotonous sequence and insufficiently explored values or workplace context.

\hypertarget{client-simulations}{%
\subsubsection*{Client Simulations}\label{client-simulations}}

Three themes characterized comments on client realism. First, dialogue style was overly compliant and repetitive, with formal or translation-like phrasing and formulaic endings. Second, emotional expression was narrow: clients showed little resistance, ambivalence, or negative affect and frequently accepted suggestions without qualification. Third, character development lacked concrete age-, workplace-, and family-related context. Experts recommended more natural backchanneling and pauses, wider emotional variation, plausible disagreement or hesitation, and richer situational detail. These limitations are relevant to SST because minimal sustain talk reduces the opportunity to observe how a counselor responds to resistance or ambivalence.

\hypertarget{s8-data-code-registration-and-materials-availability}{%
\section*{S8 Data, Code, Registration, and Materials Availability}\label{s8-data-code-registration-and-materials-availability}}

The public-sharing dataset and analysis code are archived at Zenodo (version 2026.08.26): https://doi.org/10.5281/zenodo.22106028.

The analysis code, generated transcripts, prompts, cleaned LLM-evaluation outputs, expert-rating summaries, mixed-model outputs, qualitative-summary outputs, reliability summaries, and diagnostic sensitivity outputs are maintained in the companion analysis repository. The public-sharing subset includes the Fable 5 outputs used in the analyses and four-system outputs for mixed models, ICCs, equivalence tests, power diagnostics, transcript-level score differences, and qualitative analyses. The earlier Fable 5 GUI evaluation is preserved only for provenance and is excluded from the analyses; outputs from the lower reasoning/effort configuration are stored separately.

The OSF registration and amendments are available at https://doi.org/10.17605/OSF.IO/VU286. The companion OSF project is available at https://osf.io/p8c39/. Each amendment was registered before the analysis newly specified in that amendment was conducted. The amendment concerning Fable 5 was registered after its ratings had been collected and checked for completeness but before the four-system analyses incorporating those ratings were conducted. A later post-analysis amendment clarified the complete-session benchmark and scope of inference; it specified no new analysis and did not change the dataset, hypotheses, outcomes, exclusions, statistical models, contrasts, inference criteria, or results.

Evaluator identities are not linked to ratings in the manuscript or Supplementary Material tables. The shared materials document analyses of synthetic transcripts and configured evaluator-system outputs; they do not include patient records, counseling-client data, or raw identifiable expert-evaluator records.

\end{document}